\documentclass[journal]{IEEEtran}
\usepackage{pstricks} \usepackage{eepic}

\ifCLASSINFOpdf
   \usepackage[dvips]{graphicx}
   \DeclareGraphicsExtensions{.pdf,.jpeg,.png}
\else
   \usepackage[dvips]{graphicx}
   \DeclareGraphicsExtensions{.eps}
\fi
\usepackage{amssymb, dsfont}
\usepackage[cmex10]{amsmath}

\usepackage{epsfig, psfrag}
\usepackage{tikz}
\usetikzlibrary{shapes,arrows, calc}
\tikzstyle{block} = [rectangle, draw, text width=9em, text centered, rounded corners, minimum height=2.3em]% styles for flowcharts
\tikzstyle{box} = [rectangle, draw, text width=18pc, text centered, minimum height=1em]% styles for flowcharts
\usepackage{pgfplots}

\usepackage{array}
\usepackage{mdwmath}
\usepackage{mdwtab}
\usepackage{eqparbox}
\usepackage[tight,footnotesize]{subfigure}
\usepackage[caption=false,font=footnotesize]{subfig}
\usepackage{psfrag}
\usepackage{multirow}
\begin{document}
\title{Inter-View Depth Consistency Testing in Depth Difference Subspace}

\author{Pravin~Kumar~Rana~%,~\IEEEmembership{Student Member,~IEEE,}
        and~Markus~Flierl,~\IEEEmembership{Member,~IEEE}% <-this % stops a space
\thanks{Pravin Kumar Rana is with Tobii AB (publ), Danderyd 182 17, Sweden, e-mail:\{pravin.rana\}@tobii.com.}
\thanks{Markus Flierl is with School of Electrical Engineering and Computer Science, KTH Royal Institute of Technology, Stockholm, 100 44, Sweden, e-mail:\{markus.flierl\}@kth.se.}
%\thanks{The authors are with the ACCESS Linnaeus Center, School of Electrical Engineering, KTH Royal Institute of Technology, Stockholm, SE--100 44, Sweden, e-mail:\{pravin.kumar.rana, markus.flierl\}@ee.kth.se.}% <-this % stops a space
%\thanks{Manuscript received XXXXX XX, 2012; revised XXXXX XX, 2012.}
}
\maketitle

\begin{abstract}
%\boldmath
    Multiview depth imagery will play a critical role in free-viewpoint television. This technology requires high quality virtual view synthesis to enable viewers to move freely in a dynamic real world scene. Depth imagery at different viewpoints is used to synthesize an arbitrary number of novel views. Usually, depth images at multiple viewpoints are estimated individually by stereo-matching algorithms, and hence, show lack of inter-view consistency. This inconsistency affects the quality of view synthesis negatively. This paper proposes a method for depth consistency testing in depth difference subspace to enhance the depth representation of a scene across multiple viewpoints. Furthermore, we propose a view synthesis algorithm that uses the obtained consistency information to improve the visual quality of virtual views at arbitrary viewpoints. Our method helps us to find a linear subspace for our depth difference measurements in which we can test the inter-view consistency efficiently. With this, our approach is able to enhance the depth information for real-world scenes. In combination with our consistency-adaptive view synthesis, we improve the visual experience of the free-viewpoint user. The experiments show that our approach enhances the objective quality of virtual views by up to 1.4 dB. The advantage for the subjective quality is also demonstrated.
\end{abstract}



% \fi
%\IEEEpeerreviewmaketitle

\section{Introduction}
\IEEEPARstart{F}{ree-viewpoint} Television (FTV) will change our current television experience \cite{Tanimoto2011}. FTV will enable viewers to have a dynamic natural 3D-depth impression while freely choosing their viewpoint of real world scenes. This will be facilitated by recent advances in electronic display technology and signal processing systems which permit viewing of scenes from a range of perspectives, and perhaps, for many viewers simultaneously \cite{Urey2011}. Furthermore, the availability of low-cost digital cameras enables us to record easily multiview video (MVV) for FTV. MVV is a set of videos recorded by many video cameras that capture a dynamic natural scene from many viewpoints simultaneously. A viewpoint is a defined distance and angle from which the camera views and records the scene. Usually, the sampling of a natural scene at discrete viewpoints is referred to as plenoptic sampling \cite{Chai2000}.

FTV technology requires to store or transmit an enormous amount of MVV imagery. The aim is to provide a seamless transition among interactively selected viewpoints while maintaining the quality of the perceived 3D-depth impression with multiview displays. The quality of immersive displays is expected to improve in the future by increasing the number of displayed views~\cite{Benzie2007}. The commercialization of FTV will further increase the demand for high-capacity multimedia transmission networks~\cite{Flierl2007}. In recent years, FTV attracted wide attention among researchers and, as a result, many compression techniques have been proposed for MVV imagery \cite{Flierl2007}, \cite{Girod2003}, \cite{Smolic2007}. The Joint Video Team (JVT) of MPEG and VCEG proposed multiview video coding (MVC) as an extension to the existing H.264/AVC compression technology. MVC is a promising approach to transmit a vast amount of MVV imagery~\cite{Vetro2011}. As MVV is a result of capturing the same dynamic natural scene from various viewpoints, the imagery exhibits high inter-view and temporal similarities. MVC exploits efficiently inherent similarities in the MVV imagery for compression. The resulting transmission cost for MVC is approximately proportional to the number of coded views~\cite{Mueller2011}. Therefore, a large number of views cannot be efficiently transmitted using MVC. With only a limited subset of captured texture images, high quality view synthesis is not feasible~\cite{Mueller2011}. However, by utilizing information on the scene geometry such as depth maps, the quality can be improved significantly.

\begin{figure}[t!]
\begin{center}
\graphicspath{{./imgs/inconsistency/}}
\begin{tikzpicture}[node distance=0em, auto, >=stealth]
    %dancer
    \node (00)              [             align=center]{\includegraphics[width=6.5pc, height=4.9pc]{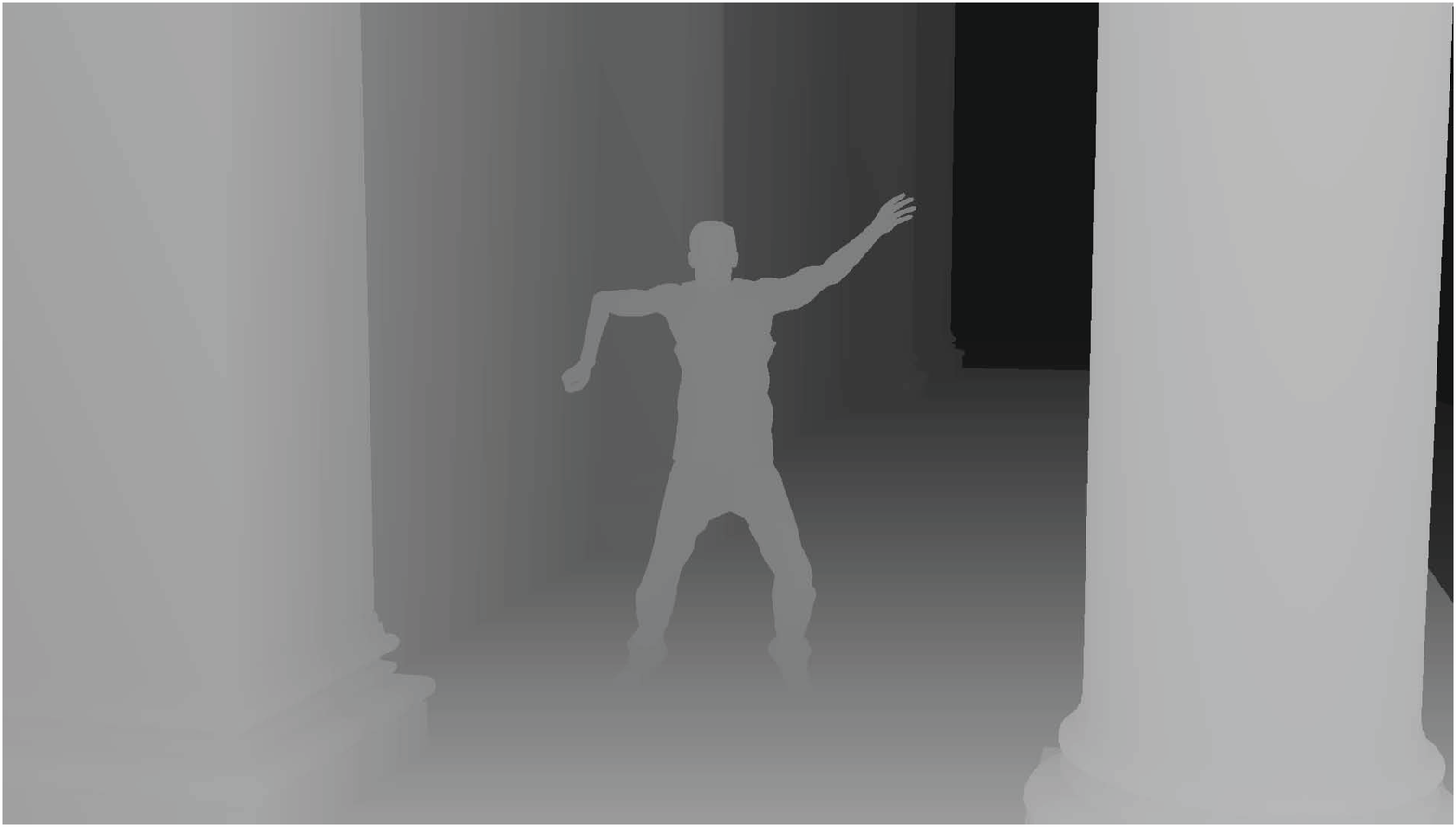}};
    \node (01) at (00.east) [anchor=west, align=center]{\includegraphics[width=6.5pc, height=4.9pc]{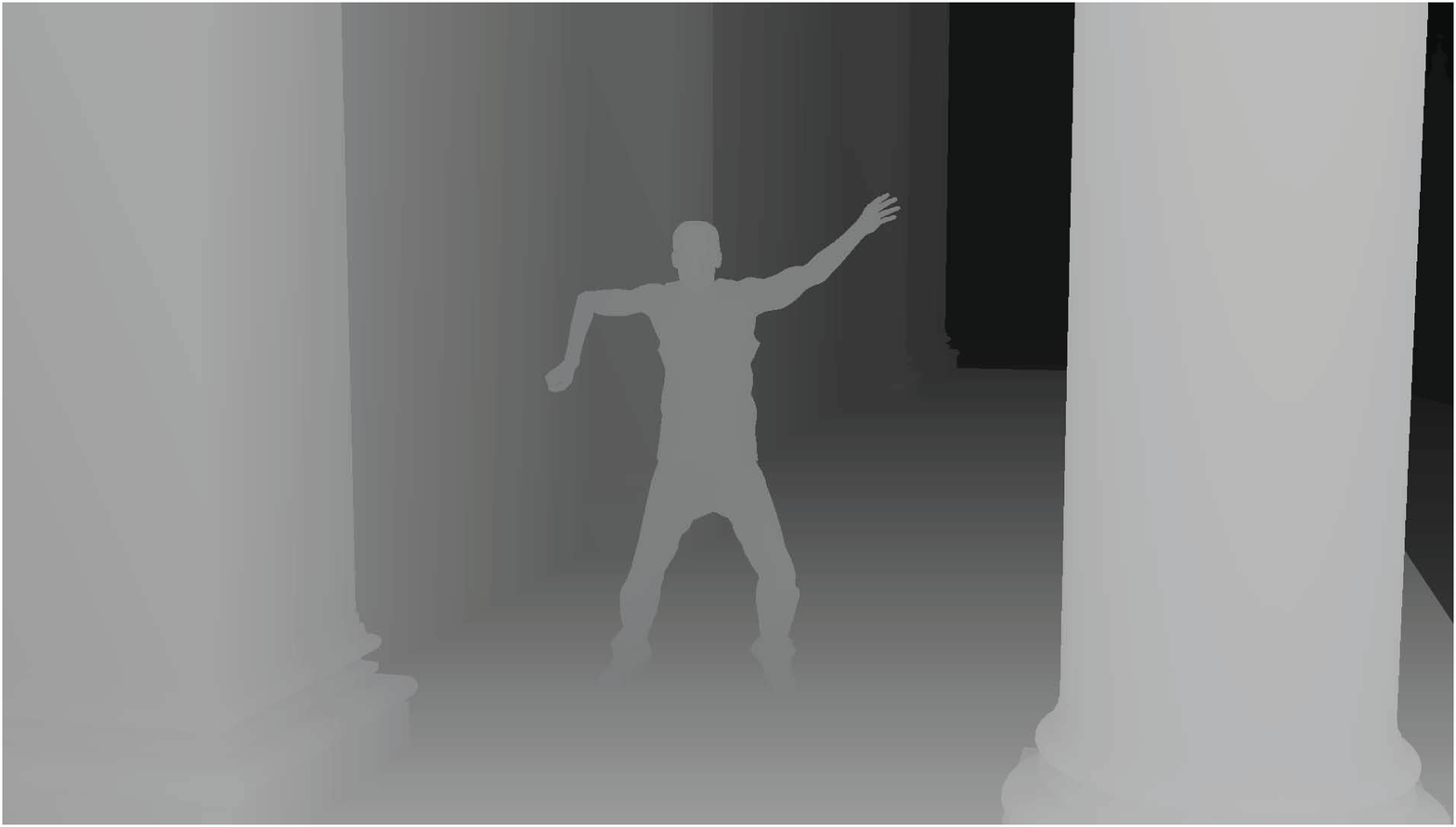}};
    \node (02) at (01.east) [anchor=west, align=center]{\includegraphics[width=6.5pc, height=4.9pc]{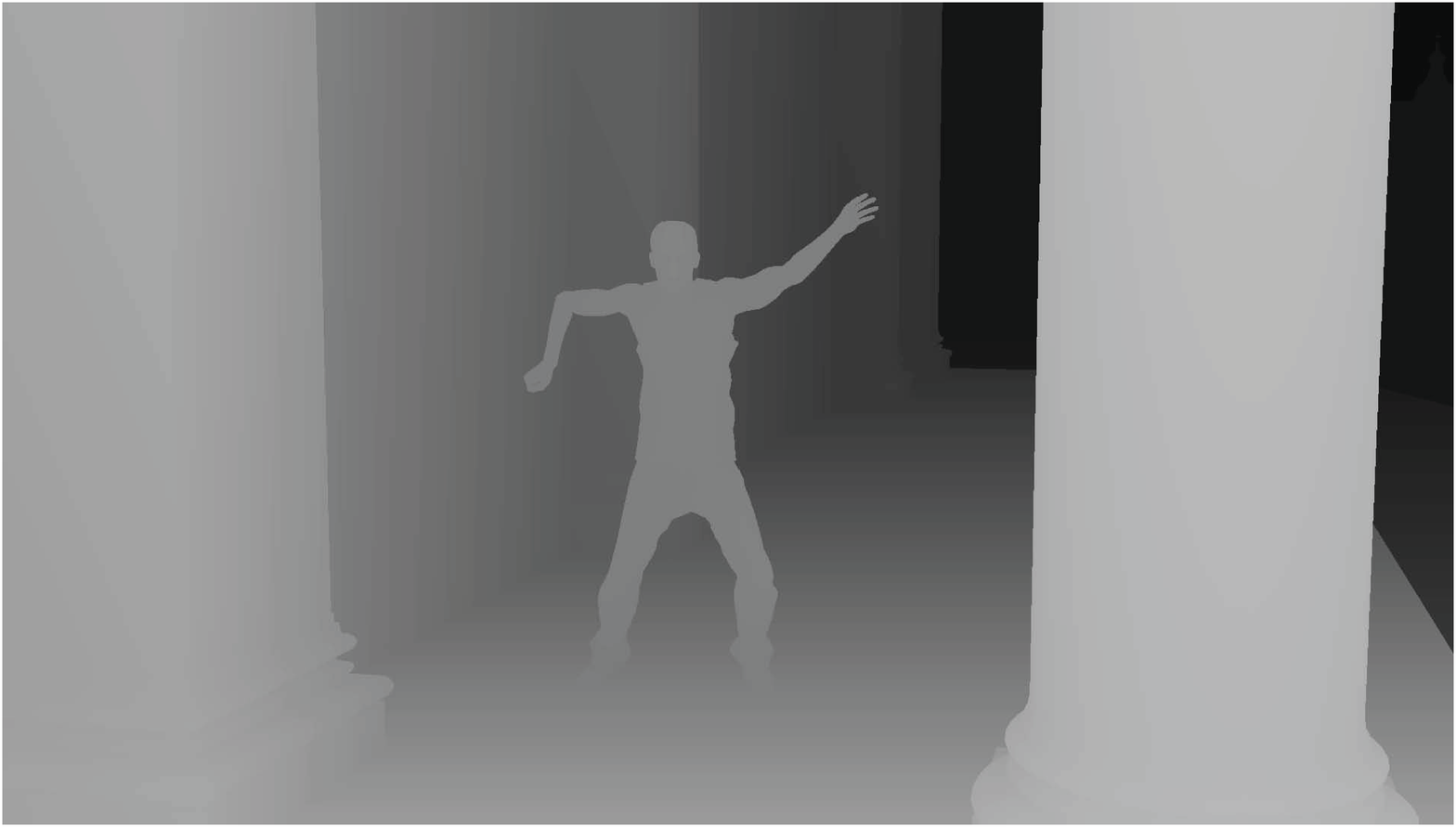}};
    \node (03) at (00.south)[anchor=north] {\footnotesize view $\sharp$ $k-1$};
    \node (04) at (01.south)[anchor=north] {\footnotesize view $\sharp$ $k$ };
    \node (05) at (02.south)[anchor=north] {\footnotesize view $\sharp$ $k+1$};
    \node (06) at (04.south)[anchor=north,align=center]{\footnotesize{(a) Dancer.}};
    %kendo
    \node (11) at (06.south)[anchor=north,align=center]{\includegraphics[width=6.5pc, height=4.9pc]{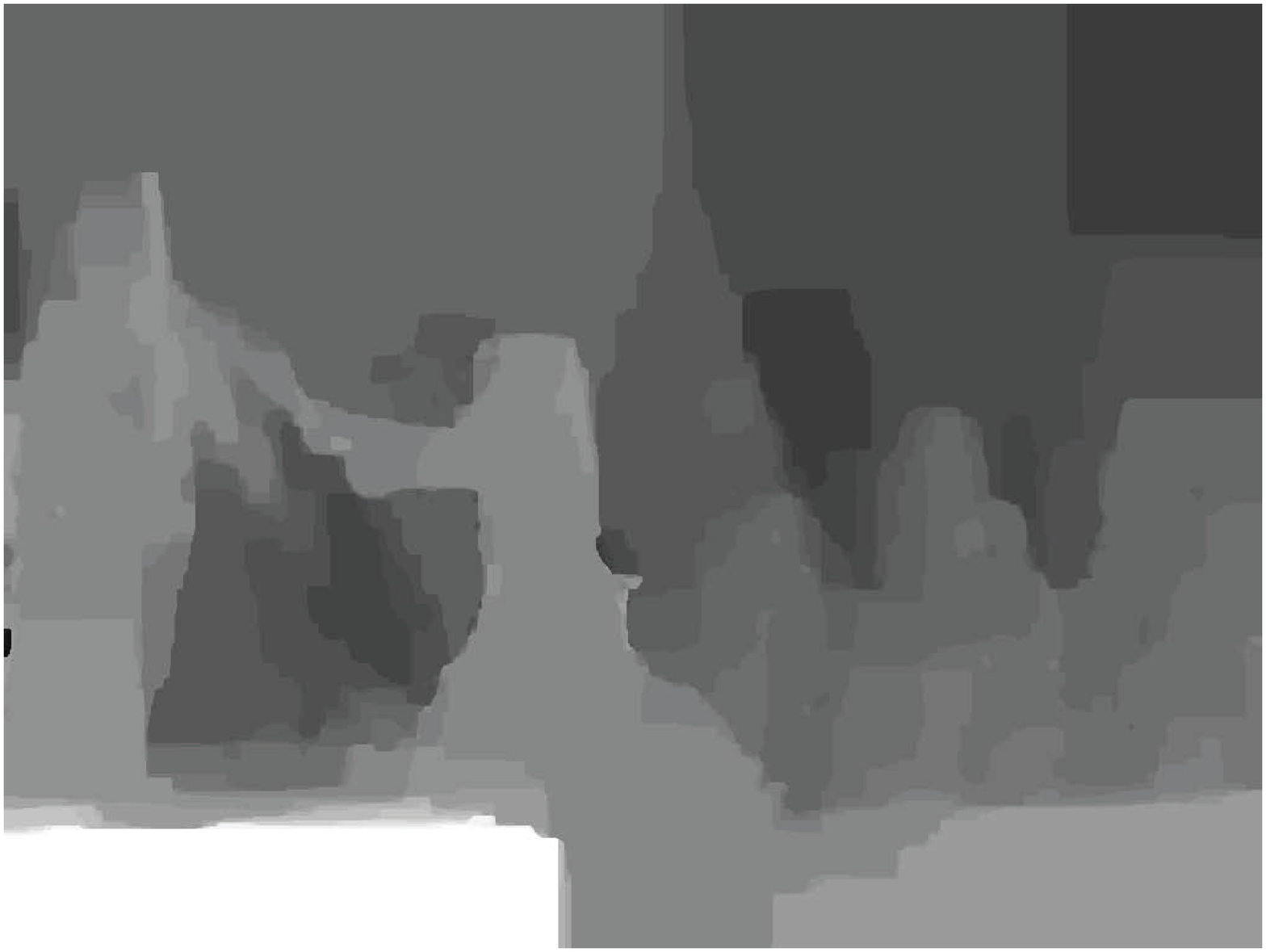}};
    \node (10) at (11.west) [anchor=east, align=center]{\includegraphics[width=6.5pc, height=4.9pc]{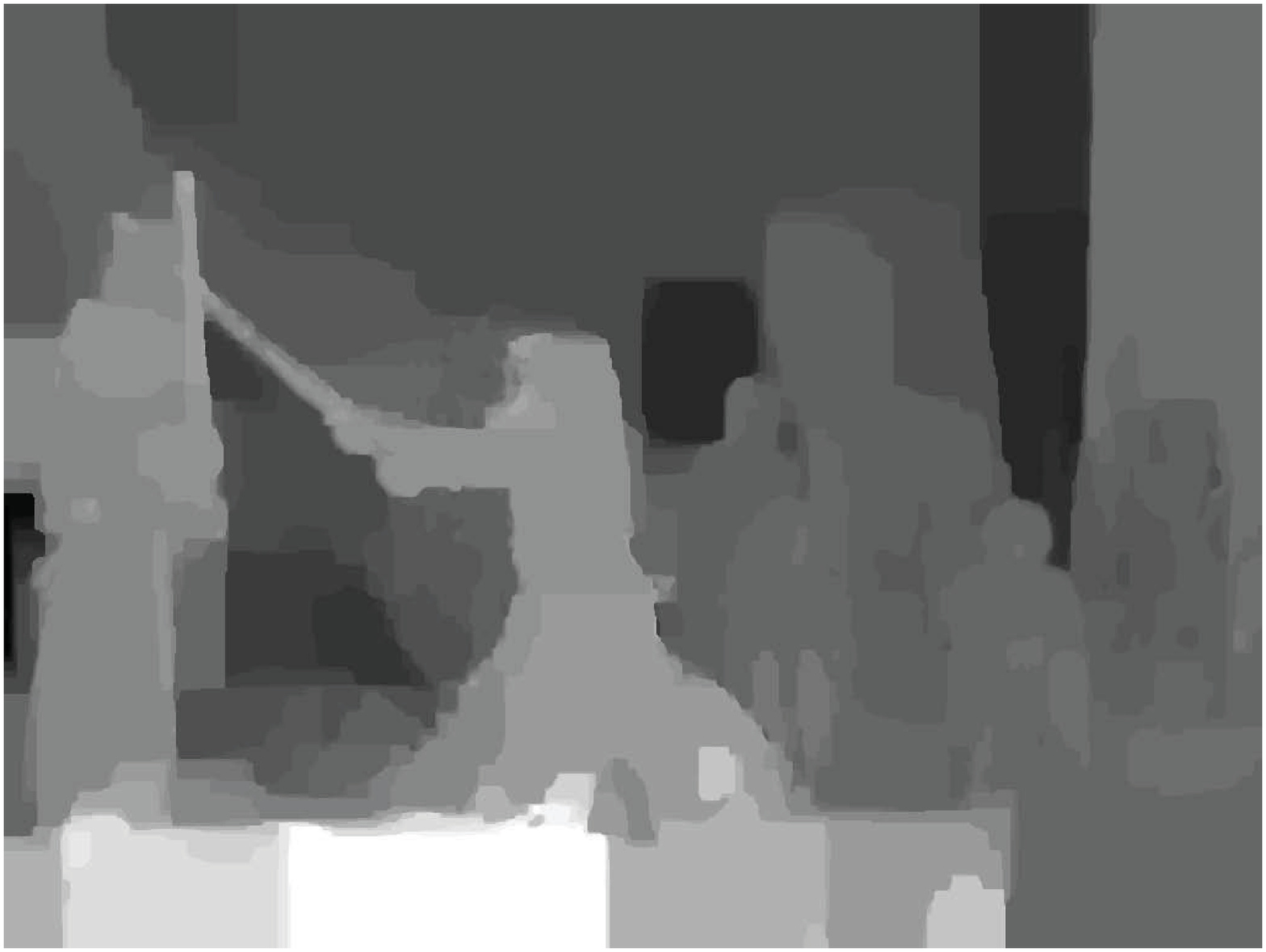}};
    \node (12) at (11.east) [anchor=west, align=center]{\includegraphics[width=6.5pc, height=4.9pc]{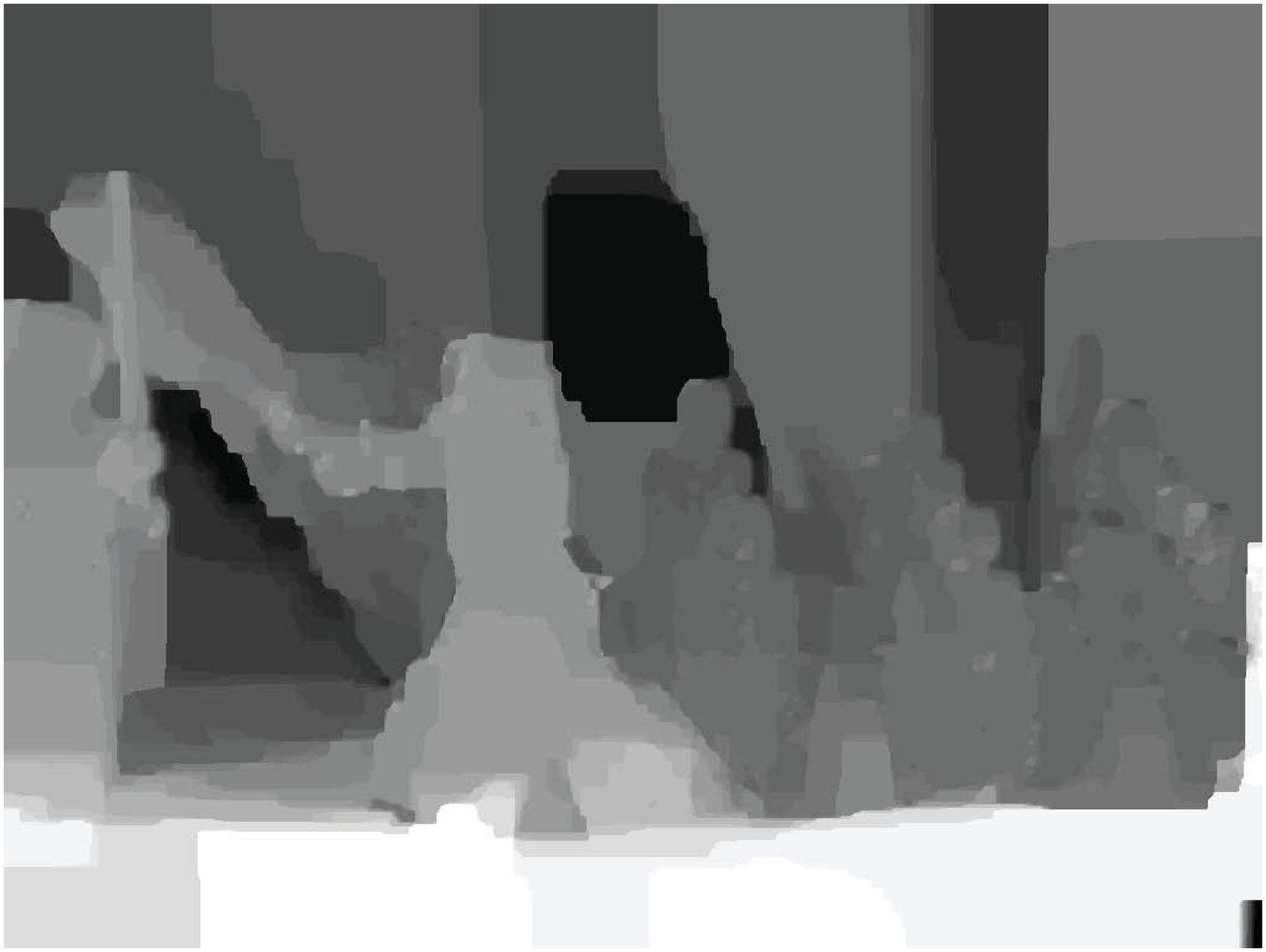}};
    \node (13) at (10) [circle, minimum size=1cm, line width=.1pc, draw, red, align=center]{};
    \node (14) at (11) [circle, minimum size=1cm, line width=.1pc, draw, red, align=center]{};
    \node (15) at (12) [circle, minimum size=1cm, line width=.1pc, draw, red, align=center]{};
    \node (16) at (10.south)[anchor=north] {\footnotesize view $\sharp$ $k-1$};
    \node (17) at (11.south)[anchor=north] {\footnotesize view $\sharp$ $k$};
    \node (18) at (12.south)[anchor=north] {\footnotesize view $\sharp$ $k+1$};
    \node (19) at (17.south)[anchor=north,align=center]{\footnotesize{(b) Kendo.}};
    \draw [line width=.1pc, red][->] (13.east) -- (14.west);
    \draw [line width=.1pc, red][->] (14.east) -- (15.west);
\end{tikzpicture}
\caption{\label{fig:inconsistency}Inter-view inconsistency among multiview depth maps at different viewpoints for two different multiview video imagery as provided by~\cite{MPEG:N12036}. Note, the Dancer test data is a synthetic test material and has consistent depth maps across all viewpoints in (a). For the estimated Kendo depth imagery in (b), the red circles mark prominent inconsistent areas in the depth maps.}
\end{center}
\end{figure}

A depth map is a single channel gray scale image. Each pixel in the depth map represents the shortest distance between the corresponding object point and the given camera plane. Generally, depth maps are compressed by existing video codecs as they contain large smooth areas of constant grey levels. Given a small set of MVV images and its corresponding set of multiview depth (MVD) images, an arbitrary number of views can be synthesized by using depth image based rendering~(DIBR)~\cite{Fehn2004}. The quality of these synthesized views depends significantly on the consistency of the MVD imagery. Usually, depth maps for different viewpoints are estimated independently by establishing stereo correspondences between nearby views only~\cite{Scharstein2002}. The resulting depth information at different viewpoints usually lacks inter-view consistency due to limitations of stereo-matching algorithms, as shown in Fig.~\ref{fig:inconsistency}. Furthermore, depth estimation does not consider inherent temporal similarities within the MVV imagery. This results in temporal depth inconsistency. These inconsistencies affect the quality of view synthesis negatively, and hence, FTV users experience visual discomfort.

The consistency of depth maps is also critical for the efficiency of FTV data formats such as layered depth video (LDV)~\cite{Mueller2008} and structured depth map (SDM)~\cite{Rana2011}. LDV is a format which comprises a reference view and a corresponding reference depth map with additional multiple residual layers to tackle occlusions with respect to the reference viewpoint. In contrast to LDV, SDM consists of a reference depth map and a set of auxiliary depth values at given multiple reference viewpoints. A format similar to SDM is the global view and depth map (GVD) format~\cite{Ishibashi2012} \cite{Suzuki2013}, which has been proposed recently. GVD also seeks consistency among depth maps.

Many methods have been proposed to repair temporal inconsistencies in MVD imagery, for example, by using belief propagation~\cite{Cigla2009}, motion estimation~\cite{Lee2010}, and by exploiting local temporal variations in the MVV imagery~\cite{Fu2010}. In our work~\cite{Rana2010}, an improved DIBR based view synthesis is proposed by exploiting inter-view depth consistency information. However, with recent MPEG activities on 3D video standardization~\cite{MPEG:N12036}, the inter-view depth inconsistency problem became an active research topic and received attention by several researchers. For example, a content adaptive median filtering is proposed in~\cite{Ekmekcioglu2011} to improve temporal and inter-view consistency of depth maps by adapting to edges, motion, and depth range. \cite{Kurc2012} presents an algorithm to reduce the inter-view inconsistency at the preprocessing stage of MVD coding. Joint view depth filtering (JVF) is proposed in~\cite{Li2012} to tackle inter-view depth inconsistency in the coding loop of 3D video coding~\cite{Rusanovskyy2011}. It should be noted, that the JVT solution and our proposed algorithm in~\cite{Rana2010} are very similar in nature. Further, JVF has been adopted by MPEG 3DV for the 3D-AVC specification~\cite{Hannuksela2012} and deployed in the JCT-3V/MPEG reference software 3DV-ATM~\cite{Nokia}. JVF works on real-world depth values, whereas~\cite{Rana2010} operates on depth pixel values. However, these methods do not fully exploit the inherent inter-view similarity to achieve a high-quality FTV user experience.

Our objective in this paper is to exploit efficiently the underlying inter-view similarity among multiview depth maps such that the overall quality of the FTV experience improves significantly. First, the proposed method warps more than two depth maps from multiple reference viewpoints to a predefined viewpoint using the principles of projective geometry~\cite{Rana2010}, where each warped depth value is referred to as a depth hypothesis. Second, it tests the consistency among all depth hypotheses at the predefined viewpoint to obtain inter-view consistency information. For this at any predefined viewpoint pixel, we define a loop difference vector by using all depth hypotheses such that its covariance matrix is always singular. This will help us to find a subspace for our depth difference measurements in which we can test the inter-view consistency efficiently. This is the main idea of this paper. The resulting inter-view consistency information is finally utilized to enrich the free-viewpoint experience by improving the visual quality of the synthesized views at any arbitrary viewpoint. For this, we propose view synthesis based on consistency information.  Furthermore, in contrast to~\cite{Rana2010} and~\cite{Rana2012A}, this paper proposes a method to efficiently enhance depth representations at multiple viewpoints by using inter-view consistency constraints. With our enhanced information on the scene geometry, we demonstrate experimentally that the visual quality of synthesized views improves significantly when compared to conventional algorithms.

The paper is organized as follows: In Section~\ref{sec:analysis of multiview depth imagery}, we briefly discuss multiview depth imagery in the context of depth consistency testing. In Section~\ref{sec:subspace_depth consistency testing}, the proposed depth consistency testing algorithm is described. Section~\ref{sec:applications} discusses the utilization of the resulting inter-view consistency information for depth map enhancement and virtual view synthesis. We present our assessment of the proposed methods in Section~\ref{sec:results and discussion}. Finally, Section~\ref{sec:conclusion} gives concluding remarks.
\section{Multiview Depth Imagery}
\label{sec:analysis of multiview depth imagery}
Consistent and precise depth information on natural 3D scenes is highly desirable for high-quality FTV. Several approaches are available for efficient and reliable depth estimation \cite{Scharstein2002}. Usually, stereo matching algorithms are used  first to establish correspondences between two or more camera images at different viewpoints. The accuracy of the correspondences affects the resulting disparities. For 1D parallel camera arrangements, disparity values $\delta$ as obtained by stereo-matching algorithms are related to real world depth values $z$ by the relation\begin{align}\label{eq:disparity}
    \delta = \frac{f\cdot\Delta l}{z},
\end{align}where $f$ is the focal length of a perspective camera and $\Delta l$ the camera interval. Several techniques have been proposed to refine depth estimates. For example, graph-cut~\cite{Kolmogorov2004}, belief propagation~\cite{Sun2003},~\cite{Felzenszwalb2004A},~\cite{Klaus2006},~\cite{Yang2009}, and modified plane sweeping with segmentation~\cite{Cigla2007B}. Despite such refinements, the quality of depth maps is limited by mismatches due to varying illumination between stereo views and occlusions. Furthermore, independent estimation of depth maps at different viewpoints usually entails inter-view inconsistencies.

Depth maps are commonly represented by eight bit single channel gray scale images. For a given viewpoint $i$  at any time instant, the depth pixel $d_{i}(x,y)\in[0, 255]$ at pixel location $(x,y)$ is related to the real world depth value $z_i(x,y)$ of an object point by the relation
\begin{align}\label{eq:d}
\displaystyle{z_{i}(x,y)}&=\displaystyle{\left[ \frac{d_{i}(x, y)}{255}  \left(\frac{1}{z_{min}}-\frac{1}{z_{max}}\right) + \frac{1}{z_{max}} \right]^{-1}},
\end{align}where $z_{max}$ and $z_{min}$ are maximum and minimum depth values of the captured scene, respectively.

For a given natural dynamic scene, we suppose that the MVV imagery is captured by perspective cameras at multiple viewpoints. The MVD imagery is estimated by stereo-matching algorithms, such as~\cite{MPEG:M16923}, that use the captured MVV imagery. Before we are able to analyze the consistency of the estimated depth information across all viewpoints, we need to align the MVD imagery spatially.
\subsection{Multiple Depth Hypotheses}
\label{subsec:multiple depth warping}
Based on the principles of perspective geometry, 3D warping is a DIBR technique to warp any view from an arbitrary viewpoint to a predefine viewpoint by using depth information and camera calibration parameters. We use 3D warping to ensure spatial alignment of the MVD imagery. We create $k$ depth hypotheses by warping estimated depth maps from $k$ viewpoints to a single viewpoint, say the principal viewpoint $p$ In the following, we briefly review the 3D warping technique for mapping depth maps from a viewpoint $i=1,\ldots,k,$ to the principal viewpoint $p$.

We assume a perspective camera $\mathbf{P}_i$ at a viewpoint $i$, which is described by the intrinsic parameter matrix $\mathbf{A}_i \in\mathbb{R}^{3\times 3}$ and two extrinsic parameters, the rotation matrix $\mathbf{R}_i \in\mathbb{R}^{3\times 3}$ and the translation vector $\mathbf{t}_i \in\mathbb{R}^{3}$. We represent the world and image points in homogeneous coordinates~\cite{Zisserman2004}. The matrices $\mathbf{A}_i$, $\mathbf{R}_i$, and $\mathbf{t}_i$ can be combined efficiently into a single homogeneous camera projection matrix $\mathbf{P}_i=\mathbf{A}_i[\mathbf{R}_i|\mathbf{t}_i] \in\mathbb{R}^{3\times 4}$. The projection of an object point $\displaystyle{[u, v, w]^\mathsf{T}}$ in 3D world coordinates to an image pixel point $[x, y]^\mathsf{T}$ at viewpoint $i$ in 2D image coordinates is given by the following perspective projection relation~\cite{Tian2009}
\begin{align}\label{eq:w1}
    \displaystyle{\alpha_i[x, y, 1]^\mathsf{T}} &= \mathbf{P}_i [u, v, w, 1]^\mathsf{T} = \mathbf{A}_i[\mathbf{R}_i|\mathbf{t}_i][u, v, w, 1]^\mathsf{T},
\end{align}
where $\displaystyle{[u,v,w,1]^\mathsf{T} }$ represents the object point $[u, v, w]^\mathsf{T}$ in homogeneous world coordinates and $[x, y, 1]^\mathsf{T}$ represents the corresponding projected image pixel  position $[x,y]$ in homogeneous image coordinates using the camera matrix $\mathbf{P}_i$. Here, ${(\cdot)}^\mathsf{T}$ is used to represent the transpose operation and $\alpha_i$ is an arbitrary non-zero scalar. The pixel $[x,y]$ can be projected back into the world coordinates by the relation
\begin{align}\label{eq:w2}
    \displaystyle{[\tilde{u}, \tilde{v},  \tilde{w}]^\mathsf{T}} &= \displaystyle{\mathbf{R}_i^{-1} \mathbf{A}_i^{-1} [x, y, 1]^\mathsf{T} z_i(x,y)-\mathbf{R_i^{-1}\mathbf{t}_i}},
\end{align}where $\displaystyle{[ \tilde{u},  \tilde{v},  \tilde{w}]^\mathsf{T}}$ is an erroneous estimate of the object point coordinates $[{u}, {v}, {w}]^\mathsf{T}$. This is due to erroneously estimated depth information ${z_i(x,y)}$ from~(\ref{eq:d}). If the principal camera calibration parameters $\mathbf{A}_p$, $\mathbf{R}_p$, and  $\mathbf{t}_p$ are known, then this erroneous estimate propagates into the back projection of image pixel $[x', y']^\mathsf{T}$ at the principal viewpoint. The relation between the world coordinates of the point $[ \tilde{u},  \tilde{v}, \tilde{w}]^\mathsf{T}$ and the corresponding image pixel $[x', y']^\mathsf{T}$ in the principal view is
\begin{align}\label{eq:w3}
    \displaystyle{\alpha_p [x',y', 1]^\mathsf{T}}&= \displaystyle{\mathbf{A}_p[\mathbf{R}_p|\mathbf{t}_p][\tilde{u}, \tilde{v},\tilde{w}, 1]^\mathsf{T}},
\end{align} where $\displaystyle{[x', y', 1]^\mathsf{T}}$ is the warped image pixel in homogeneous image coordinates and $\alpha_p$ is an arbitrary non-zero scalar. In combination with (4), this
relation describes the 3D warping from a viewpoint $i$ to the principal viewpoint $p$ by using depth information $z_i(x,y)$ from the depth pixel $d_i(x,y)$ of viewpoint $i$. In the following, we denote the warping of depth pixels from the viewpoint $i$ to the viewpoint $p$ by
\begin{align}\label{eq:w4}
    \displaystyle{\hat{d}_p(i; x, y)}&= \displaystyle{d_i(i; x',y')},
\end{align}
where $(x,y)$ denotes the pixel location in view $p$ and $(x',y')$ the corresponding location in view $i$. These warped depth maps are burdened by erroneous depth estimates, and hence, establish depth hypotheses $\displaystyle{\hat{d}_p(i; x,y)}$  at the principal viewpoint $p$. The accuracy of the warping process is limited by the discrete-valued depth information and by the resampling error due to non-integer accurate disparity values. It is common to bound the resampling error by fractional sub-pixel accurate disparity values. Usually, the quantization of depth values as well as the resampling error cause minor artifacts around sharp depth discontinuities. On the other hand, inter-view depth inconsistencies usually lead to severe artifacts such as ghosting. In our work, we focus on the depth inconsistencies and use above depth hypotheses in our testing algorithm to remove efficiently the inter-view inconsistencies among multiview depth imagery.

Moreover, regions which are occluded in the view at viewpoint $i$ may become visible at the principal viewpoint. This is the disocclusion problem of warping, where the information regarding the newly exposed regions is not available for the principal viewpoint. As we use multiview depth imagery, the disocclusion problem at certain viewpoints can be compensated by other available viewpoints where the regions in question are not occluded. In the case that an object point pixel  in the principal view is disoccluded when warping from a given view, the contribution  from that given viewpoint will not be considered during our consistency analysis. However, if the same object point in the principal view is visible from other viewpoints, we will consider these. Note, for efficient testing, we need at least three viable contributions. In certain cases, decisions can be made with two.

\subsection{Consistency Analysis}
\label{subsec:consistency_analysis}
In order to analyze the inter-view depth consistency at any pixel in the principal view, let us define the \emph{loop difference vector}
\begin{equation}
\mathbf{\Delta} = [{\Delta}_{12}, {\Delta}_{23},\ldots, {\Delta}_{k1}]^\mathsf{T} \in \mathcal{L}^k
\end{equation}
as a vector of inter-view depth differences in $k$-dimensional loop space $\mathcal{L}^k$. It uses the available $k$ depth hypotheses at the principal viewpoint, where ${\Delta}_{ij} =\displaystyle{\hat{{d}}_p(i;x,y)- \hat{{d}}_p(j; x,y)}$ is the inter-view depth difference between warped depth values from views $i$ and $j$ to the principal view $p$, where $i, j =1, \ldots, k$. We can interpret ${\Delta}_{ij}$ as a \emph{depth inconsistency evidence} between the corresponding warped depth pair $\displaystyle{(\hat{{d}}_p(i; x,y), \hat{{d}}_p(j; x,y))}$ at the principal pixel  $p$, and hence, between the estimated depth values at viewpoints $i$ and $j$ for a given 3D-point. Note, for any principal pixel, $\mathbf{\Delta}$ satisfies the following \emph{zero-sum constraint},
\begin{align}\label{eq:zero-sum-constraint}
\mathbf{1}^\mathsf{T}\mathbf{\Delta} &= 0,
\end{align}i.e., the loop is closed, where $\mathbf{1}$ is the $k$-dimensional vector with each element equal to one. Due to constraint (\ref{eq:zero-sum-constraint}), elements of a closed loop difference vector are linearly dependent, and hence, are highly correlated. To analyze the consistency efficiently, we represent the loop difference vector in such a way that its elements are uncorrelated, and that most of its energy is concentrated in a low-dimensional subspace. This can be achieved by an orthonormal transformation.

Let $\mathbf{U}=[\mathbf{u}_1,\ldots, \mathbf{u}_k] \in \mathbb{R}^{k \times k}$ be a linear orthonormal transform that maps $\mathbf{\Delta}$ according to\begin{align}
{\mathbf{{\Psi}}} &= \mathbf{U}^\mathsf{T}\mathbf{\Delta},
\end{align}
where $\mathbf{\Psi}= [\Psi_1, \Psi_2,\ldots, \Psi_k]^\mathsf{T} \in \mathbb{R}^{k\times1}$ is the transformed loop difference vector in the loop space and $\mathbf{u}_l\in \mathbb{R}^{k \times 1}$, $l=1, \ldots, k$, are the orthonormal basis vectors of $\mathbf{U}$ such that $\mathbf{U}^\mathsf{T}\mathbf{U} = \mathbf{I}$, where $\mathbf{I}$ is the identity matrix.

Now, let the loop difference vector $\mathbf{\Delta} \in  \mathbb{R}^{k\times1}$ be modeled by a $k$-dimensional random vector with zero mean and covariance matrix
\begin{equation}\label{eq:covrainace_matrix_def}
\mathbf{C}_{\Delta\Delta} := \mathds{E}[\mathbf{\Delta\Delta}^\mathsf{T}] \in  \mathbb{R}^{k \times k},
\end{equation}
where $\mathds{E}[\cdot]$ denotes the expectation operator. As the covariance matrix is a symmetric matrix, the spectral theorem holds and there exists an orthonormal basis consisting of eigenvectors of the covariance matrix. The eigenvectors satisfy
\begin{align}\label{eq:eigen_equation}
\mathbf{C}_{\mathbf{\Delta}\mathbf{\Delta}}\mathbf{u}_l &=\lambda_l \mathbf{u}_l,~\mathbf{u}_l\neq 0,
\end{align}where $\mathbf{u}_l$ is an eigenvector of $\mathbf{C}_{\mathbf{\Delta}\mathbf{\Delta}}$ and $\lambda_l$ the corresponding eigenvalue. The orthonormal basis $\mathbf{U}$ diagonalizes the covariance matrix
\begin{equation}
\mathbf{\Lambda} = \mathbf{U}^\mathsf{T} \mathbf{{C}_{\Delta\Delta}} \mathbf{U},
\end{equation}
where $\mathbf{\Lambda} = diag[\lambda_1, \ldots, \lambda_k] \in  \mathbb{R}^{k \times k}$ is a diagonal matrix whose elements are the eigenvalues of $\mathbf{C}_{\Delta\Delta}$. If we multiply $\mathbf{C}_{\Delta\Delta}$ by the $k$-dimensional vector $\mathbf{1}$, we will get
\begin{align}
\mathbf{C}_{\mathbf{\Delta}\mathbf{\Delta}}\mathbf{1} &= {a}\mathbf{1},
\end{align}where $a$ is a scalar and $\mathbf{1}$ an eigenvector of $\mathbf{C}_{\mathbf{\Delta}\mathbf{\Delta}}$. Due to the zero-sum constraint~(\ref{eq:zero-sum-constraint}), we have
\begin{align}
{a} &= 0 = \lambda_1,
\end{align}where $\lambda_1$ is an eigenvalue of $\mathbf{C}_{\mathbf{\Delta}\mathbf{\Delta}}$.

\subsection{Statistical Model}
\label{subsec:statistical_model}
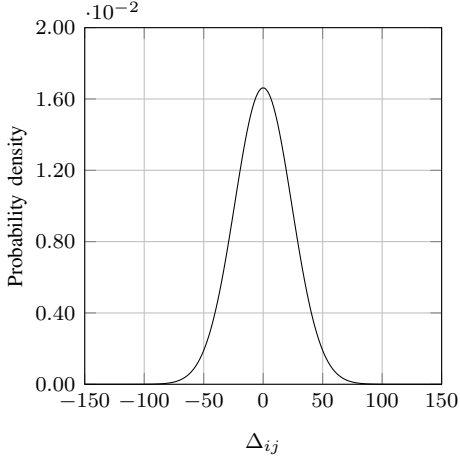
\begin{figure}[t!]
\centering
\pgfmathdeclarefunction{gauss}{2}{\pgfmathparse{1/(#2*sqrt(2*pi))*exp(-((x-#1)^2)/(2*#2^2))}}
\begin{tikzpicture}
\begin{axis}[
  every axis plot post/.append style={domain=-150:150, samples=100, smooth}, enlargelimits=upper, height=15pc, width=15pc,ymax=0.02, xlabel=$\Delta_{ij}$, ylabel={Probability density}, ytick={0, 0.0040, 0.0080, 0.0120, 0.0160, 0.0200, 0.0240}, xtick={-150, -100, -50, 0, 50, 100, 150}, enlargelimits=false, clip=false, grid = major, x tick style={font=\footnotesize}, y tick style={font=\footnotesize}, x tick label style={font=\footnotesize}, y tick label style={font=\footnotesize, /pgf/number format/.cd, fixed, fixed zerofill, precision=2, /tikz/.cd},  y label style={at={(0.07,0.5)}, font={\footnotesize}}, x label style={font=\footnotesize},
  legend style={nodes=right, font=\footnotesize}, legend pos= north east,
  ]
  \addplot[black]{gauss(0,24)};
\end{axis}
\end{tikzpicture}
\caption{\label{fig:delta distribution}Distribution of the elements $\Delta_{ij}$ of the loop difference vector $\mathbf{\Delta}$.}
\end{figure}

Fig.~\ref{fig:delta distribution} shows that the observed distribution of $\mathbf{\Delta}_{ij}$ is well approximated by a normal distribution. Therefore, let us assume that $\mathbf{\Delta}$ follows a wide-sense stationary $k$-variate normal distribution with covariance matrix ${\mathbf{C}_{\mathbf{\Delta}\mathbf{\Delta}}}$ and zero mean. As there is no preference in ordering the elements in $\mathbf{\Delta}$, there is no loss in generality if we assume that, the variance $\mathbf{\sigma}^{2}$ is the same for all $k$ elements of the loop difference vector. With the same argument, there is no preference among pairs of elements in the loop difference vector. As a result, one correlation coefficient $\rho_{\Delta}$ between any two loop difference vector elements is sufficient to capture the correlation. With above assumptions, we write the covariance matrix of a loop difference vector $\mathbf{\Delta}$ as~\cite{Flierl2002}
{\setlength\arraycolsep{0.5em}
\begin{equation}\label{eq:covariance_matrix}
 \displaystyle{\mathbf{C}_{\mathbf{\Delta}\mathbf{\Delta}}}= {\mathbf{\sigma}^{2} \left(
  \begin{array}{cccc}
    1                         & \rho_{\mathbf{\Delta}}  &  \hdots     & \rho_{\mathbf{\Delta}} \\
    \rho_{\mathbf{\Delta}}    & 1                       &  \hdots     & \rho_{\mathbf{\Delta}} \\
    \vdots                    & \vdots                  &  \ddots     & \vdots\\
    \rho_{\mathbf{\Delta}}    & \rho_{\mathbf{\Delta}}  &  \hdots     & 1\\
  \end{array}
\right).}
\end{equation}}The covariance matrix $\displaystyle{\mathbf{C}_{\mathbf{\Delta}\mathbf{\Delta}}}$ can be written in terms of the identity matrix $\mathbf{I}$ and the matrix $\mathbf{1}\mathbf{1}^\mathsf{T}$,
\begin{align}\label{eq:covariance_matrix_rewritten}
\mathbf{C}_{\mathbf{\Delta}\mathbf{\Delta}} = \sigma^2 \left[\rho_{\mathbf{\Delta}} \mathbf{1}\mathbf{1}^\mathsf{T} -  (\rho_{\mathbf{\Delta}}-1)\mathbf{I}\right].
\end{align}%the covariance matrix $\displaystyle{\mathbf{C}_{\mathbf{\Delta}\mathbf{\Delta}}}$ is singular,
In view of (\ref{eq:zero-sum-constraint}), $\displaystyle{\mathbf{C}_{\mathbf{\Delta}\mathbf{\Delta}}}$ is singular
%In view of the zero-sum constraint (\ref{eq:zero-sum-constraint}), $\displaystyle{\mathbf{C}_{\mathbf{\Delta}\mathbf{\Delta}}}$ is singular
%\begin{align}
%\mathbf{1}^\mathsf{T}\mathbf{C}_{\mathbf{\Delta}\mathbf{\Delta}}\mathbf{1} &= \mathbb{E}\left[\mathbf{1}^\mathsf{T}\mathbf{\Delta}\mathbf{\Delta}^\mathsf{T}\mathbf{1}\right]=0.
%\end{align} Therefore, solving
and therefore, solving $\det(\mathbf{C}_{\mathbf{\Delta}\mathbf{\Delta}})=0$
%\begin{align}\label{eq:det}
%\det(\mathbf{C}_{\mathbf{\Delta}\mathbf{\Delta}})&=\det \left(\mathbf{1}\mathbf{1}^\mathsf{T} - \frac{\rho_{\mathbf{\Delta}}-1}{\rho_{\mathbf{\Delta}}}\mathbf{I}\right)=0
%\end{align}
gives two singularities for $\mathbf{C}_{\mathbf{\Delta}\mathbf{\Delta}}$, one at $\rho_{\mathbf{\Delta}}=1$ and another at $\rho_{\mathbf{\Delta}}=\frac{1}{1-k}$. As a consequence, the correlation coefficient  in (\ref{eq:covariance_matrix}) has the limited range~\cite{Flierl2002}
\begin{equation}\label{eq:limited_range}
\frac{1}{1-k}\leq \rho_{\mathbf{\Delta}}\leq 1
\end{equation}
which is dependent on the number of depth hypotheses $k$. In practice, the covariance matrix $\mathbf{C}_{\mathbf{\Delta}\mathbf{\Delta}}$ has to be estimated by using the loop difference vectors of all principal pixels.

By setting the characteristic polynomial of (\ref{eq:eigen_equation}) equal to zero
\begin{align}\label{eq:characteristic_polynomial}
p(\lambda) = \det(\mathbf{C}_{\mathbf{\Delta}\mathbf{\Delta}}-\lambda \mathbf{I}) = 0,
\end{align}
we obtain the following eigenvalues for $\mathbf{C}_{\mathbf{\Delta}\mathbf{\Delta}}$
\begin{align}\label{eq:eigen1}
 \lambda_{1} &= \displaystyle{\mathbf{\sigma}^{2}\left[1+(k-1)\rho_{\mathbf{\Delta}}\right]},\\ \label{eq:eigen2}
 \lambda_{2},& \lambda_{3},\ldots,\lambda_{k} = \mathbf{\sigma}^{2}(1-\rho_{\mathbf{\Delta}}),
 \end{align}where $\lambda_1$ is a non-degenerate eigenvalue and $\lambda_q$, $q=2, \ldots, k$, is a $(k-1)$-fold degenerate eigenvalue for $\rho_\Delta \neq 0$. As $\mathbf{C}_{\mathbf{\Delta}\mathbf{\Delta}}$ is singular and $\rho_\Delta = 1$ is not observed in the data, (\ref{eq:eigen1}) and (\ref{eq:eigen2}) lead to
\begin{align}\label{eq:eigen3}
\lambda_1  &=0,\\ \label{eq:eigen3b}
\lambda_l  &= \frac{k}{k-1}\sigma^2.
\end{align}These relations hold for any value of $k>1$, where $l=2,\ldots, k$.

With $\lambda_1=0$, we solve (\ref{eq:eigen_equation}) by using
\begin{align}\label{eq:eigen4}
 \mathbf{u}_{1} &= \frac{1}{\sqrt{k}}[1, 1, \ldots, 1]^\mathsf{T}\in \mathbb{R}^{k}.
\end{align}
For the remaining basis vectors $\mathbf{u}_{2},\ldots,  \mathbf{u}_{k}$ and the $(k-1)$-degenerate eigenvalue $\lambda_l$ we get
\begin{align}\label{eq:eigen5}
 \mathbf{1}^\mathsf{T}\mathbf{u}_{l} &= 0
\end{align}for $l=2,\ldots,k$. Hence, there is a $(k-1)$-dimensional linear subspace and any basis vector in this subspace is an eigenvector with the same eigenvalue. We can arbitrarily choose $k-1$ linearly independent vectors in  this subspace and orthonormalize them by the Gram-Schmidt process such that the $(k-1)$-dimensional subspace is orthogonal to $\mathbf{u}_1$.

\begin{figure}[t]
\centering
\scalebox{0.9}
{\begin{tikzpicture}
    \node (0)              [            ]{\includegraphics[width=10pc, height=7.5pc]{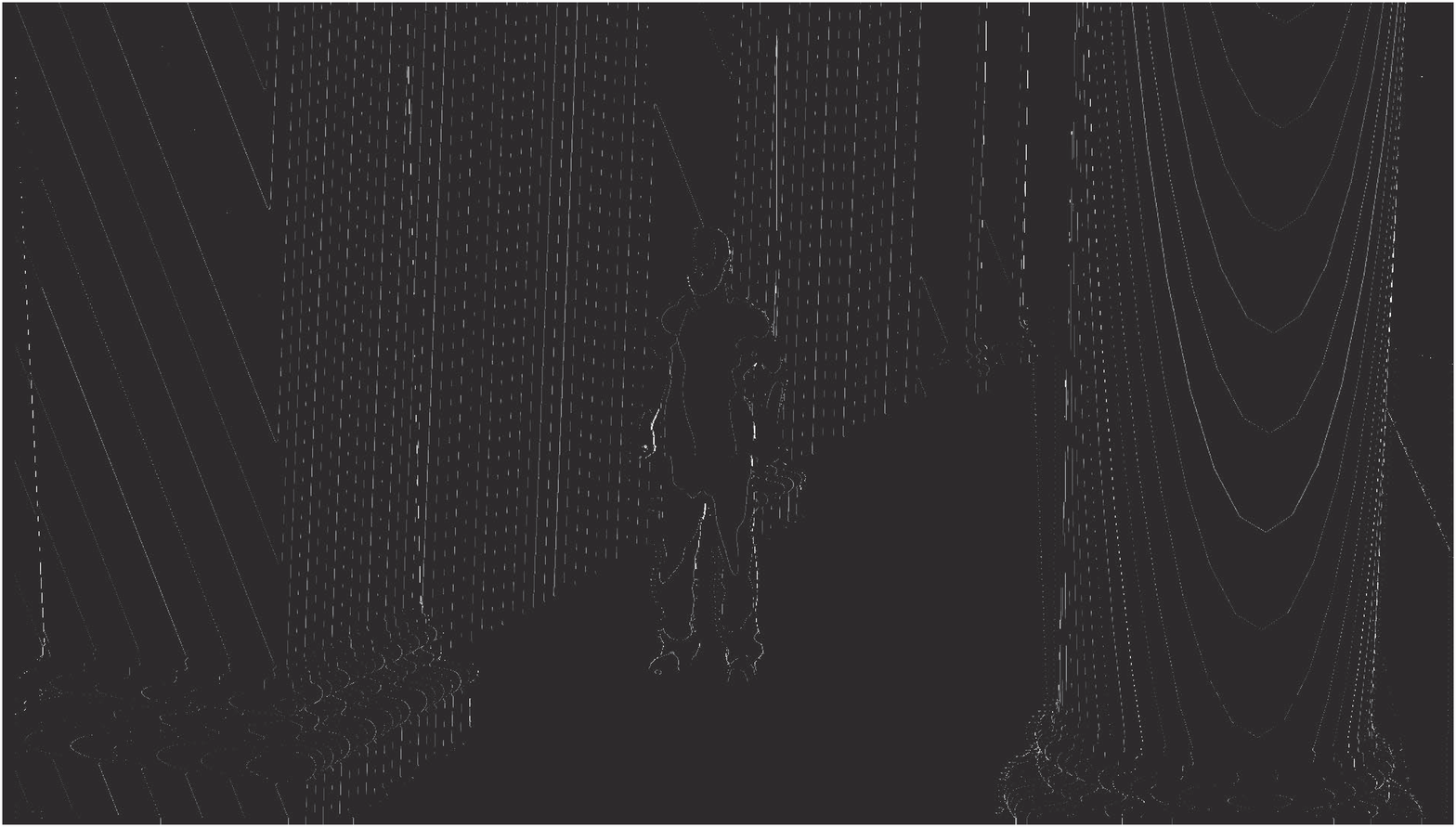}};
    \node (1) at (0.east)  [anchor=west ]{\includegraphics[width=10pc, height=7.5pc]{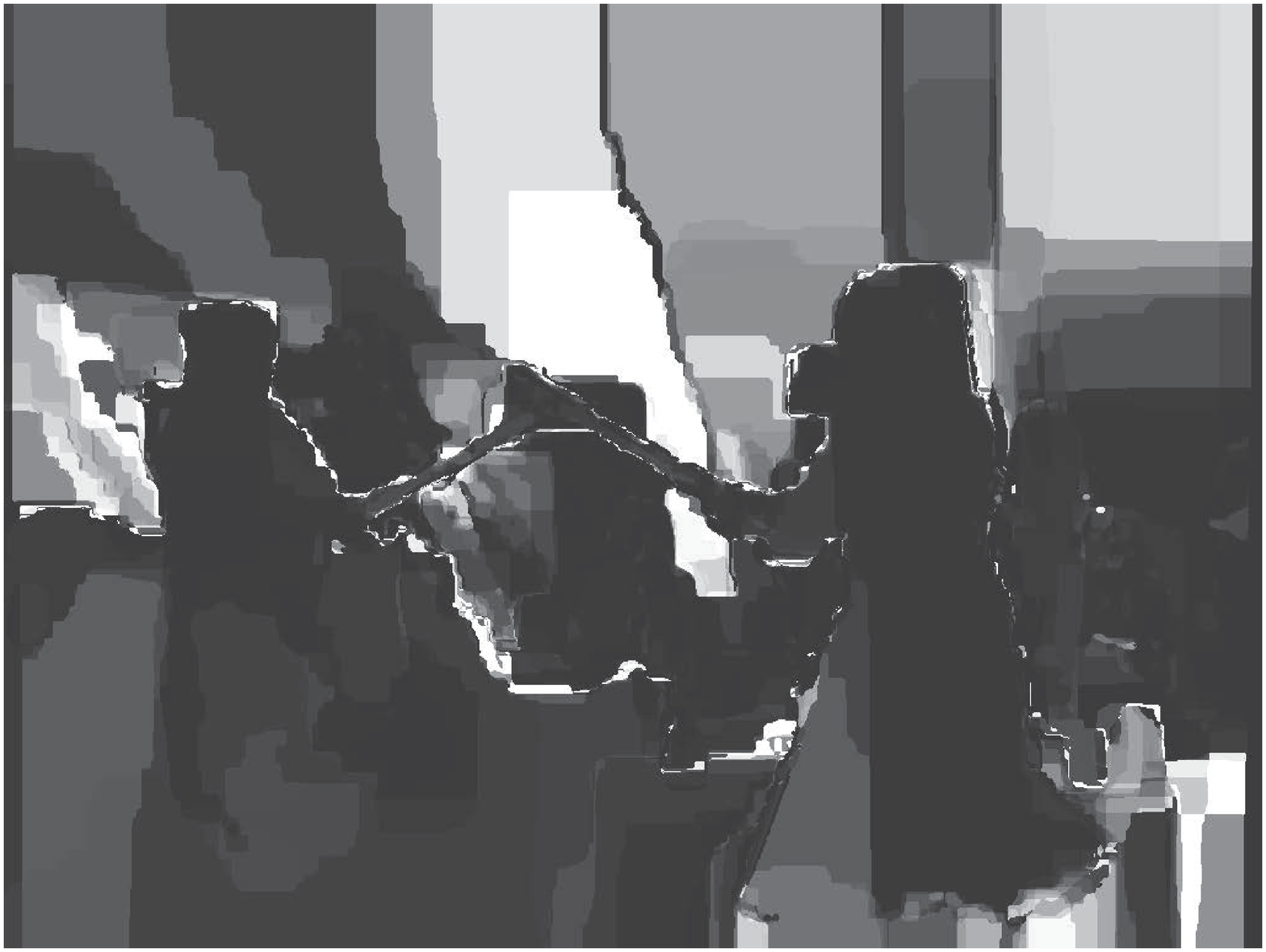}};
    \node (A) at (0.south) [anchor=north]{\footnotesize (a)~Dancer.};
    \node (B) at (1.south) [anchor=north]{\footnotesize (b)~Kendo.};
\end{tikzpicture}}
\caption{\label{fig:subspace energy_image}The energy of the loop difference $\mathrm{E}_{3}(\mathbf{\Delta})$ for all principal pixels using three depth hypotheses, i.e., $k=3$. The computer generated Dancer depth imagery is highly inter-view consistent when compared to the estimated Kendo depth imagery. Note that images are in grayscale, where pixel intensities range from 0-255. The zero pixel value represents zero loop energy, where the 255 pixel value represents the largest loop energy, i.e., the weakest inter-view consistency.}
\end{figure}

\begin{figure}[t!]
\centering
\begin{tikzpicture}
\draw [color=gray](0,0) circle [radius=1.3];
\draw [->, black] (-2,0) -- (2,0) node at (2.2,0) {\small $\Psi_{2}$};
\draw [->, black] (0,-2) -- (0,2) node at (0,2.2) {\small $\Psi_{3}$};
\node at (1.8, 1.8) {\small $\left({\Psi}_{1}=0\right)$};
\node at (0.65, -.3) {\footnotesize $\sqrt{\lambda_2}$};
\node at (.3, 0.65) {\footnotesize $\sqrt{\lambda_3}$};
\draw (-.2,1.3) -- (.2, 1.3);
\draw (1.3,-.2) -- (1.3,.2);
\end{tikzpicture}
\caption{\label{fig:constant_loop_enegery}Subspace of the transformed loop vector $\mathbf{\Delta}$ for the three reference viewpoint scenario ($k=3$). The ($k-2$)-sphere (circle) with constant loop energy $\mathrm{E}(\mathbf{\Delta})=\lambda_2=\lambda_3$ is shown.}
\end{figure}
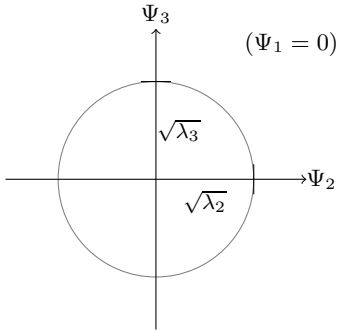

\section{Testing in Depth Difference Subspace}
\label{sec:subspace_depth consistency testing}
Inter-view depth consistency testing (IVDCT) will give us information about the inter-view consistency among estimated depth values in the multiview depth imagery. IVDCT at any principal pixel starts by defining the loop difference vector $\mathbf{\Delta}$ and using all available $k$ depth hypotheses as described in~\ref{subsec:consistency_analysis}. Second, the subspace orthogonal transform $\mathbf{U} $ is obtained from the measurements $\mathbf{\Delta}$ and the so-called loop energy $\mathrm{E}_{k}(\mathbf{\Delta})$ is calculated. Finally, we test the inter-view depth consistency with respect to an inter-view consistency threshold. This testing provides inter-view depth consistency information across multiple viewpoints. In the following, we define the loop energy of the loop difference vector $\mathbf{\Delta}$ and discuss the testing algorithm in detail.

We define the loop energy of the loop difference vector $\mathbf{\Delta}$ by the inner product of the loop difference vector.\begin{align}\label{eq:energy0}
 \mathrm{E}_{k}(\mathbf{\Delta})&= \mathbf{\Delta}^\mathsf{T} \mathbf{\Delta} = \mathbf{\Psi}^\mathsf{T} \mathbf{\Psi}
\end{align}
Here, $\Psi_1$ is zero always because  $\mathbf{u}_{1}^\mathsf{T} \mathbf{\Delta}= \mathbf{1}^\mathsf{T} \mathbf{\Delta} = 0$, whereas $\Psi_2, \ldots, \Psi_k$ are uncorrelated and Gaussian distributed. Then, the sum of their squares, i.e., $\mathrm{E}_{k}(\mathbf{\Delta})$, is distributed according to the chi-squared ($\chi_{k-1}^2$) distribution with $k-1$ degrees of freedom for any $k$-dimensional loop space. Fig.~\ref{fig:constant_loop_enegery} shows the subspace and an example of a constant loop energy for the three reference viewpoint scenario.

In the best case, which we call the \emph{zero error event}, the depth values for a visible world point across all viewpoints are perfectly consistent. Consequently, the loop difference vector for any zero error event is given by
\begin{align}
\mathbf{\Delta}_{\circ} &= [0, 0, \ldots, 0]^\mathsf{T} \in \mathbb{R}^{k}
\end{align}and the corresponding loop energy is zero, i.e., $\mathrm{E}_{k}(\mathbf{\Delta}_{\circ})= 0$ $\forall~k$, The \emph{non-zero error events}, $\displaystyle{\mathrm{E}_{k}(\mathbf{\Delta})>0}$ $\forall k$, reflect inconsistency in the observed depth values across multiple viewpoints. Therefore, the energy of the loop difference vector is directly related to the severity of the inconsistency across multiple viewpoints. Fig.~\ref{fig:subspace energy_image} shows the loop energy $\mathrm{E}_{k}(\mathbf{\Delta})$ for two different scenarios, one computer generated depth imagery and another for estimated depth imagery. Especially, we note that the computer generated depth imagery, which is highly consistent, shows mostly zero error events due to integer rounding, whereas the estimated depth imagery shows mostly non-zero error events. In the following, we use the energy of the loop difference vector as a measure of depth inconsistency.

Let us first define the inter-view consistency information $I_{k}$ as a binary information which specifies whether $k$ depth hypotheses, $\hat{d}_{p}(1;x,y), \ldots,\hat{d}_{p}(k;x,y)$, at any principal pixel are consistent $(1)$ or inconsistent $(0)$. The inter-view consistency information $I_{k}$ is obtained by checking the corresponding loop energy  $\mathrm{E}_{k}(\mathbf{\Delta})$,\begin{equation}\label{eq:inter-view_consistency info}
I_{k} = \left\{
\begin{array}{cc}
  1 &  \text{if $\mathrm{E}_{k}(\mathbf{\Delta}) \leq \vartheta $,} \\
  0 & \text{otherwise,}
\end{array}
\right.
\end{equation}where $\vartheta$ is an \emph{inter-view consistency threshold} which is defined as the energy of a loop difference vector at a principal pixel having the desired quality of inter-view consistency. In order to relate $\vartheta$ to the desired quality of inter-view consistency, we define
\begin{align}\label{eq:inter-view consistency_threshold_define}
\vartheta & := \alpha^2 \lambda_2,
\end{align}where $\lambda_2$ is the smallest non-zero eigenvalue and where the weight $\alpha \in [0, 1]$ is empirically chosen. By using the relation (\ref{eq:eigen3b}), we get  \begin{align}\label{eq:inter-view consistency_threshold_define}
\vartheta & = \alpha^2 \frac{k}{k-1}\sigma^2,
\end{align}where $\frac{k}{k-1}$ helps us to adjust the consistency threshold to any given $k$-dimensional depth difference space. Further, with the variance ${\sigma}^{2}$, we adjust to the variance of the depth inconsistency evidence $\Delta_{ij}$. Note that if vector $\mathbf{\Delta}$ is wide-sense stationary $k$-variate normal distributed then this statistical inter-view consistency test falls into the category of  the chi-squared test as $\mathrm{E}_{k}(\mathbf{\Delta}) \sim \chi_{k-1}^2$.

For any non-zero error event, where $\mathrm{E}_{k}(\mathbf{\Delta}) \leq \vartheta$, we accept all $k$ depth hypotheses and consider them as sufficiently consistent. Subsequently, we assume that all the corresponding $k$ depth pixels have a consistent depth representation and describe the same 3D object point in world coordinates. Using this consistency information and the perspective projection as described in \ref{subsec:multiple depth warping}, the corresponding depth values in the reference depth maps can be used to determine an improved depth estimate for the given principal pixel. Moreover, if $\mathrm{E}_{k}(\mathbf{\Delta}) = 0$, i.e., the zero error event, all the corresponding $k$ depth hypotheses are assumed to be perfectly consistent. Finally, if $\mathrm{E}_{k}(\mathbf{\Delta}) > \vartheta$, we reject all $k$  depth hypotheses and assume that we do not have a consistent depth representation at the given principal pixel. We refer to such events as \emph{extreme error events}.

\begin{figure}[t!]
\centering
    \begin{tikzpicture}
    %\draw [help lines,step=0.5cm] grid (7,5);
     %second layer
    \fill (00.00,04.00) circle (2pt) node (A) [below left=-2pt ] {\footnotesize $\hat{d}_1$};
    \fill (01.00,04.00) circle (2pt) node (D) [below right=-2pt] {\footnotesize $\hat{d}_2$};
    \fill (01.00,05.00) circle (2pt) node (C) [above right=-2pt] {\footnotesize $\hat{d}_3$};
    \fill (00.00,05.00) circle (2pt) node (B) [above left=-2pt ] {\footnotesize $\hat{d}_4$};
    \draw [thick] (00.00,05.00) -- (01.00,05.00) node[pos=0.5, above] {\footnotesize $-\mu$};
    \draw [thick] (01.00,04.00) -- (01.00,05.00) node[pos=0.5, right] {\footnotesize $\mu$};
    \draw [thick] (01.00,04.00) -- (00.00,05.00) node[pos=0.5, sloped, below] {\footnotesize $0$};
    %\draw [thick, dashed] (01.00,04.00) -- (00.00,04.00) {};
%    \draw [thick, dashed] (00.00,04.00) -- (00.00,05.00) {};

    %\node at (00.50, 03.25) [above] {\footnotesize (a)};

    \fill (02.00,04.00) circle (2pt) node (A) [below left=-2pt] {\footnotesize $\hat{d}_1$};
    \fill (03.00,04.00) circle (2pt) node (D) [below right=-2pt] {\footnotesize $\hat{d}_2$};
    \fill (03.00,05.00) circle (2pt) node (C) [above right=-2pt] {\footnotesize $\hat{d}_3$};
    \fill (02.00,05.00) circle (2pt) node (B) [above left=-2pt] {\footnotesize $\hat{d}_4$};
    \draw [thick] (02.00,04.00) -- (02.00,05.00) node[pos=0.5, left ] {\footnotesize $0$};
    \draw [thick] (02.00,05.00) -- (03.00,05.00) node[pos=0.5, above] {\footnotesize $-\mu$};
    \draw [thick] (03.00,05.00) -- (02.00,04.00) node[pos=0.5, sloped, below] {\footnotesize $\mu$};
    %\draw [thick, dashed] (02.00,04.00) -- (03.00,04.00) {};
%    \draw [thick, dashed] (03.00,04.00) -- (03.00,05.00) {};
    %\node at (02.50, 03.25) [above] {\footnotesize (b)};

    \fill (04.00,04.00) circle (2pt) node (A) [below left=-2pt ] {\footnotesize $\hat{d}_1$};
    \fill (05.00,04.00) circle (2pt) node (D) [below right=-2pt] {\footnotesize $\hat{d}_2$};
    \fill (05.00,05.00) circle (2pt) node (C) [above right=-2pt] {\footnotesize $\hat{d}_3$};
    \fill (04.00,05.00) circle (2pt) node (B) [above left=-2pt] {\footnotesize $\hat{d}_4$};
    \draw [thick] (04.00,04.00) -- (04.00,05.00) node[pos=0.5, left ] {\footnotesize $\mu$};
    \draw [thick] (04.00,05.00) -- (05.00,04.00) node[pos=0.5, sloped, above] {\footnotesize $0$};
    \draw [thick] (05.00,04.00) -- (04.00,04.00) node[pos=0.5, below] {\footnotesize $-\mu$};
   % \draw [thick, dashed] (05.00,04.00) -- (05.00,05.00) {};
%    \draw [thick, dashed] (05.00,05.00) -- (04.00,05.00) {};
    %\node at (04.50, 03.25) [above] {\footnotesize (c)};

    \fill (06.00,04.00) circle (2pt) node (A) [below left=-2pt] {\footnotesize $\hat{d}_1$};
    \fill (07.00,04.00) circle (2pt) node (D) [below right=-2pt] {\footnotesize $\hat{d}_2$};
    \fill (07.00,05.00) circle (2pt) node (C) [above right=-2pt] {\footnotesize $\hat{d}_3$};
    \fill (06.00,05.00) circle (2pt) node (B) [above left=-2pt] {\footnotesize $\hat{d}_4$};
    \draw [thick] (06.00,04.00) -- (07.00,04.00) node[pos=0.5, below] {\footnotesize $-\mu$};
    \draw [thick] (07.00,04.00) -- (07.00,05.00) node[pos=0.5, right] {\footnotesize $\mu$};
    \draw [thick] (07.00,05.00) -- (06.00,04.00) node[pos=0.5, sloped, above] {\footnotesize $0$};
   % \draw [thick, dashed] (06.00,04.00) -- (06.00,05.00) {};
%    \draw [thick, dashed] (06.00,05.00) -- (07.00,05.00) {};
    %\node at (06.50, 03.25) [above] {\footnotesize (d)};
\end{tikzpicture}
\caption{\label{fig:single error events}All possible cases of single error events for the scenario where we select three out of four depth hypotheses.}
\end{figure}
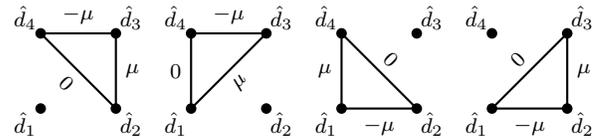

If the consistency test fails with available $k$ depth hypotheses, we repeat the consistency analysis and testing with $k-1$ out of $k$ available depth hypotheses. However, there are
\begin{align}
{}^{k}\mathsf{C}_{k-1}&= \frac{k!}{(k-1)!(k-(k-1))!} = k
\end{align} ways to select $k-1$ out of $k$ available depth hypotheses and to define the corresponding $k$ unique loop difference vectors of $k-1$ dimension.  Fig.~\ref{fig:single error events} shows examples of four possible combinations to define loop difference vectors for the single error event using three out of four available depth hypotheses.  We therefore perform $k$ consistency analyses and tests with $k$ different loop difference vectors. If multiple consistency tests out of k are successful, we only accept the test which satisfies $\vartheta$ with the smallest loop. If all $k$ consistency tests with $k-1$ depth hypotheses fail, we repeat the process of consistency analysis and testing with a reduced number of depth hypotheses for each possible combination of available depth hypotheses until the smallest possible number $k=2$ is reached. When all tests failed, we mark the corresponding principal pixel by a mask which allows other techniques to improve the current depth value. For example, we may use inpainting ~\cite{Bertalmio2001} in such cases.

In a nutshell, the fundamental approach of this work is rooted in the combination of the zero-sum constraint~(\ref{eq:zero-sum-constraint}) with the threshold constraint (\ref{eq:inter-view_consistency info}).  We measure the differences between related depth values and use them as evidence. Due to (\ref{eq:zero-sum-constraint}),  the covariance matrix of the vector of evidence values, i.e., loop difference vector, is singular. With the threshold constraint (i.e. a constraint on the variance), we are able to find a subspace of the evidence in which the zero-sum constraint is satisfied at a lower variance. We use this approach to find consistent evidence at several threshold levels according to (\ref{eq:inter-view consistency_threshold_define}). In other words, by removing outliers, we find subspaces of the evidence that satisfy the zero-sum constraint at various levels of the variance of the evidence. The algorithm is summarized in Fig.~\ref{fig:depth_consistency_testing_algo}.

\begin{figure}[t!]
 \begin{center}
 \scalebox{0.98}{
  \begin{tikzpicture}[node distance=3.1em, auto, >=stealth]
   \node        (0)  [          ] {\footnotesize Multiview Depth Imagery};
   \node[block] (1)  [below of=0] {\footnotesize Multiple Depth Warping};
   \node[block] (2)  [below of=1] {\footnotesize Depth Hypothesis Selection};
   \node[block] (X)  [below of=2] {\footnotesize Loop Difference Vector};
   \node[block] (3)  [below of=X] {\footnotesize Mapping into Subspace};
   \node[block] (4)  [below of=3, node distance=3.2em] {\footnotesize Inter-view Consistency Testing};
   \node        (5)  [below of=4, node distance=3.1em ] {\footnotesize Inter-view Consistency Information};
   \draw[->, thick] (0.south) -- (1.north);
   \draw[->, thick] (1.south) -- (2.north);
   \draw[->, thick] (2.south) -- (X.north);
   \draw[->, thick] (X.south) -- (3.north);
   \draw[->, thick] (3.south) -- (4.north);
   \draw[->, thick] (4.south) -- (5.north) node[pos=0.5, align=center, right]{\footnotesize Test: Succeed};
   \draw[->, thick, rounded corners] (1.east) -- +(+.7,0) |- (4.east);
   \draw[->, thick, rounded corners] (4.west) -- +(-.7,0) node[below]{\footnotesize Test: Fail} |- (2.west);
   \draw[rounded corners] (2.west) -- +(-.7,0) node[above,  align=center, text width= 1.2cm]{\footnotesize Dimension Reduction} |- (4.west);
   \end{tikzpicture}}
  \caption{\label{fig:depth_consistency_testing_algo}Inter-view depth consistency testing in depth difference subspace.}
 \end{center}
\end{figure}
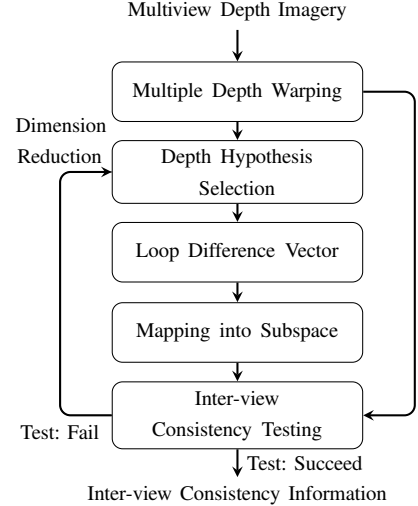

%%OK
\section{Applications}
\label{sec:applications}

The resulting consistency information is advantageous for improving many aspects of FTV. Consistent descriptions of the scene geometry can be obtained. Further, the visual quality of synthesized virtual views can be improved by exploiting the consistency information.
\subsection{Multiview Depth Image Enhancement}
\label{subsec:multiview depth imagery enhancement}
To have a consistent depth representation across $k$ viewpoints, we first utilize the IVDCT to obtain inter-view consistency information at a principal viewpoint which coincides with one of the reference viewpoints, i.e., $p = i$, where $i=1, \ldots, k$. Next, the resulting consistency information at $i$ is used to update the principal depth pixel. By updating, we mean that we replace the previous depth pixel value at $i$ by a new improved value at $i$.  Usually, the improved depth value at $i$ is determined by averaging the chosen depth hypothesis values as per consistency information out of available $k$ warped depth hypotheses.  However, if the reference viewpoints are irregularly spaced, the depth value at $i$ is updated by weighted-baseline averaging of the chosen depth hypothesis values. The enhanced depth values are then used to update the corresponding depth map value in the viewpoint $i$. We apply a similar procedure to update the depth maps at each viewpoint. The resulting depth maps show improved inter-view consistency across the $k$ viewpoints. The resulting depth pixel value at $i$ is then used as an improved input when testing the next viewpoint. We repeat this process until each viewpoint satisfies our stopping criterion which is the  thresholded relative difference between the loop energies of successive iterations. When the given threshold is achieved, the iterations are stopped. This iterative algorithm improves, the inter-view depth consistency across all viewpoints. Fig.~\ref{fig:multiview depth imagery enhancement} shows an example of enhanced depth maps as obtained by the proposed algorithm.
\begin{figure}[t!]
\begin{center}
\graphicspath{{./imgs/sdct_depthmaps/}}
\centering
\begin{tikzpicture}
    \node (0) at (1,   9)  [align=center             ]{\includegraphics[width=6.5pc, height=4.9pc]{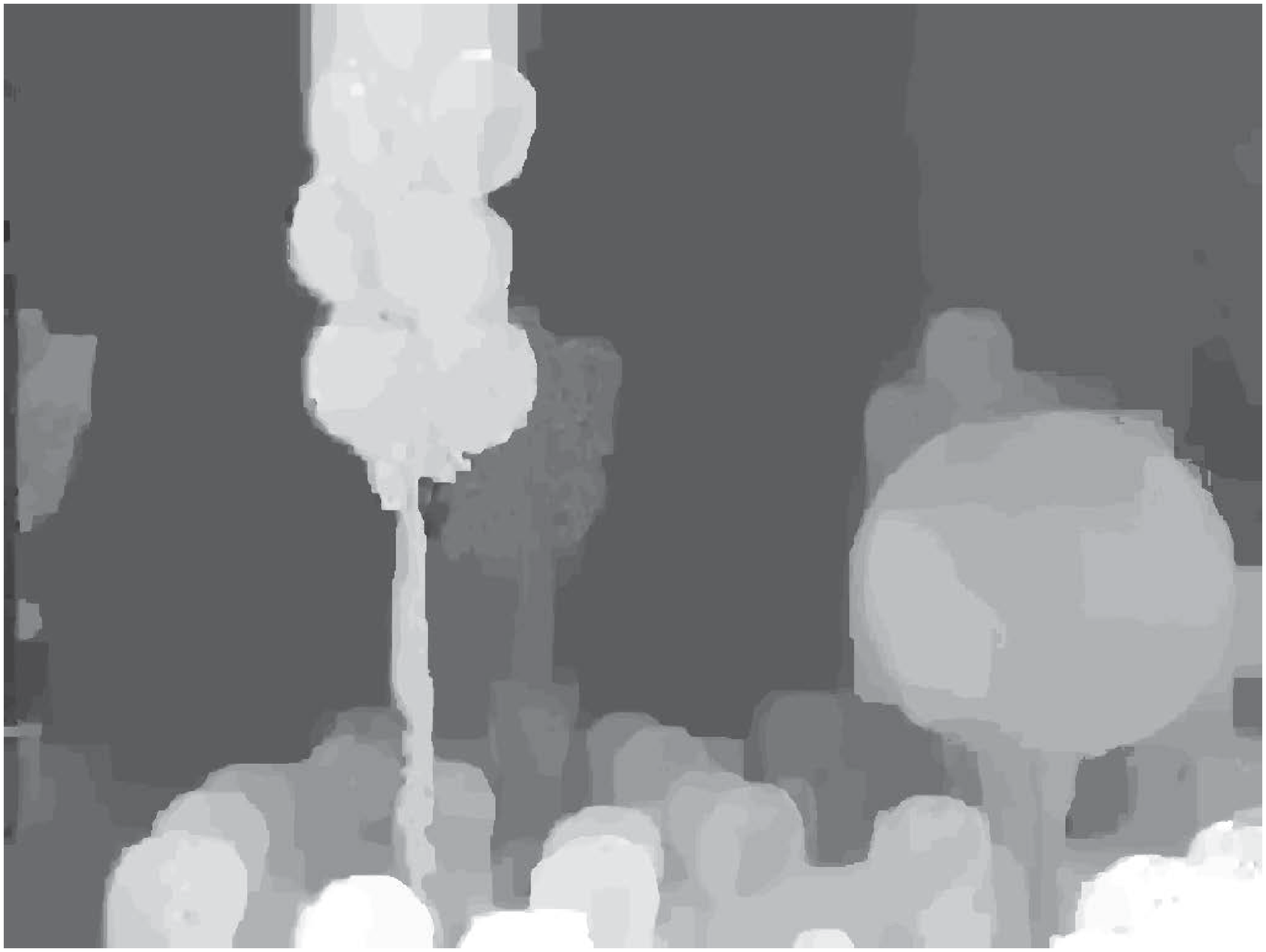}};
    \node (1) at (0.east)  [anchor=west, align=center]{\includegraphics[width=6.5pc, height=4.9pc]{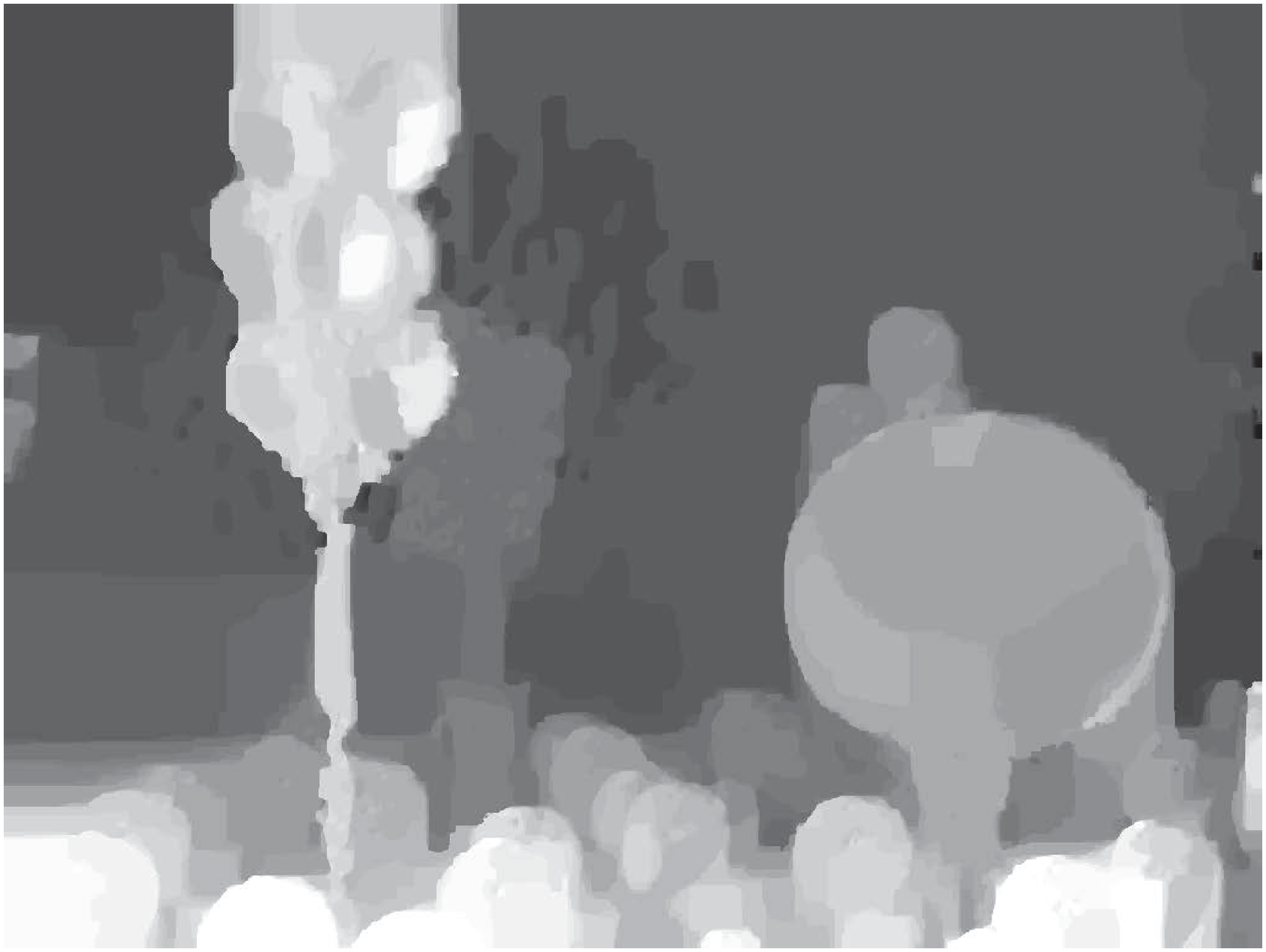}};
    \node (2) at (1.east)  [anchor=west, align=center]{\includegraphics[width=6.5pc, height=4.9pc]{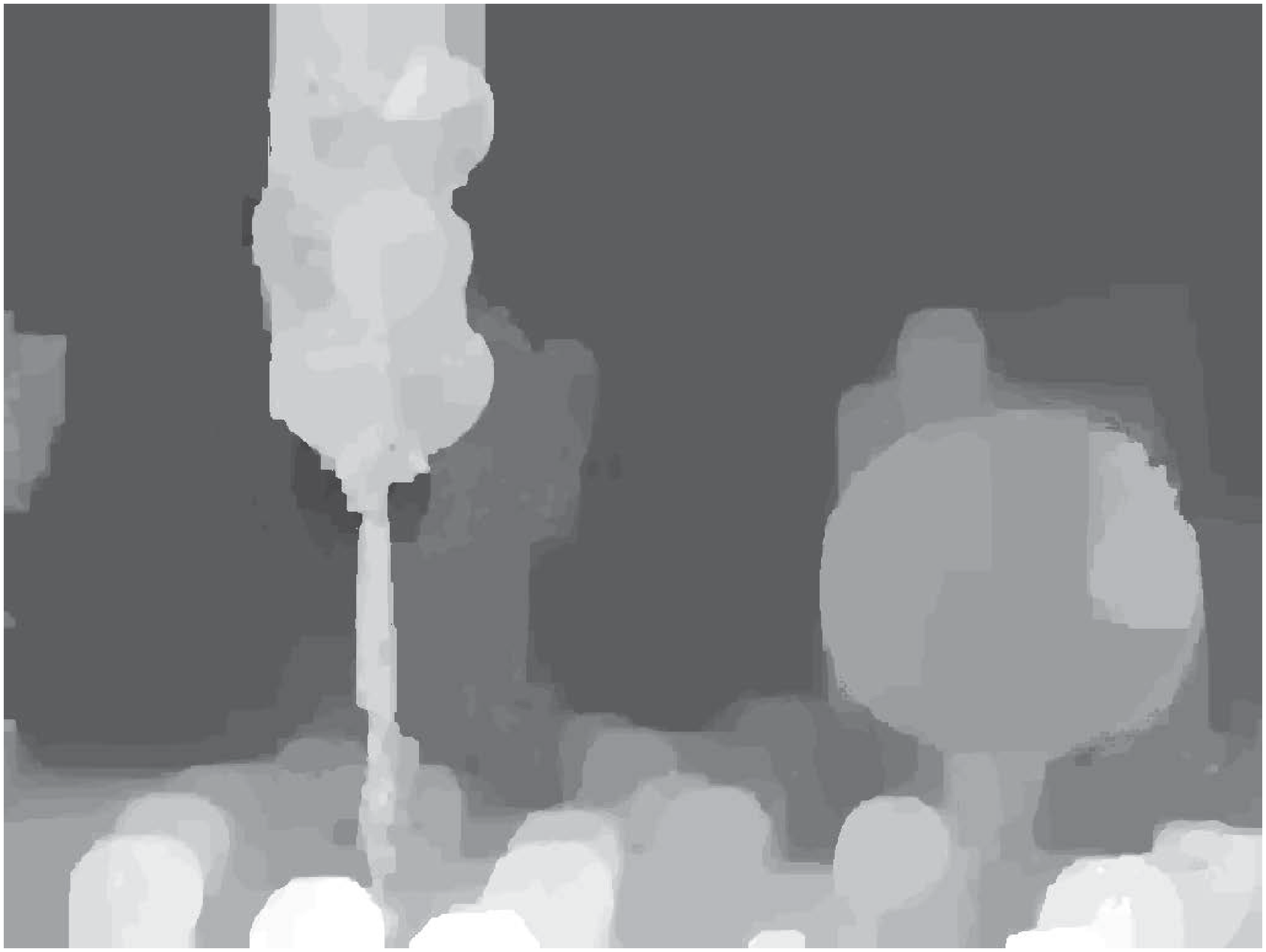}};
    \node (3) at (01.90,8.70) [circle, minimum size=0.85cm, line width=.1pc, draw, red, align=center]{};
    \node (4) at (04.82,8.70) [circle, minimum size=0.85cm, line width=.1pc, draw, red, align=center]{};
    \node (5) at (07.85,8.70) [circle, minimum size=0.85cm, line width=.1pc, draw, red, align=center]{};
    \node (6) at (01.00,9.40) [circle, minimum size=0.85cm, line width=.1pc, draw, red, align=center]{};
    \node (7) at (03.98,9.40) [circle, minimum size=0.85cm, line width=.1pc, draw, red, align=center]{};
    \node (8) at (07.01,9.40) [circle, minimum size=0.85cm, line width=.1pc, draw, red, align=center]{};
    \draw [line width=.1pc, red][->] (3.east) -- (4.west);
    \draw [line width=.1pc, red][->] (4.east) -- (5.west);
    \draw [line width=.1pc, red][->] (6.east) -- (7.west);
    \draw [line width=.1pc, red][->] (7.east) -- (8.west);
    \node    [anchor=north] at (0.south) {\footnotesize view $\sharp 1$};
    \node (9)[anchor=north] at (1.south) {\footnotesize view $\sharp 3$};
    \node    [anchor=north] at (2.south) {\footnotesize view $\sharp 5$};
    \node (10) at (9.south) [anchor=north, align=center]{\footnotesize {(a) Depth maps of the Balloons test sequence before enhancement.}};
\end{tikzpicture}
\begin{tikzpicture}
    \node (0) at (1 ,9)    [align=center             ]{\includegraphics[width=6.5pc, height=4.9pc]{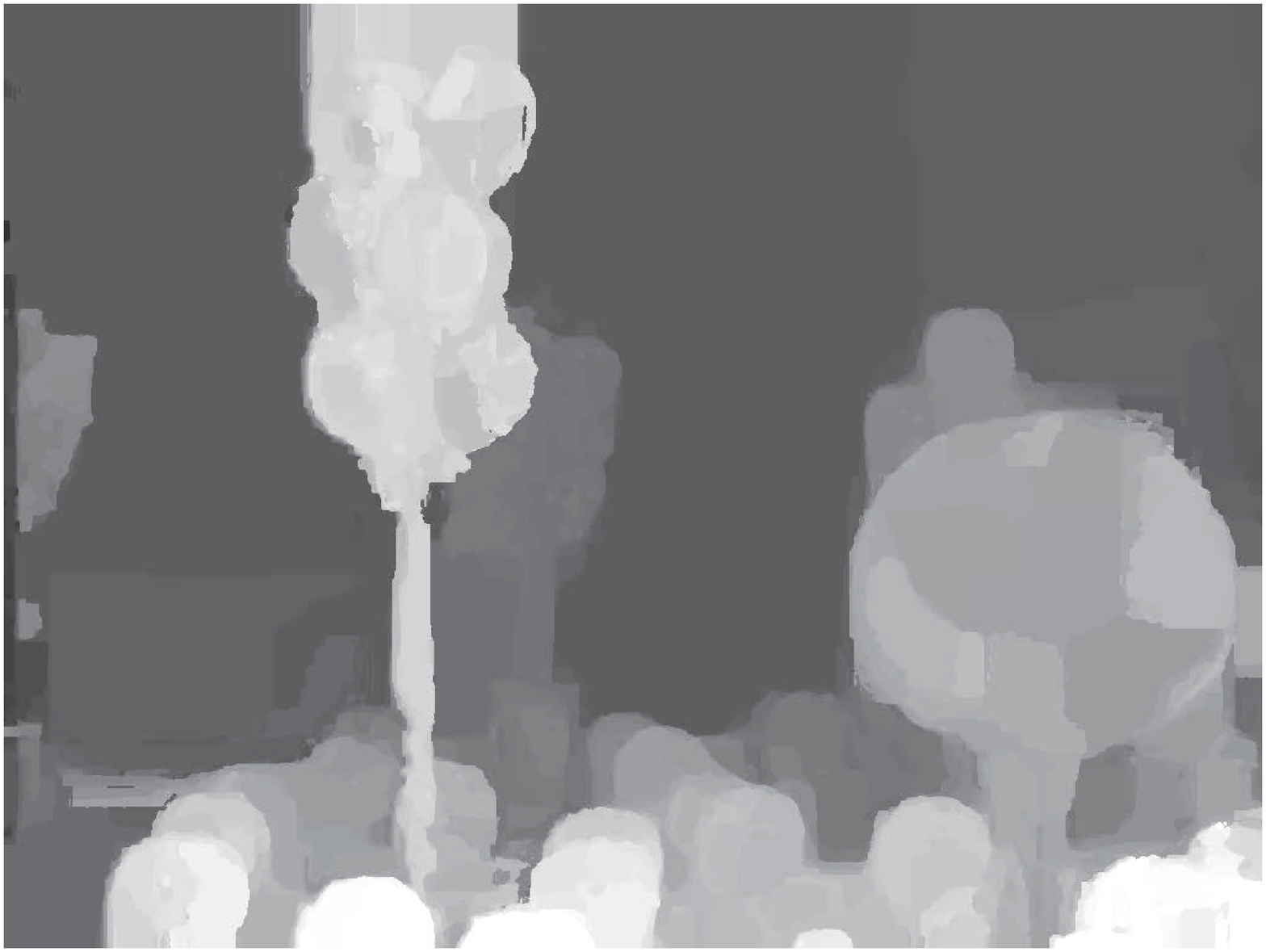}};
    \node (1) at (0.east)  [anchor=west, align=center]{\includegraphics[width=6.5pc, height=4.9pc]{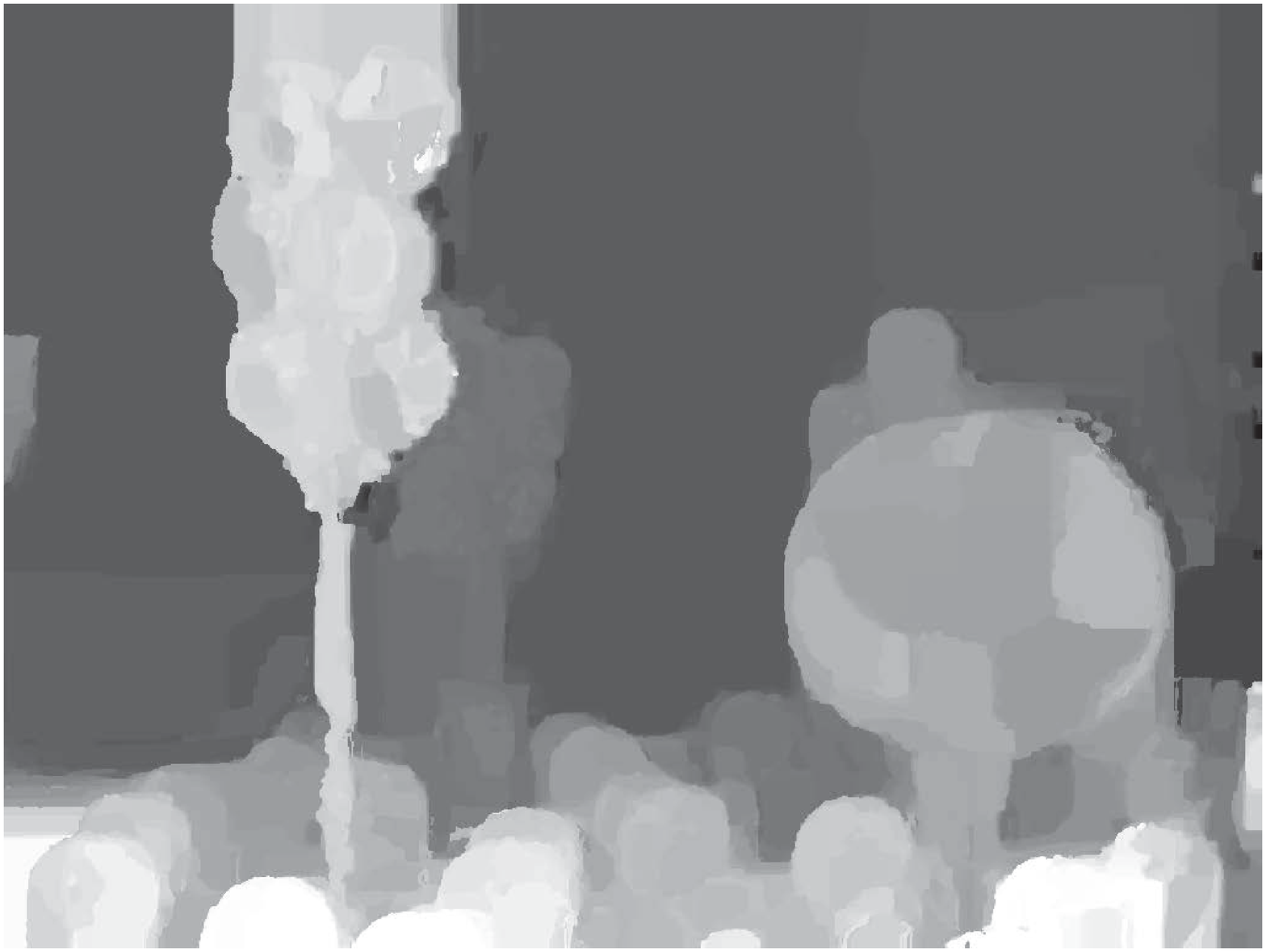}};
    \node (2) at (1.east)  [anchor=west, align=center]{\includegraphics[width=6.5pc, height=4.9pc]{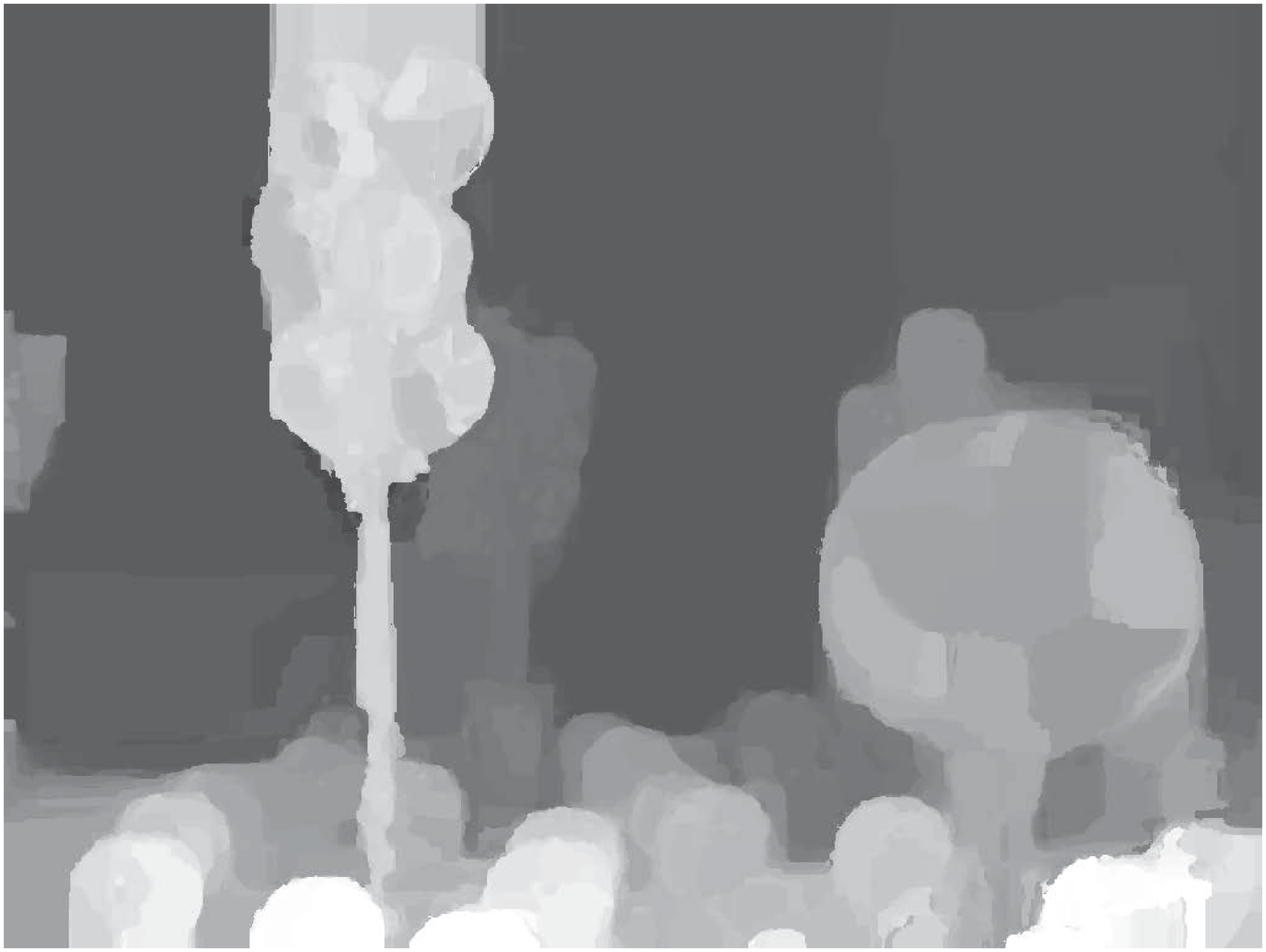}};
    \node (3) at (01.90,8.70) [circle, minimum size=0.85cm, line width=.1pc, draw, green, align=center]{};
    \node (4) at (04.82,8.70) [circle, minimum size=0.85cm, line width=.1pc, draw, green, align=center]{};
    \node (5) at (07.85,8.70) [circle, minimum size=0.85cm, line width=.1pc, draw, green, align=center]{};
    \node (6) at (01.00,9.40) [circle, minimum size=0.85cm, line width=.1pc, draw, green, align=center]{};
    \node (7) at (03.98,9.40) [circle, minimum size=0.85cm, line width=.1pc, draw, green, align=center]{};
    \node (8) at (07.01,9.40) [circle, minimum size=0.85cm, line width=.1pc, draw, green, align=center]{};
    \draw [line width=.1pc, green][->] (3.east) -- (4.west);
    \draw [line width=.1pc, green][->] (4.east) -- (5.west);
    \draw [line width=.1pc, green][->] (6.east) -- (7.west);
    \draw [line width=.1pc, green][->] (7.east) -- (8.west);
    \node    [anchor=north] at (0.south) {\footnotesize view $\sharp 1$};
    \node (9)[anchor=north] at (1.south) {\footnotesize view $\sharp 3$};
    \node    [anchor=north] at (2.south) {\footnotesize view $\sharp 5$};
    \node (10) at (9.south) [anchor=north, align=center]{\footnotesize{(b) Depth maps of the Balloons test sequence after iterative enhancement.}};
\end{tikzpicture}
\caption{\label{fig:multiview depth imagery enhancement} For 1D-parallel camera arrangements, the depth value of a unique 3D point is the same in all depth maps, but located at different positions in the maps. Therefore, depth observations at different viewpoints should be consistent and related areas in different viewpoints should show the same depth values, but shifted. This is not always the case in (a). After the enhancement in (b), more regions are consistent. The red circles mark areas with inter-view inconsistency in the depth maps before enhancement. The green circles mark corresponding areas with improved inter-view consistency  after enhancement.}
\end{center}
\end{figure}
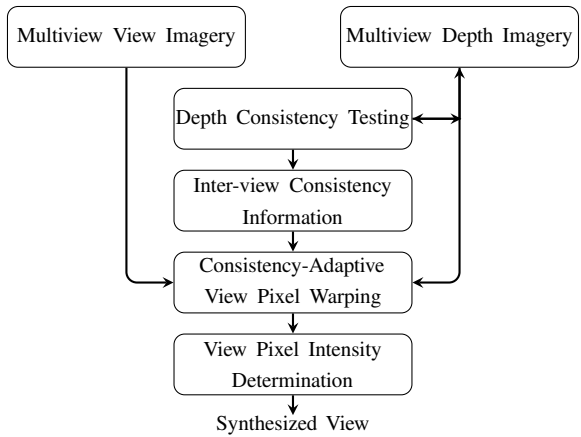
\begin{figure}[t!]
 \centering
  \begin{tikzpicture}[node distance=6.5em, auto, >=stealth, inner sep=0pt,remember picture]
    \node[block] (2)  [                                ] {\footnotesize Multiview Depth Imagery};
    \node[block] (3)  [left of =2, node distance=12.6em] {\footnotesize Multiview View Imagery};
    \node        (4)  [left of =2, node distance=06.3em] {};
    \node[block] (5)  [below of=4, node distance=03.1em] {\footnotesize Depth Consistency Testing};
    \node[block] (6)  [below of=5, node distance=03.1em] {\footnotesize Inter-view Consistency Information};
    \node[block] (7)  [below of=6, node distance=03.1em] {\footnotesize Consistency-Adaptive\\View Pixel Warping};
    \node[block] (8)  [below of=7, node distance=03.1em] {\footnotesize View Pixel Intensity Determination};
    \node        (9)  [below of=8, node distance=02.3em] {\footnotesize Synthesized View};
    \node       (10)  [right of=5, node distance=6.3em]{};
    \draw[->, thick] (5.south) -- (6.north);
    \draw[->, thick] (6.south) -- (7.north);
    \draw[->, thick] (7.south) -- (8.north);
    \draw[->, thick] (2.south) -- +(0,0) |- (5.east);
    \draw[<->,thick,rounded corners] (2.south) -- +(0,-.2)|- (7.east);
    \draw[->, thick,rounded corners] (3.south) -- +(0,-.2)|- (7.west);
    \draw[->, thick] (8.south) -- (9.north);
    \draw[->, thick] (5.east)  -- (10.east);
   \end{tikzpicture}
 \caption{\label{fig:cavs_block}Consistency-adaptive view synthesis.}
\end{figure}

\begin{table*}[t!]
\begin{center}

\caption{\label{tab:results}Objective quality of the synthesized virtual views}% for depth maps enhanced by the proposed algorithms and }
\resizebox{\textwidth}{!}{\begin{tabular}{|@{}c@{}|@{}c@{}|@{}c@{}|@{}c@{}|c|c|c|c|c|c|c|c|c|c|} \hline
Test         & \multicolumn{2}{c|}{Input Views} &~Virtual~~& No. of   &\multicolumn{3}{c|}{MPEG/D$\to$VSRS 3.5 } &\multicolumn{3}{c|}{IVDCT/ED$\to$VSRS 3.5} &\multicolumn{3}{c|}{MPEG/D$\to$CAVS }       \\\cline{2-3}\cline{6-14}
Sequence     &~VSRS~     & ~IVDCT~                & Views   & Frames      & PSNR [dB] & SSIM  &IW-SSIM& PSNR [dB]  & SSIM  &IW-SSIM& PSNR [dB] & SSIM &IW-SSIM\\\hline \hline
Dancer       &2-5      & 2-5-9                  & 3       & 250      &  38.8     & 0.977    &0.996 &     38.8   & 0.977   &0.996 &   40.0& 0.985 &0.9982\\ \hline %$\lambda = 0.5$
Kendo        &3-5      & 1-3-5                  & 4       & 300      &  37.6       & 0.969  &0.988&     38.2      & 0.971  & &38.3   &  0.971& 0.9887\\ \hline %38.17
Balloons     &3-5      & 1-3-5                  & 4       & 300      &  36.6        &0.965  &  0.9873 &   36.8       & 0.966 & 0.9880 &37.0     & 0.969 & 0.9883 \\ \hline %36.80
Lovebird1    &6-8      & 4-6-8                  & 7       & 240      &  29.0        & 0.883 & 0.955  &  29.2       & 0.887 & &29.2    & 0.887 &0.9576\\ \hline %29.16
~Newspaper~  &4-6      & 2-4-6                  & 5       & 300      &  32.3     &  0.943   &0.981 &    33.0     &  0.945  &0.9818 & 33.5& 0.952& 0.9824\\ \hline %33.48  $\lambda = 0.25$
\end{tabular}}
\end{center}
\end{table*}

\begin{figure*}[t]
\centering
\begin{tikzpicture}
    %dancer
    \node (A) [                                 align=center] {\includegraphics[width=07.5pc, height=7.5pc]{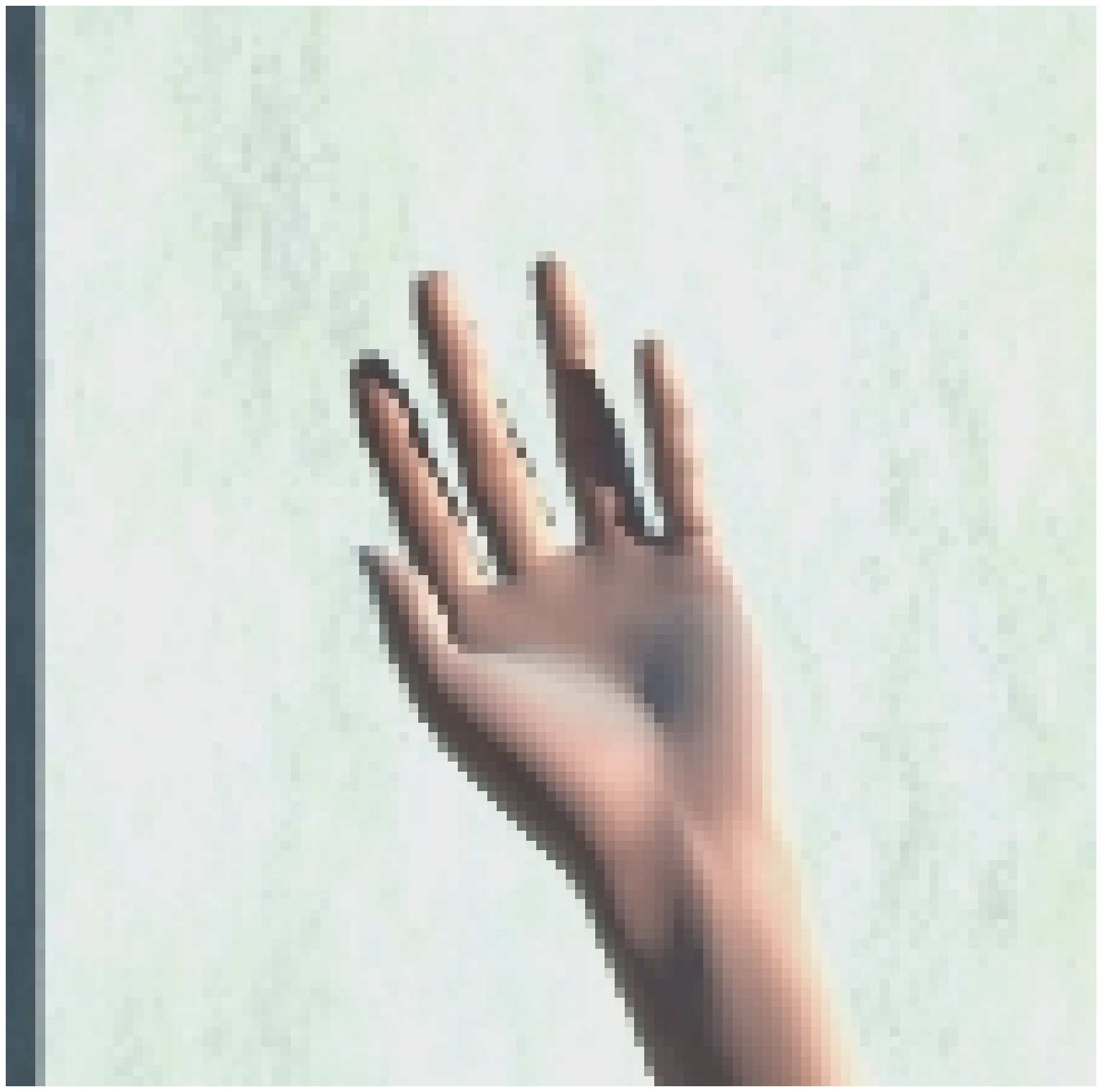}};
    \node (B) [below of =A,node distance=9.85pc,align=center] {\includegraphics[width=07.5pc, height=7.5pc]{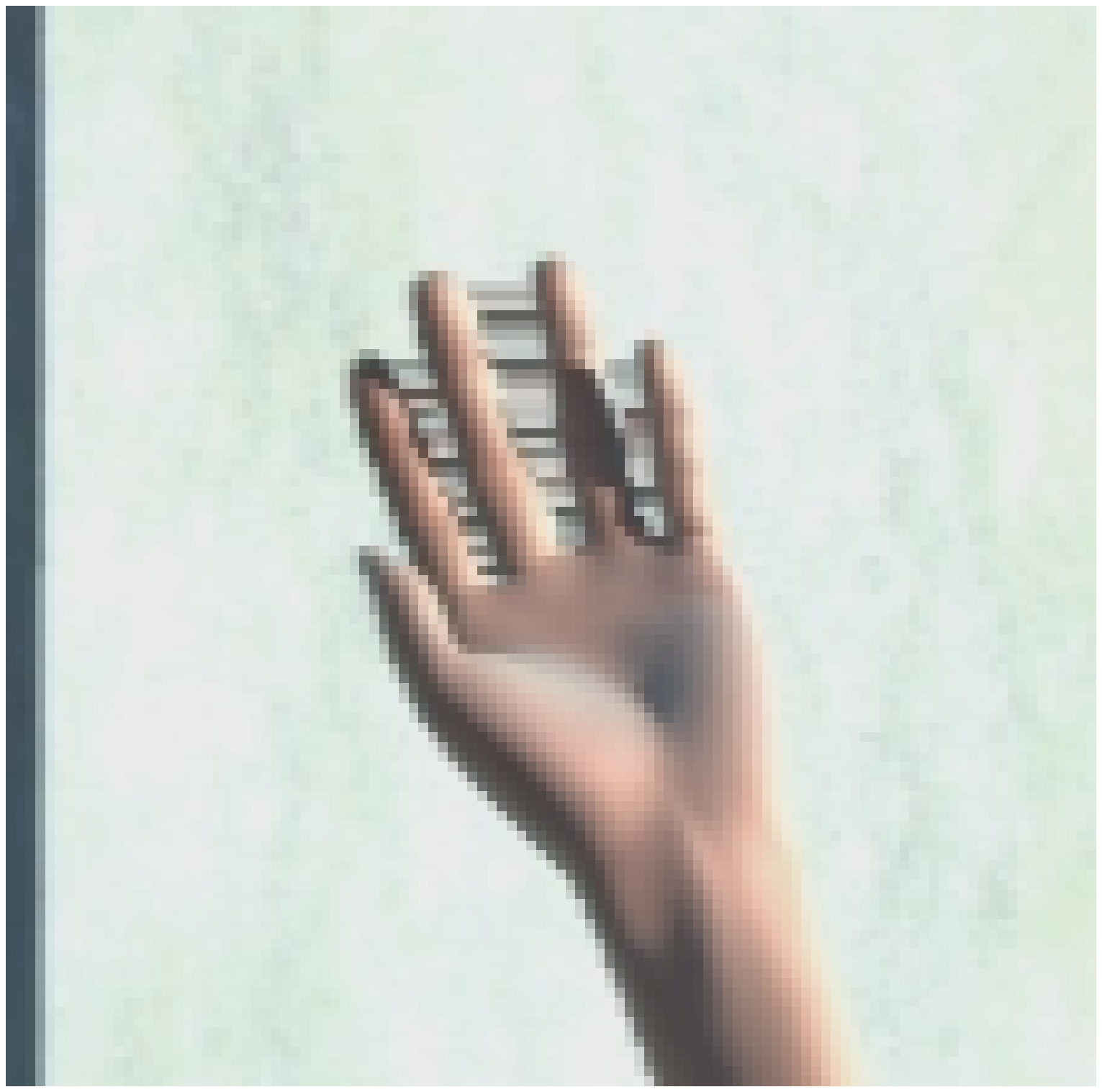}};
    \node (C) [below of =B,node distance=9.85pc,align=center] {\includegraphics[width=07.5pc, height=7.5pc]{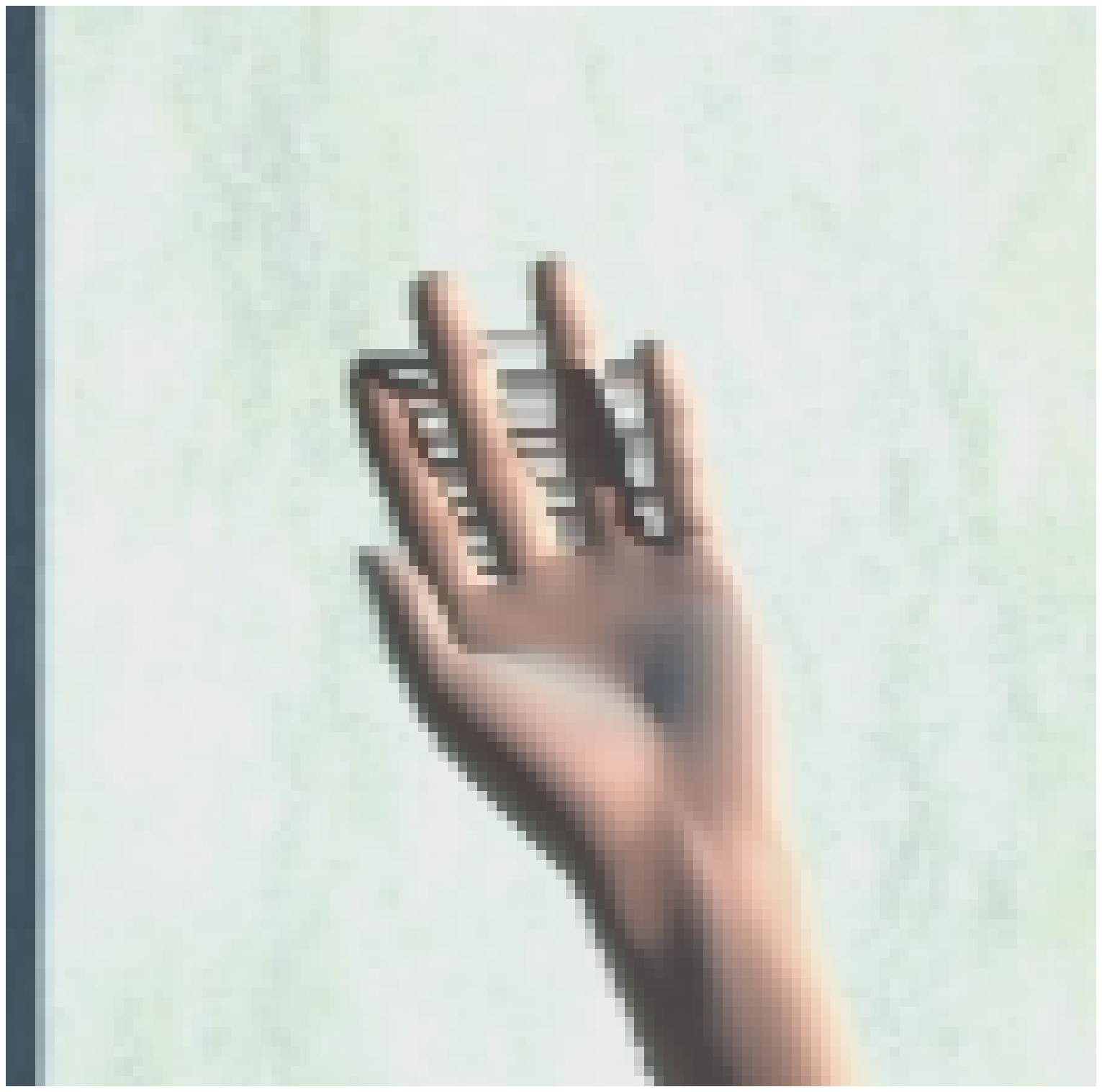}};
    \node (D) [below of =C,node distance=9.85pc,align=center] {\includegraphics[width=07.5pc, height=7.5pc]{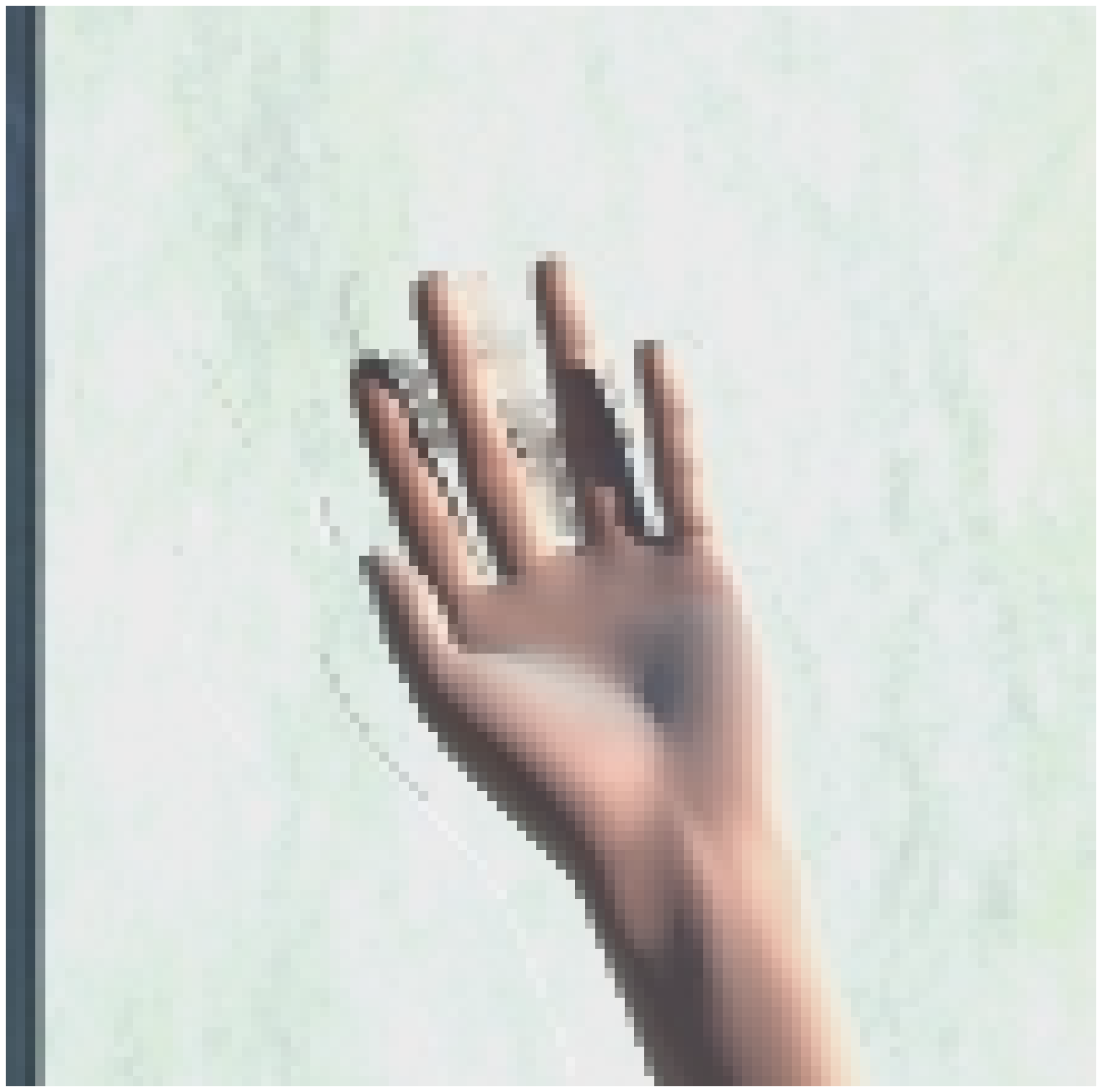}};
    \node (01) at (A) [ellipse, minimum height=5.5pc, minimum width=4.5pc, line width=.1pc, draw, red, align=center]{};
    \node (02) at (B) [ellipse, minimum height=5.5pc, minimum width=4.5pc, line width=.1pc, draw, red, align=center]{};
    \node (03) at (C) [ellipse, minimum height=5.5pc, minimum width=4.5pc, line width=.1pc, draw, red, align=center]{};
    \node (04) at (D) [ellipse, minimum height=5.5pc, minimum width=4.5pc, line width=.1pc, draw, red, align=center]{};
    \node[] (11) [below of =A , node distance=3.8pc, anchor=north, align=center]{\footnotesize{Dancer.}};
    \node[] (12) [below of =B , node distance=3.8pc, anchor=north, align=center]{\footnotesize{Dancer.}};
    \node[] (13) [below of =C , node distance=3.8pc, anchor=north, align=center]{\footnotesize{Dancer.}};
    \node[] (14) [below of =D , node distance=3.8pc, anchor=north, align=center]{\footnotesize{Dancer.}};
    %kendo
    \node[] (A2) [ right of =A  , node distance=8.25pc,align=center] {\includegraphics[width=7.5pc, height=7.5pc]{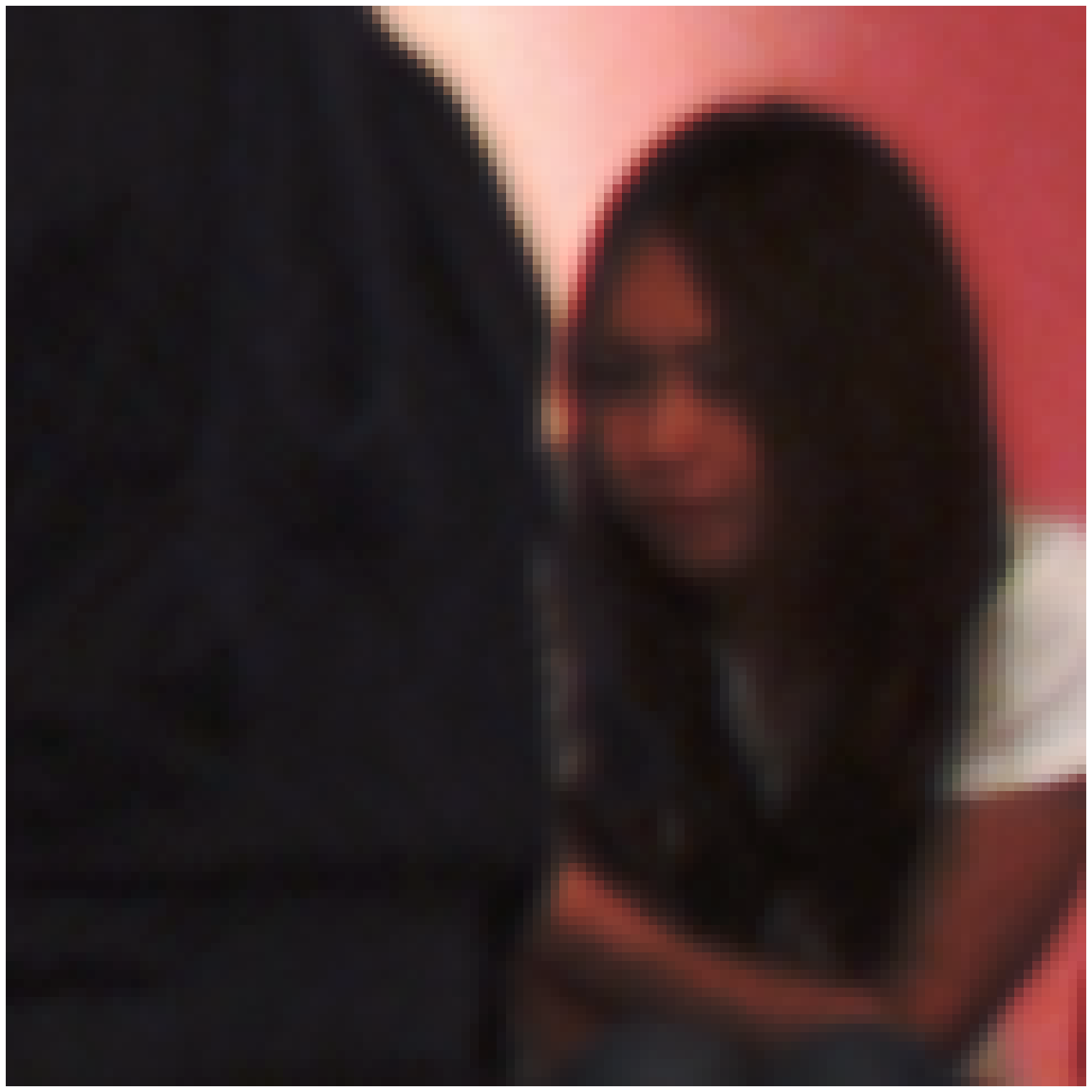}};
    \node[] (B2) [ below of =A2 , node distance=9.85pc,align=center] {\includegraphics[width=7.5pc, height=7.5pc]{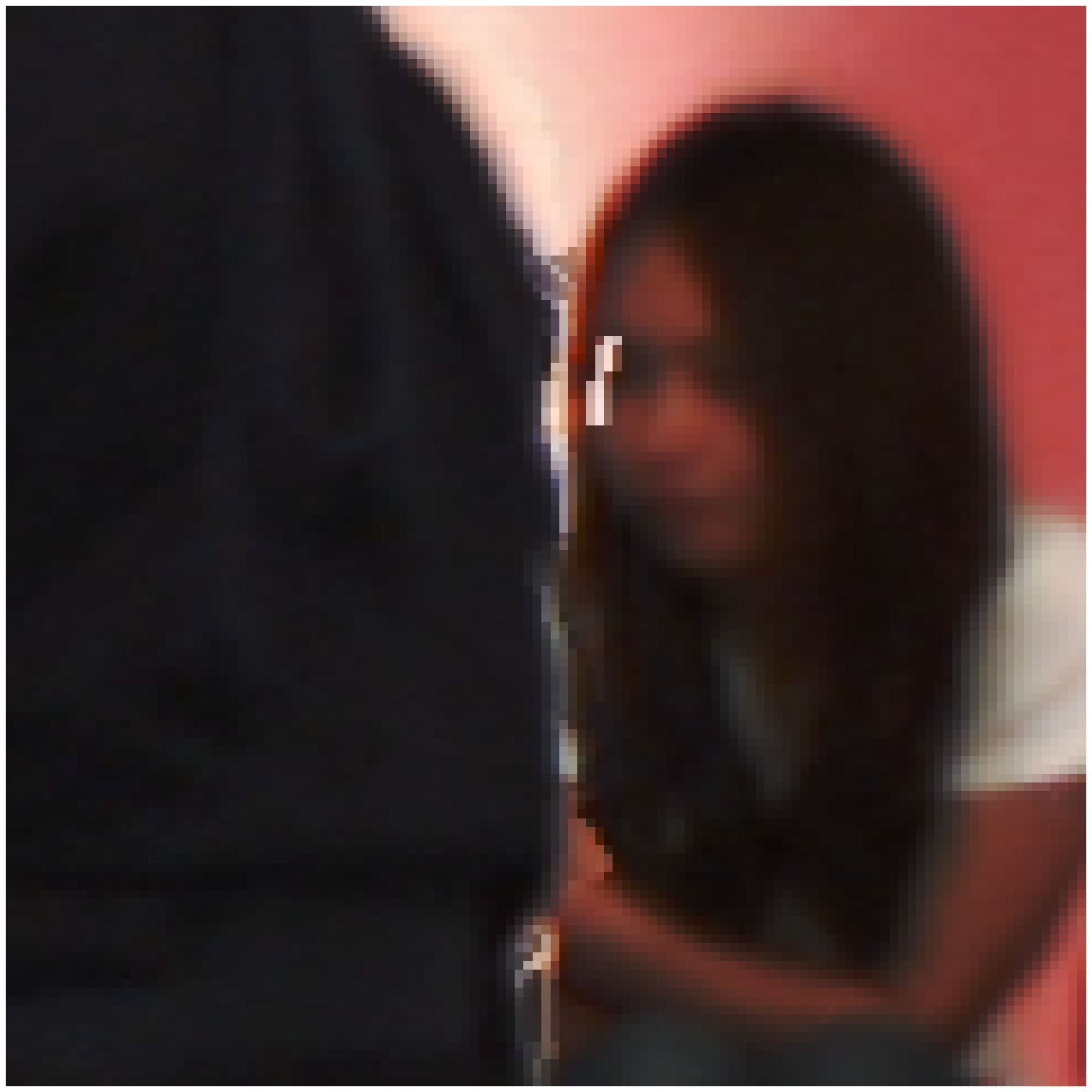}};
    \node[] (C2) [ below of =B2 , node distance=9.85pc,align=center] {\includegraphics[width=7.5pc, height=7.5pc]{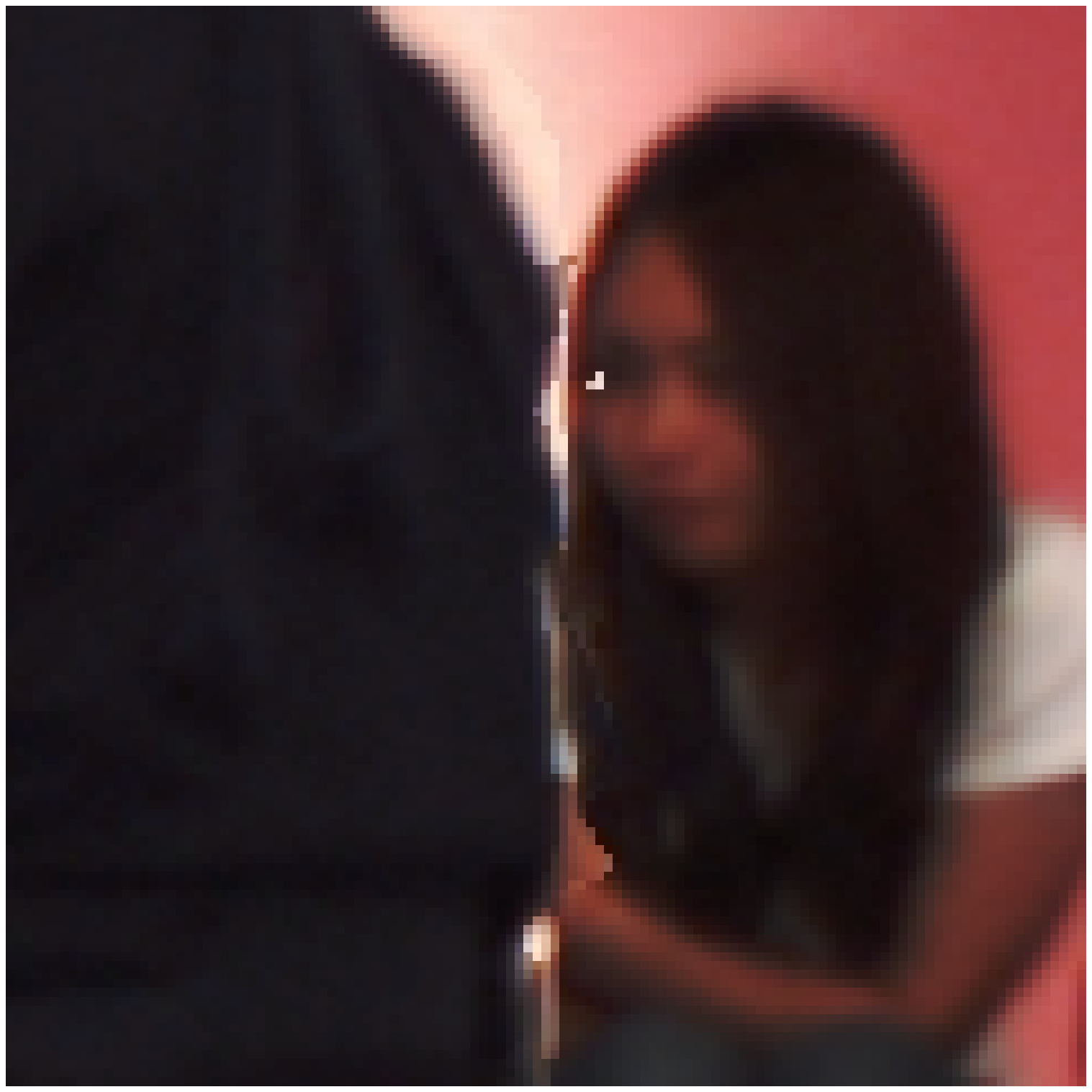}};
    \node[] (D2) [ below of =C2 , node distance=9.85pc,align=center] {\includegraphics[width=7.5pc, height=7.5pc]{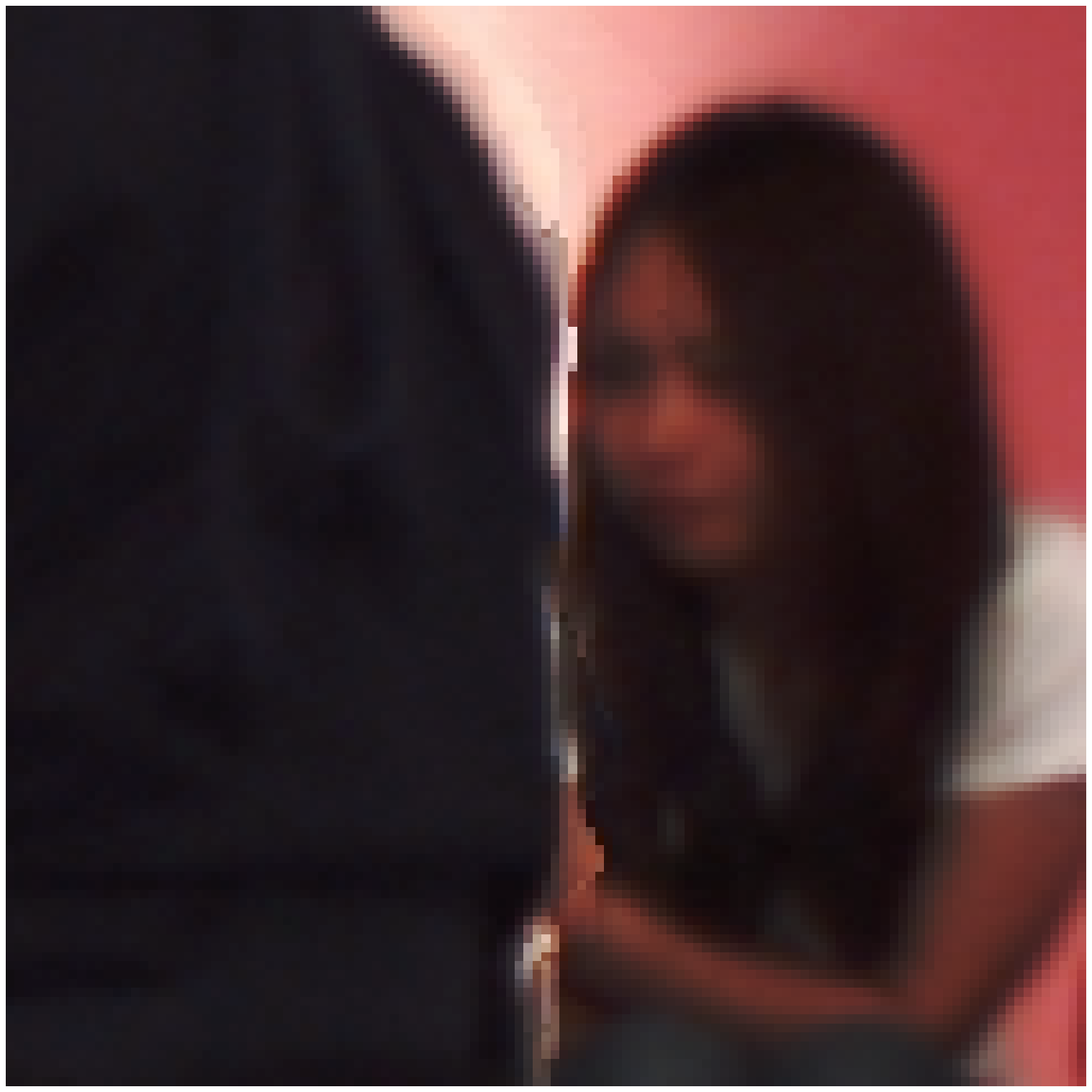}};
    \node (71) at (A2) [ellipse, minimum height=6pc, minimum width=3pc, line width=.1pc, draw, red, align=center]{};
    \node (72) at (B2) [ellipse, minimum height=6pc, minimum width=3pc, line width=.1pc, draw, red, align=center]{};
    \node (73) at (C2) [ellipse, minimum height=6pc, minimum width=3pc, line width=.1pc, draw, red, align=center]{};
    \node (74) at (D2) [ellipse, minimum height=6pc, minimum width=3pc, line width=.1pc, draw, red, align=center]{};
    \node (81) [below of =A2 , node distance=3.8pc, anchor=north, align=center]{\footnotesize{Kendo.}};
    \node (82) [below of =B2 , node distance=3.8pc, anchor=north, align=center]{\footnotesize{Kendo.}};
    \node (83) [below of =C2 , node distance=3.8pc, anchor=north, align=center]{\footnotesize{Kendo.}};
    \node (84) [below of =D2 , node distance=3.8pc, anchor=north, align=center]{\footnotesize{Kendo.}};
    % balloons
    \node[] (A0) [ right of =A2 , node distance=8.25pc,align=center] {\includegraphics[width=7.5pc, height=7.5pc]{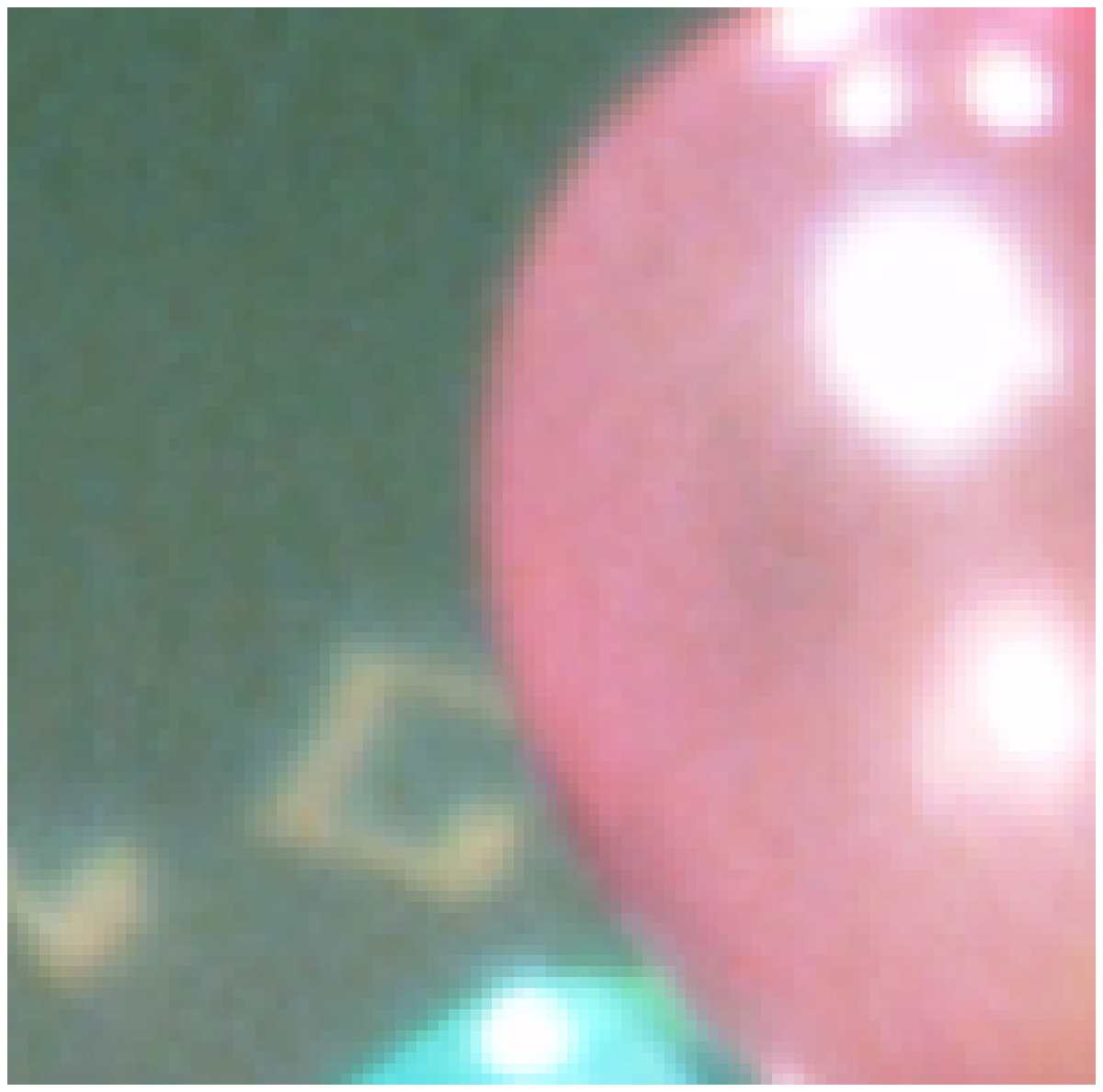}};
    \node[] (B0) [ below of =A0 , node distance=9.85pc,align=center] {\includegraphics[width=7.5pc, height=7.5pc]{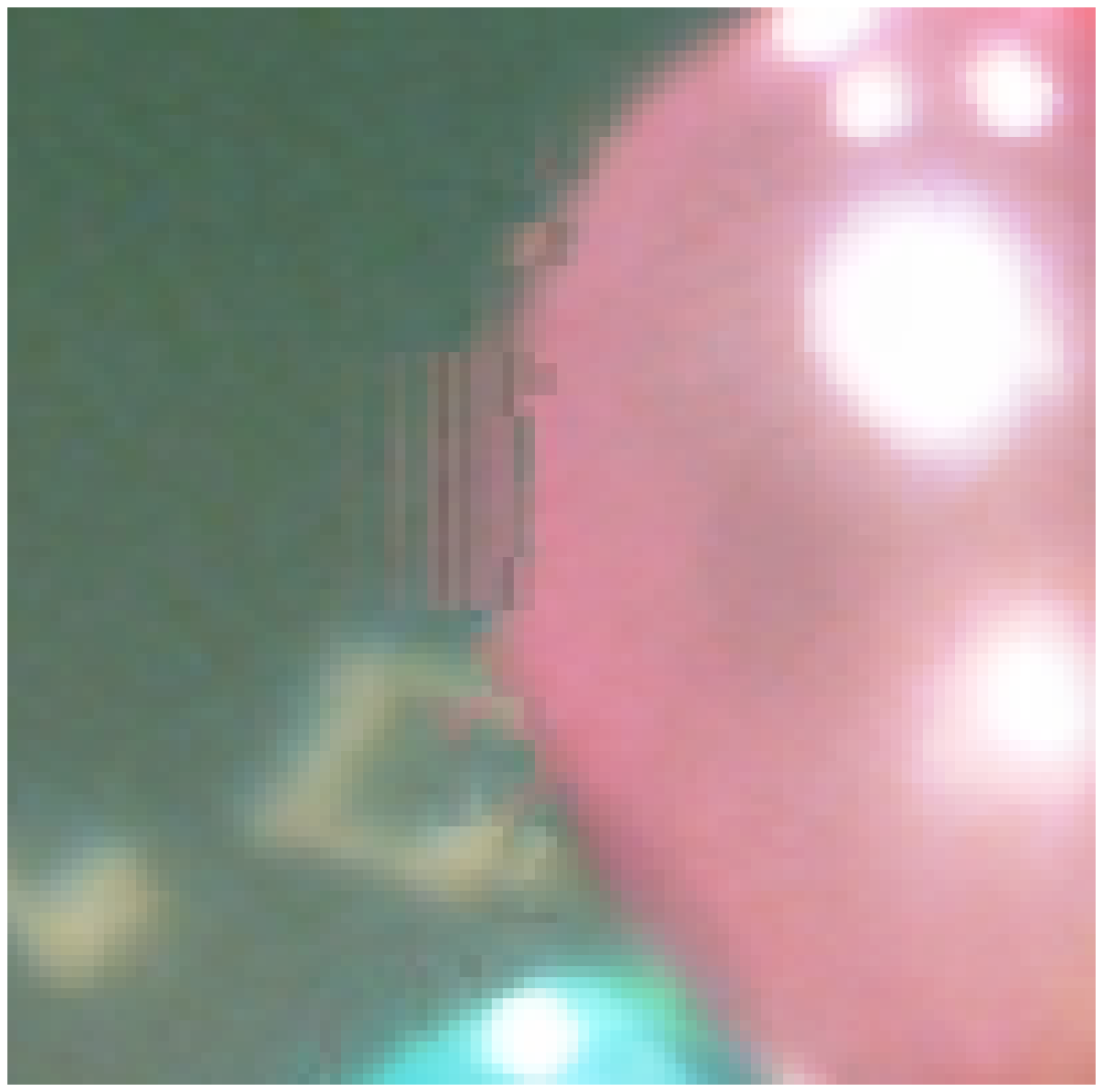}};
    \node[] (C0) [ below of =B0 , node distance=9.85pc,align=center] {\includegraphics[width=7.5pc, height=7.5pc]{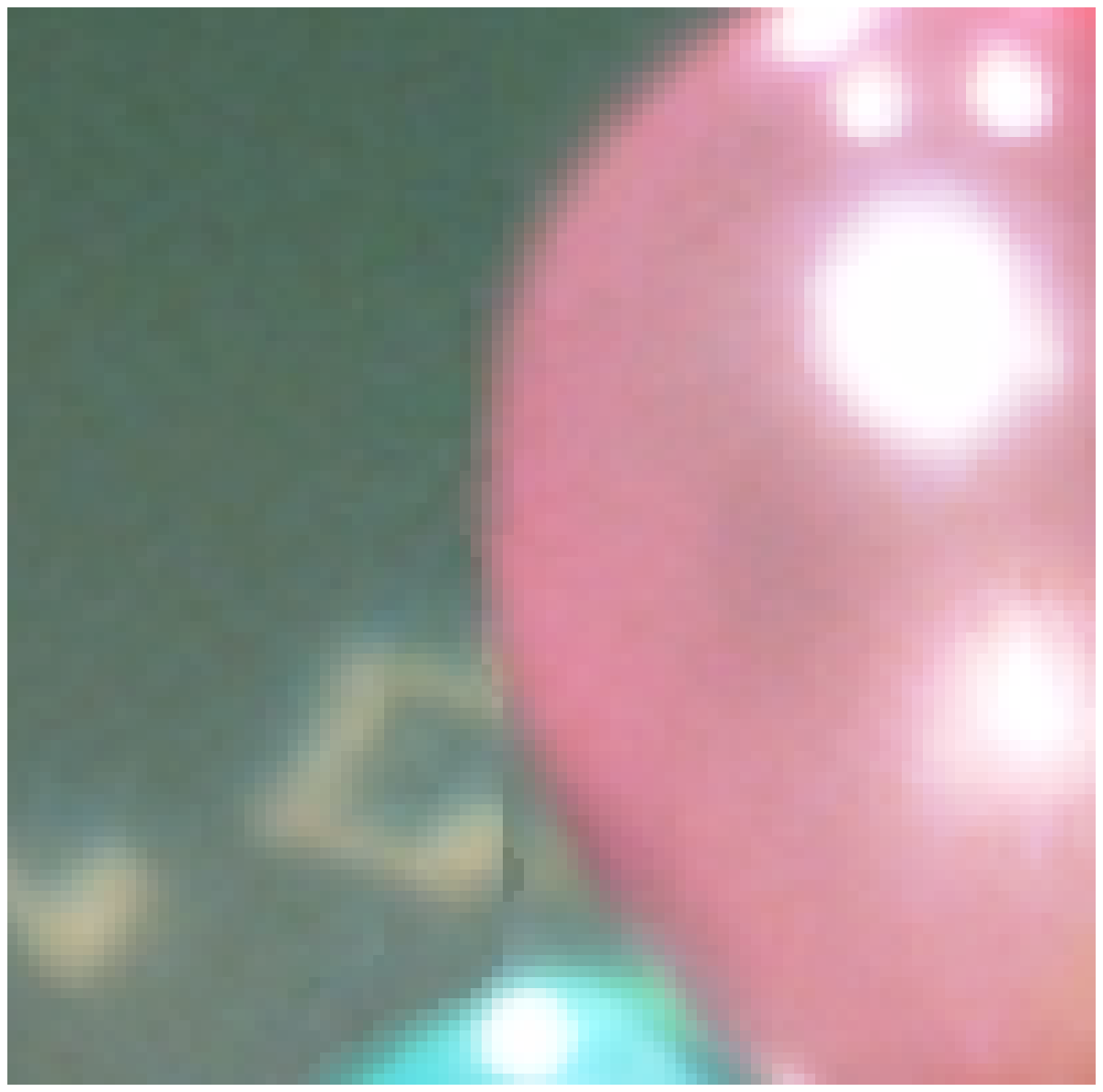}};
    \node[] (D0) [ below of =C0 , node distance=9.85pc,align=center] {\includegraphics[width=7.5pc, height=7.5pc]{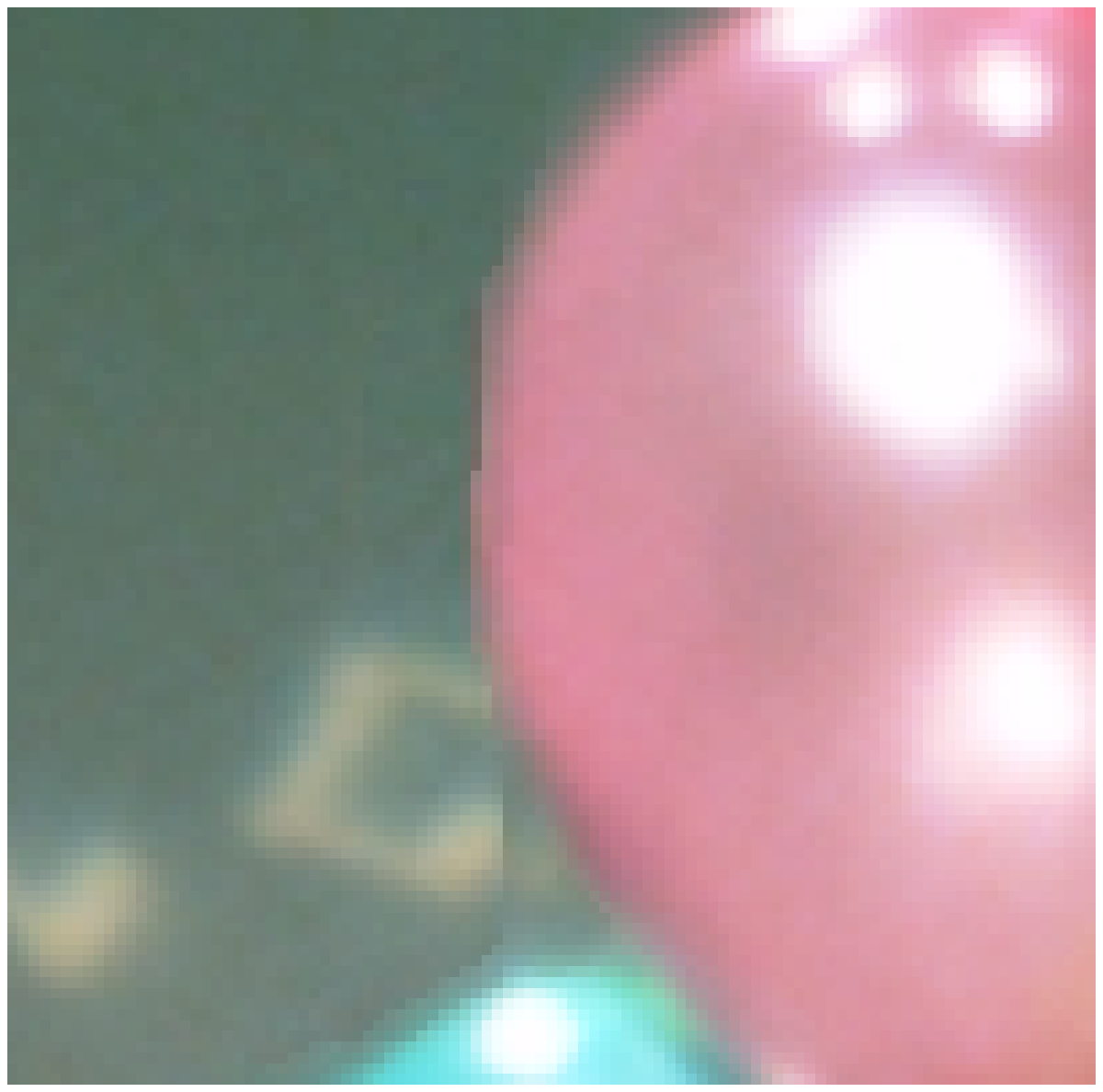}};
    \node (21) at (A0) [ellipse, minimum height=5.5pc, minimum width=4.5pc, line width=.1pc, draw, red, align=center]{};
    \node (22) at (B0) [ellipse, minimum height=5.5pc, minimum width=4.5pc, line width=.1pc, draw, red, align=center]{};
    \node (23) at (C0) [ellipse, minimum height=5.5pc, minimum width=4.5pc, line width=.1pc, draw, red, align=center]{};
    \node (24) at (D0) [ellipse, minimum height=5.5pc, minimum width=4.5pc, line width=.1pc, draw, red, align=center]{};
    \node[] (41) [below of =A0 , node distance=3.8pc, anchor=north, align=center]{\footnotesize{Balloons.}};
    \node[] (42) [below of =B0 , node distance=3.8pc, anchor=north, align=center]{\footnotesize{Balloons.}};
    \node[] (43) [below of =C0 , node distance=3.8pc, anchor=north, align=center]{\footnotesize{Balloons.}};
    \node[] (44) [below of =D0 , node distance=3.8pc, anchor=north, align=center]{\footnotesize{Balloons.}};
    %lovebird1
    \node[] (A1) [ right of =A0,  node distance=8.25pc,align=center] {\includegraphics[width=7.5pc, height=7.5pc]{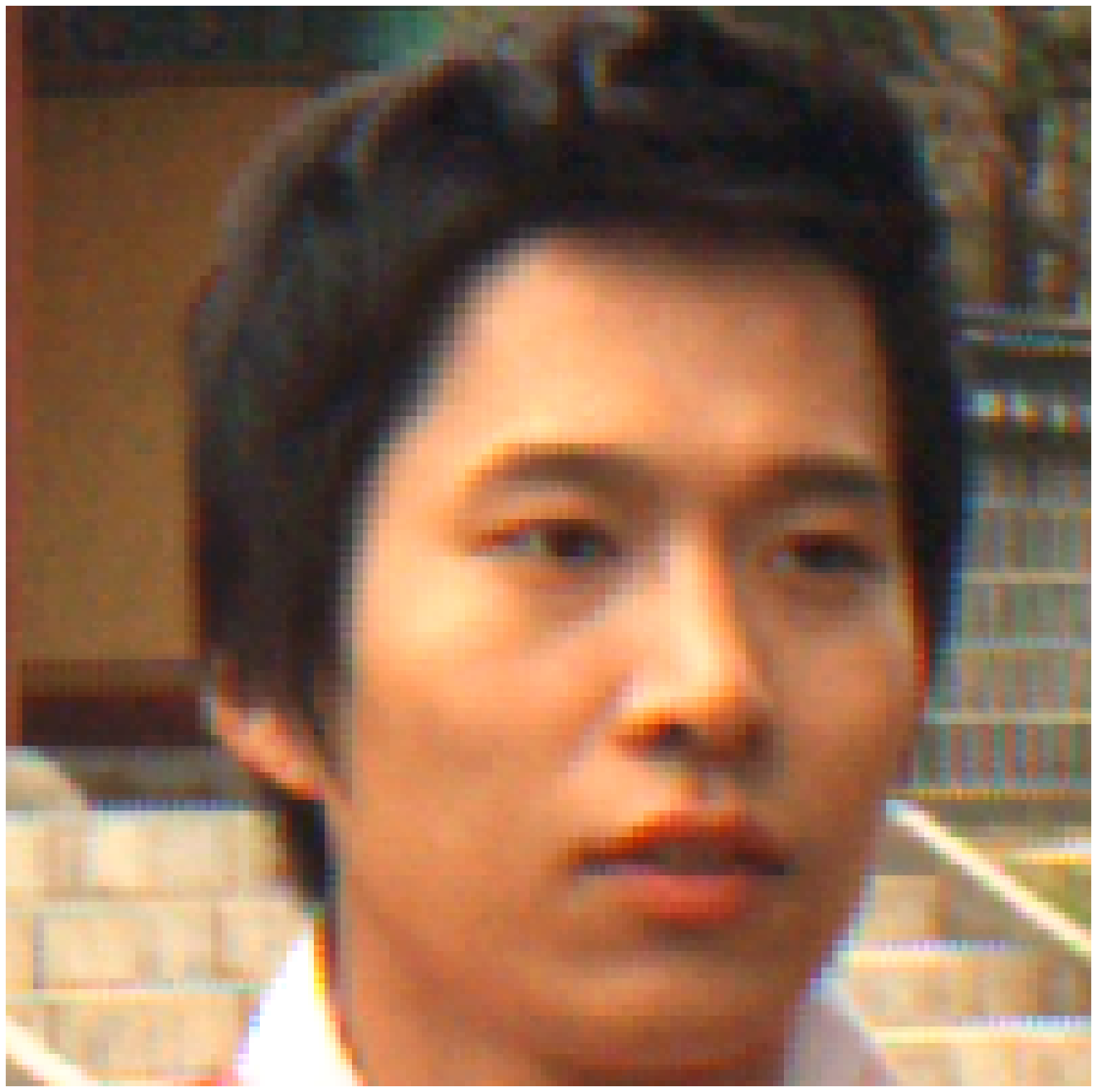}};
    \node[] (B1) [ below of =A1 , node distance=9.85pc,align=center] {\includegraphics[width=7.5pc, height=7.5pc]{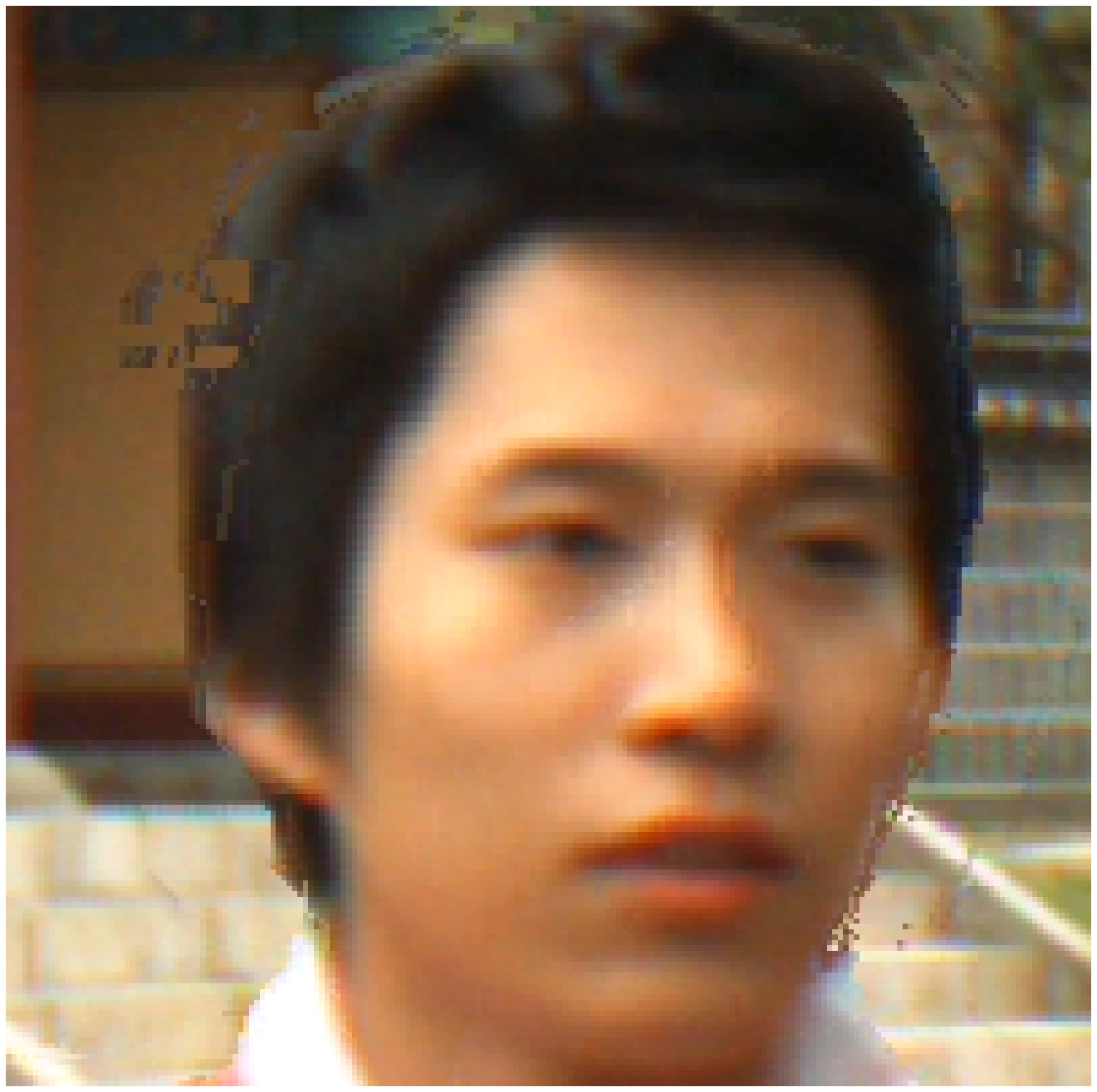}};
    \node[] (C1) [ below of =B1 , node distance=9.85pc,align=center] {\includegraphics[width=7.5pc, height=7.5pc]{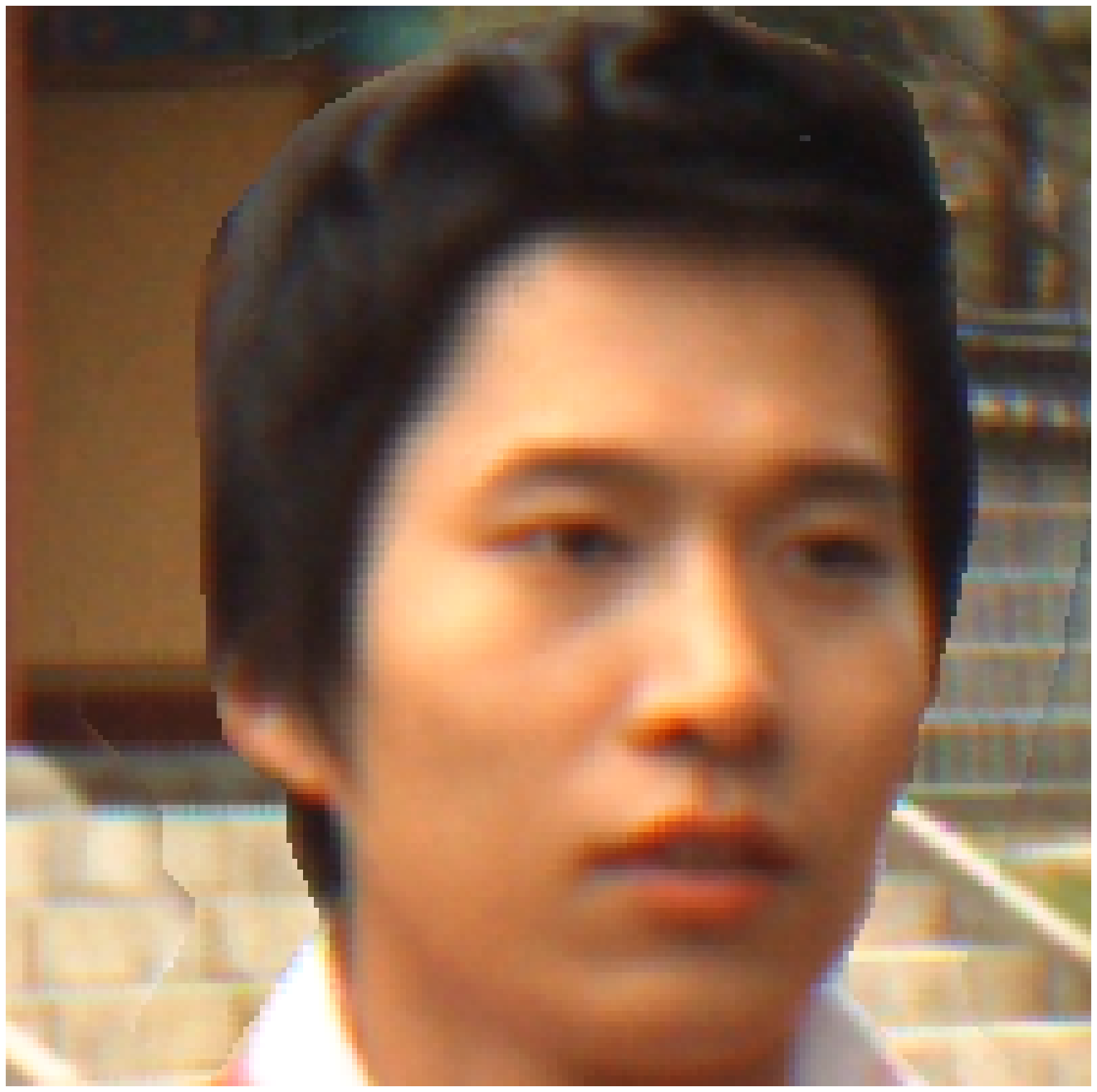}};
    \node[] (D1) [ below of =C1 , node distance=9.85pc,align=center] {\includegraphics[width=7.5pc, height=7.5pc]{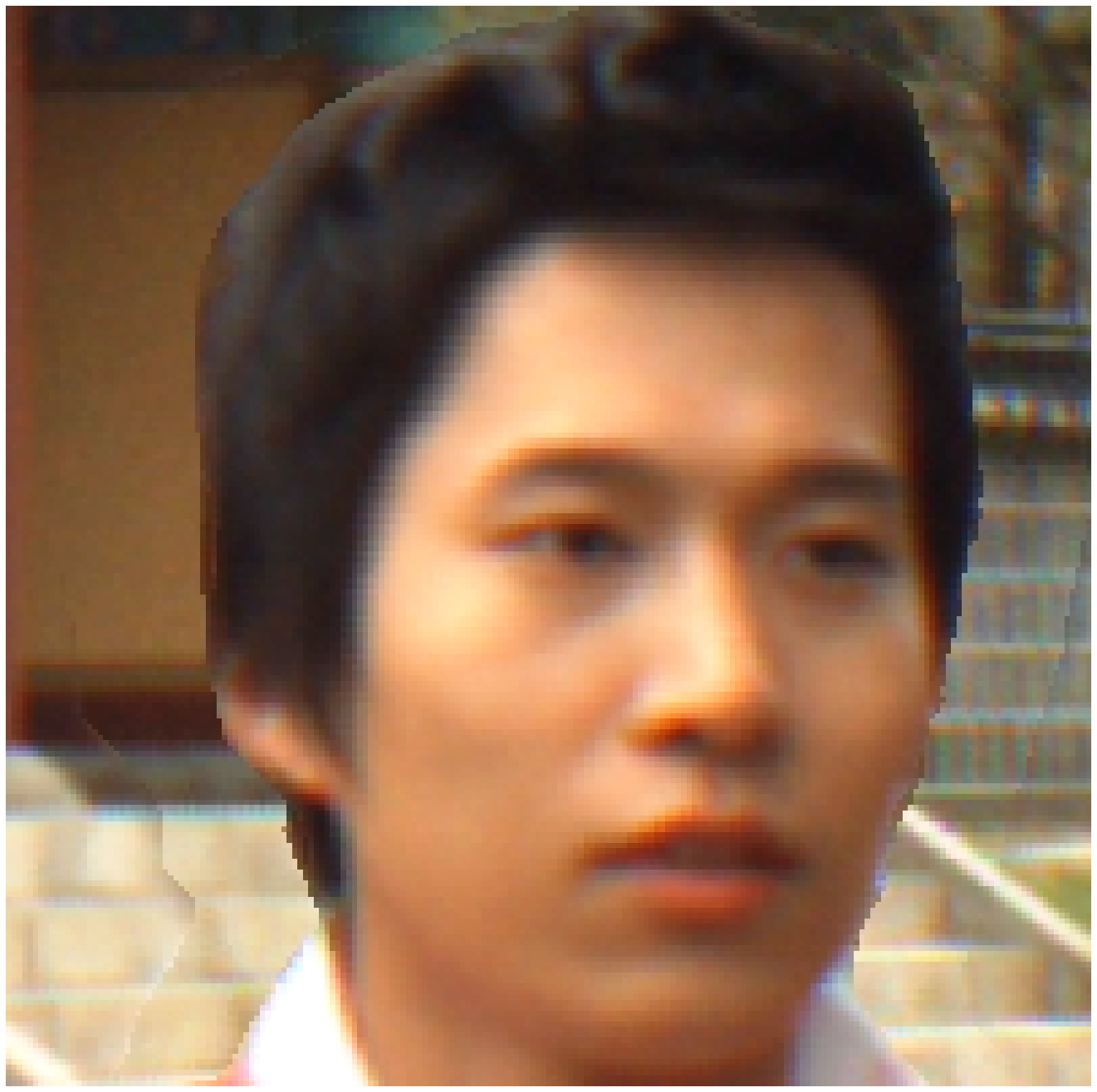}};

    \node (51) [ellipse, above left of =A1, node distance=3pc, minimum height=3.0pc, minimum width=3pc, line width=.1pc, draw, white, align=center]{};
    \node (52) [ellipse, above left of =B1, node distance=3pc, minimum height=3.0pc, minimum width=3pc, line width=.1pc, draw, white, align=center]{};
    \node (53) [ellipse, above left of =C1, node distance=3pc, minimum height=3.0pc, minimum width=3pc, line width=.1pc, draw, white, align=center]{};
    \node (54) [ellipse, above left of =D1, node distance=3pc, minimum height=3.0pc, minimum width=3pc, line width=.1pc, draw, white, align=center]{};

    \node (61) [below of =A1 , node distance=3.8pc, anchor=north, align=center]{\footnotesize{Lovebird1.}};
    \node (62) [below of =B1 , node distance=3.8pc, anchor=north, align=center]{\footnotesize{Lovebird1.}};
    \node (63) [below of =C1 , node distance=3.8pc, anchor=north, align=center]{\footnotesize{Lovebird1.}};
    \node (64) [below of =D1 , node distance=3.8pc, anchor=north, align=center]{\footnotesize{Lovebird1.}};

    % newspaper
    \node (A3) [ right of =A1 , node distance=8.25pc,align=center] {\includegraphics[width=7.5pc, height=7.5pc]{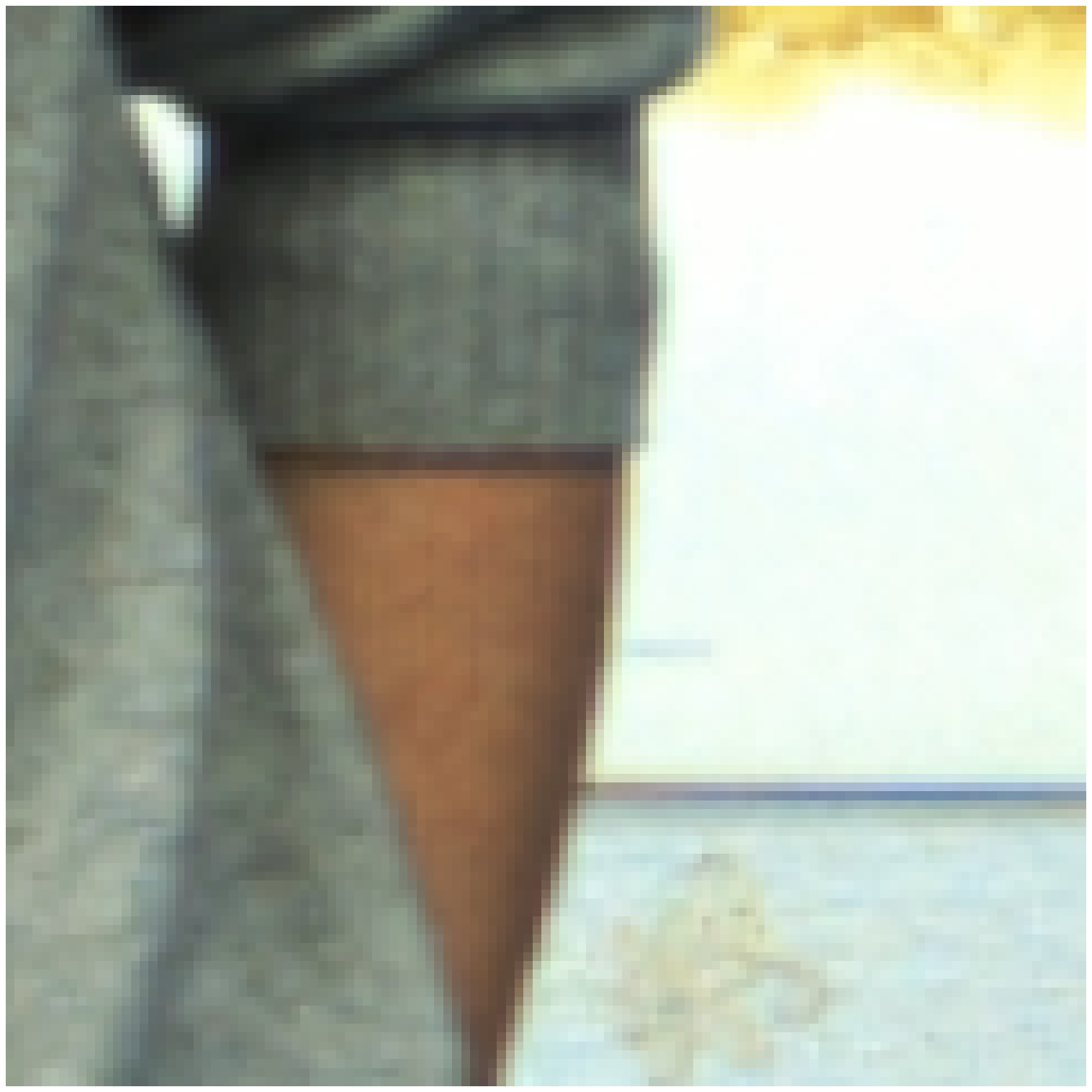}};
    \node (B3) [ below of =A3 , node distance=9.85pc,align=center] {\includegraphics[width=7.5pc, height=7.5pc]{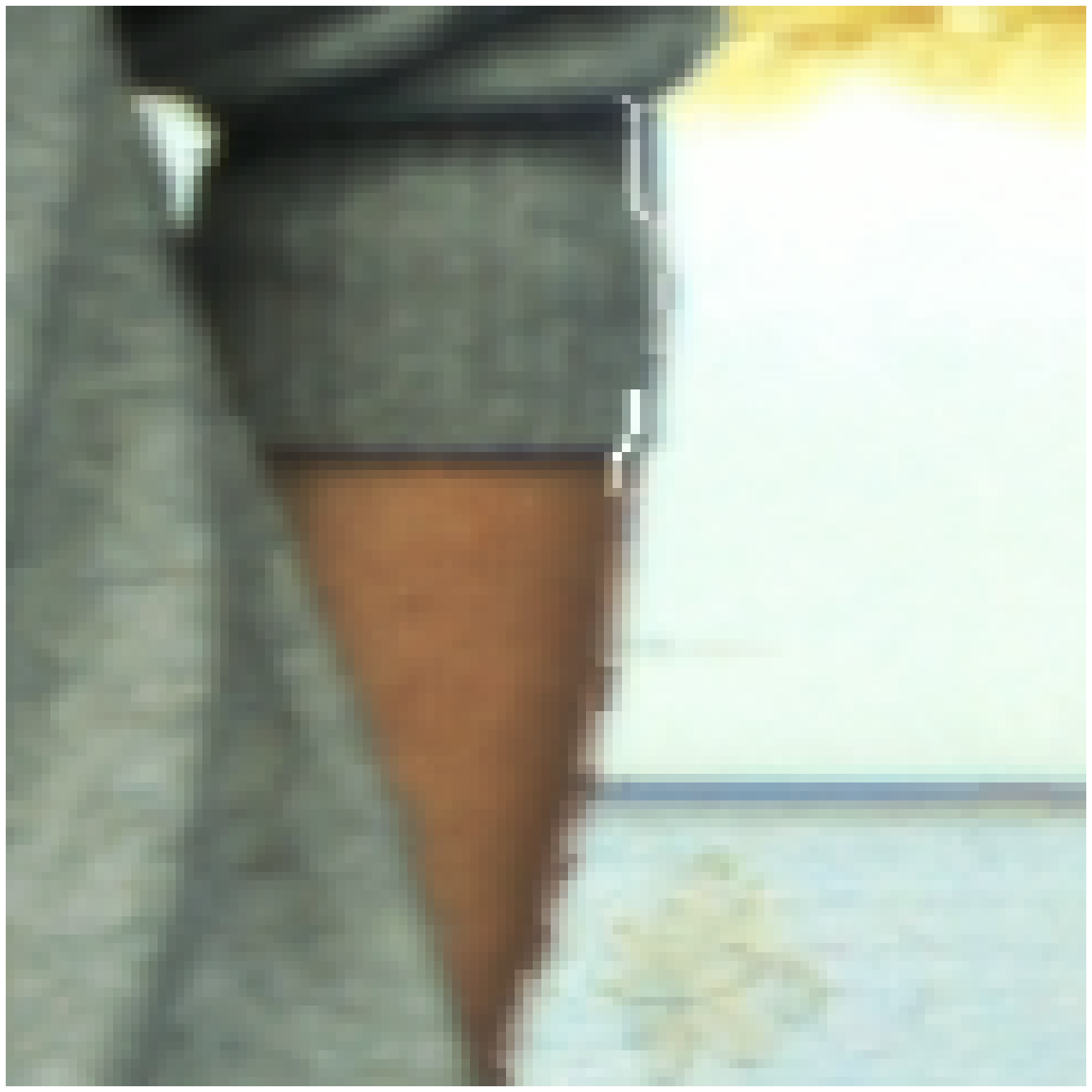}};
    \node (C3) [ below of =B3 , node distance=9.85pc,align=center] {\includegraphics[width=7.5pc, height=7.5pc]{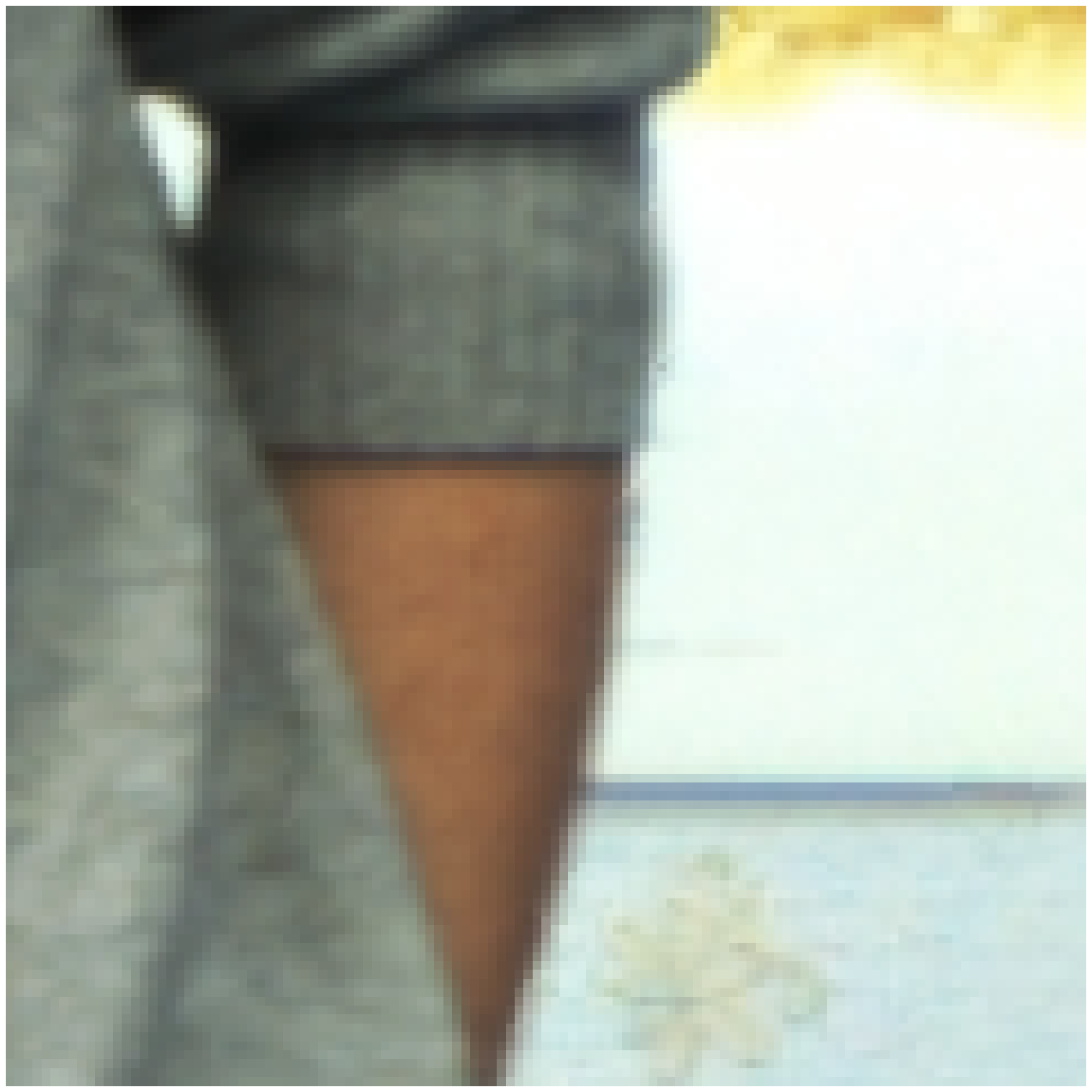}};
    \node (D3) [ below of =C3 , node distance=9.85pc,align=center] {\includegraphics[width=7.5pc, height=7.5pc]{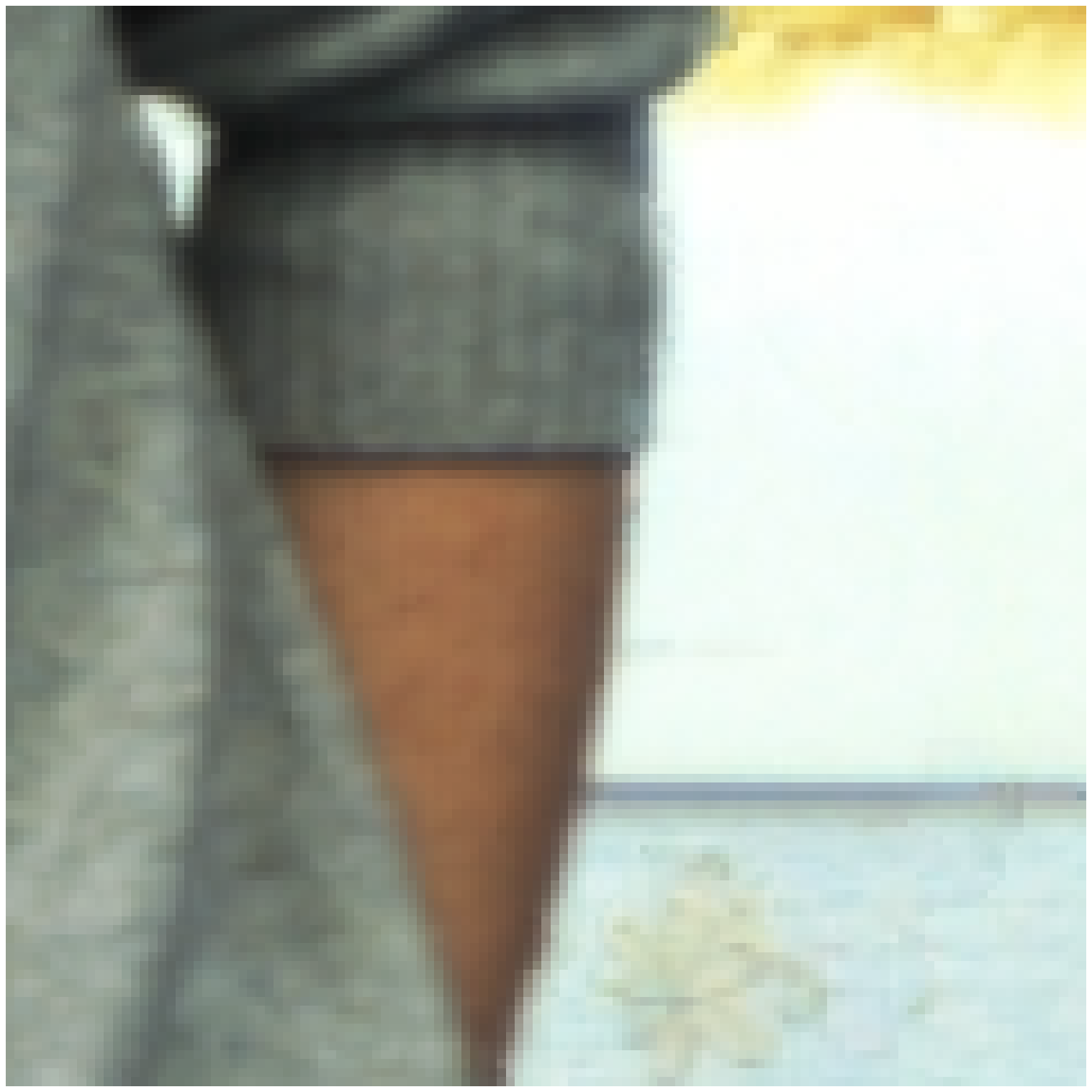}};

    \node (91) at (A3) [ellipse, minimum height=7.2pc, minimum width=4pc, line width=.1pc, draw, red, align=center]{};
    \node (92) at (B3) [ellipse, minimum height=7.2pc, minimum width=4pc, line width=.1pc, draw, red, align=center]{};
    \node (93) at (C3) [ellipse, minimum height=7.2pc, minimum width=4pc, line width=.1pc, draw, red, align=center]{};
    \node (94) at (D3) [ellipse, minimum height=7.2pc, minimum width=4pc, line width=.1pc, draw, red, align=center]{};

    \node (101) [below of =A3 , node distance=3.8pc, anchor=north, align=center]{\footnotesize{Newspaper.}};
    \node (102) [below of =B3 , node distance=3.8pc, anchor=north, align=center]{\footnotesize{Newspaper.}};
    \node (103) [below of =C3 , node distance=3.8pc, anchor=north, align=center]{\footnotesize{Newspaper.}};
    \node (104) [below of =D3 , node distance=3.8pc, anchor=north, align=center]{\footnotesize{Newspaper.}};
    \node[] (W) [below of =41, node distance=1.0pc, align=center]{\footnotesize{(a)~Original.}};
    \node[] (X) [below of =42, node distance=1.0pc, align=center]{\footnotesize{(b)~MPEG/D$\rightarrow$VSRS 3.5.}};
    \node[] (Y) [below of =43, node distance=1.0pc, align=center]{\footnotesize{(c)~MPEG/D$\rightarrow$IVDCT/ED$\rightarrow$VSRS 3.5.}};
    \node[] (Z) [below of =44, node distance=1.0pc, align=center]{\footnotesize{(d)~MPEG/D$\rightarrow$CAVS.}};
\end{tikzpicture}
\caption{\label{fig:sub_results}Examples of the synthesized views for the test sequences as generated by VSRS 3.5 when (b) using MPEG depth maps (MPEG/D) and (c) IVDCT enhanced depth maps (IVDCT/ED).  In (d), examples of the synthesized views for the test sequences are generated by CAVS when using MPEG/D. The artifacts that appear in the views when synthesized by VSRS 3.5 using MPEG/D are efficiently suppressed by using IVDCT/ED. Further, CAVS improves the quality of the synthesized views by exploiting the resulting IVDCT consistency information. The ellipses/circles mark the improved regions for a detailed comparison. (Best viewed in color).}
\end{figure*}
\subsection{Consistency-Adaptive View Synthesis}
\label{subsec:consistency information adaptive view synthesis}
Fig.~\ref{fig:cavs_block} summarizes the consistency-adaptive view synthesis (CAVS). Here, we aim to perform view synthesis at a virtual viewpoint that is the same as the principal viewpoint of IVDCT. A virtual viewpoint is a viewpoint at which no physical camera is  available to view and record the scene. IVDCT allows us to generate inter-view consistency information at the virtual viewpoint that will be helpful for view synthesis. In particular, the consistency information for a given virtual pixel is used to control the warping and fusion of view pixels from multiple reference views. However, a view pixel with no inter-view consistency information (extreme error event) is not determined by warping. Such pixels are marked by a mask that allows other techniques to fill in the missing intensity values.

If inter-view consistency information is available for a given pixel in the virtual view, we use various approaches to fuse adaptively warped inter-view consistent view pixels to obtain the final pixel intensity in the virtual view. The fusion of pixel values depends mainly on the baseline scenario and the varying illumination conditions among the reference views. If the pixel intensities of chosen reference pixels are similar, averaging of the warped pixel intensities is feasible.  To maintain color consistency, the similarity is defined in terms of the Euclidean distance. However, if the pixel intensities among the chosen references differ significantly due to varying illumination, we assume that the virtual pixel value is best described by the warped texture pixel of the nearest reference view. The reference view which has minimum baseline distance from the virtual viewpoint is defined as the nearest view. In this case, we simply set the pixel intensity in the virtual view by copying the pixel intensity from the warped view pixel of the nearest reference view that is connected. If the reference views are captured from multiple viewpoints using irregular camera baseline distances, we estimate the virtual pixel intensity by weighted-baseline averaging of the chosen references.

Information about all possible object points of a natural 3D scene is not available in a single viewpoint. This leads to disocclusion in virtual views. For example, background which is covered by foreground objects in the reference view may be disoccluded in the virtual view. Therefore,  several pixels in the virtual view cannot be specified. Increasing the number of reference views is likely to decrease the number of these pixels. However, these pixels cannot be ruled out completely. Therefore, all unspecified pixels (holes) in the virtual view are filled by inpainting~\cite{Bertalmio2001}.

\section{Results and Discussion}
\label{sec:results and discussion}
To evaluate the efficiency of the proposed consistency testing, we conducted two kinds of experiments. The first experiment studies the effect of our depth maps enhancement scheme on view synthesis. The second experiment evaluates the consistency-adaptive view synthesis. In these experiments, we assess the quality of synthesized virtual views. We measure the objective video quality of the synthesized view at a given viewpoint by means of the Peak Signal-to-Noise Ratio (PSNR) with respect to the captured view of a real camera at the same viewpoint. For the experiments, we use five standard MVV test data sets and the corresponding depth maps from three different viewpoints as provided by MPEG~\cite{MPEG:N12036}: Newspaper ($1024\times768$), Kendo~($1024\times768$), Balloons ($1024\times768$), Lovebird1 ($1024\times768$), and Dancer ($1920\times1088$). Note, the Dancer test data is a synthetic test material with consistent depth maps across all viewpoints.

% depth map enhancement
Since, in FTV scenario, depth maps are not going to be viewed by end users. Therefore, to evaluate the depth map enhancement by IVDCT, we assess the effect of depth enhancement on virtual view synthesis. The virtual views are synthesized by MPEG View Synthesis Reference Software (VSRS) 3.5 which is an DIBR approach~\cite{MPEG:M15377}. VSRS 3.5 uses two reference views, left and right, to synthesizes a virtual view at an arbitrary intermediate viewpoint by using the two corresponding reference depth maps and camera parameters. We compare the subjective and objective quality of virtual views as synthesized by VSRS 3.5 with the help of MPEG depth maps and improved depth maps from our approach. First, the depth imagery from three viewpoints is improved by utilizing the proposed IVDCT as discussed in~\ref{sec:subspace_depth consistency testing} with $\alpha=1/2$. Second, a virtual view  for a given viewpoint is synthesized by VSRS 3.5 using the improved depth maps. For view synthesis, the 1D parallel synthesis mode of VSRS 3.5 is used with half-pel precision. Table~\ref{tab:results} shows a comparison of PSNR values (in dB) for the synthesized virtual views as generated by VSRS 3.5 when using (a) MPEG depth maps (MPEG/D) and (b) IVDCT enhanced depth maps (IVDCT/ED).  Our enhancement algorithm offers an improvement of up to 0.7 dB. The improvement in quality is likely to increase with an increasing number of reference viewpoints used for the testing. It also depends on the quality of the input reference depth maps at various viewpoints. Note that our enhancement algorithm does not offer gains for the synthetic Dancer sequence because the synthetic depth maps are consistent across all viewpoints. However, VSRS 3.5 can not efficiently exploit our enhanced depth maps and the consistency information fully due to its input requirements.

Our proposed consistency-adaptive view synthesis efficiently exploit the inter-view consistency information and further improve the quality of the virtual views. To demonstrate this, CAVS is used for the view synthesis at a virtual viewpoint. For this, we first perform IVDCT at the virtual viewpoint by utilizing depth maps from three reference viewpoints with $\alpha=1/2$. The resulting consistency information is used to adaptively determine the virtual view pixel intensity by using views from three reference viewpoints, as discussed in~\ref{subsec:consistency information adaptive view synthesis}. In Table~\ref{tab:results}, the quality of CAVS virtual views are compared to the virtual views as synthesized by VSRS 3.5 using MPEG provided depth maps and IVDCT enhanced depth maps. We observe that CAVS offers a PSNR gain of up to 1.2 dB. Note, CAVS even offers gains for the synthetic Dancer sequence. This is because our CAVS efficiently supervises view pixel selection from multiple viewpoints through inter-view consistency information. Especially around the edges where information is missing from one viewpoint CAVS chooses adaptively consistent information from another viewpoint. VSRS 3.5 lacks in this aspect. Hence, inter-view consistency information is relevant for view synthesis.

In general, our algorithms improve the FTV visual experience by efficiently reducing visually annoying artifacts in the virtual views, as illustrated in Fig.~\ref{fig:sub_results} for the test sequences. These improvements are exclusively offered by the inter-view consistency information. To put emphasis on the improvements, Fig.~\ref{fig:sub_results} shows the selected regions of synthesized virtual views for the test sequences. Virtual views synthesized by VSRS 3.5 using MPEG depth maps are used as the base for the comparison. We suppress artifacts in the hand of the Dancer efficiently. The visual quality of the synthesized Kendo view improves, especially the eye of the spectator is well synthesized.  The artifacts around the balloon boundaries are efficiently suppressed by our proposed algorithms for the Balloons sequence. Artifacts around the hair of the man have been reduced for Lovebird1 by exploiting inter-view consistency information.  Furthermore, areas around the sweater sleeve edges have been improved for the Newspaper.

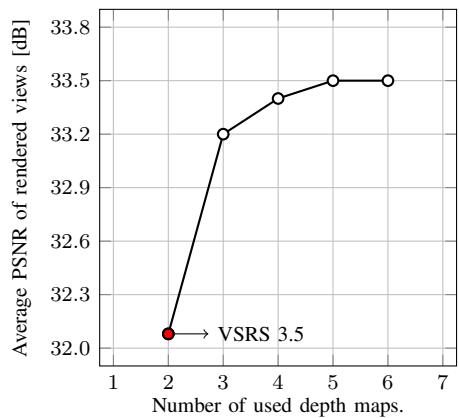
\begin{figure}[t!]
\centering
\tikzstyle{init} = [pin edge={-to,thin,black}]
\begin{tikzpicture}
  \begin{axis}[name=viewnumber,
     width=15pc, height=15pc,
     xstep=1, xmin=00.75, xmax=07.25, xtick={1,2,3,4,5,6,7},
     ystep=1, ymin=31.90, ymax=33.90, ytick={32.00,32.30,32.60,32.90,33.20,33.50,33.80},
     legend pos=south east, ylabel ={Average PSNR of rendered views [dB]}, xlabel ={Number of used depth maps.}, grid =major, y label style={at={(0.04,0.5)}},legend cell align=left, legend style={font=\footnotesize}, x tick style={font=\footnotesize}, y tick style={font=\footnotesize}, x tick label style={font=\footnotesize},  y tick label style={font=\footnotesize, /pgf/number format/.cd, fixed, fixed zerofill, precision=1, /tikz/.cd}, y label style={font=\footnotesize}, x label style={at={(0.5, 0.05)},font=\footnotesize}
     ]
    \addplot[black, thick, mark=*, mark options={fill=white}] plot coordinates {(2, 32.08) (3, 33.2) (4, 33.4) (5, 33.5) (6, 33.5)};
    \addplot[black, thick, mark=*, mark options={fill=red}  ] plot coordinates {(2, 32.08)};
    \node[coordinate, pin={[init]right:{\footnotesize VSRS 3.5}}]at(axis cs:2.08,32.08){};
  \end{axis}
\end{tikzpicture}
  \caption{\label{fig:number_of depth maps}Average PSNR of synthesized virtual views over the number of depth maps used by CAVS for synthesis. VSRS 3.5 is used as a reference for the synthesis quality and uses only two depth maps. The experiment is based on 50 frames of the data set Newspaper.}
\end{figure}

Moreover, by increasing the number of depth maps used by CAVS, both inter-view consistency and virtual view quality improve. Fig.~\ref{fig:number_of depth maps} shows this trend. For view synthesis using CAVS with three depth maps, we observe significant improvements in virtual view quality when compared to VSRS 3.5, which uses two depth maps. A further increase in the number of used depth maps gives additional small improvements. This is because the impact of reference views decreases as the distance between virtual and reference viewpoints increases.

\subsection{Quantization Noise}
\label{subsec:quantization noise}

In a FTV system~\cite{Tanimoto2011}, both our IVDCT based depth enhancement algorithm and consistency-adaptive view synthesis can be used at the receiver side. Usually, due to bit-rate budget constraints, depth maps are coded at a quantization parameter and transmitted to the receiver. The receiver reconstructs the quantized depth maps for view synthesis. The quality of the synthesized view is a indirect measure of the quality of the reconstructed depth maps~\cite{Mueller2011}.

\begin{figure*}[t!]
  \centering
  \begin{tikzpicture}
  %dancer
  \begin{axis}[name=dancer,
    width=15pc, height=15pc, grid =major,
    xstep=1, xmin=23.25, xmax=42.75,
    ystep=1, ymin=32.875, ymax=39.125,
    legend pos=south west,
    xlabel ={Quantization parameter (QP)}, xtick={24,27,30,33,36,39,42}, x label style={at={(0.5, 0.05)}, font=\footnotesize}, x tick style={font=\footnotesize}, x tick label style={font=\footnotesize},
    ylabel ={Average PSNR of rendered views [dB]}, y label style={at={(0.07,0.5)}, font={\footnotesize}}, y tick style={font=\footnotesize}, y tick label style={font=\footnotesize},
    ytick={33.00,34.00,35.00,36.00,37.00,38.00,39.00},
    y tick label style={font=\footnotesize, /pgf/number format/.cd, fixed, fixed zerofill, precision=1, /tikz/.cd},
    legend cell align=left, legend style={font=\footnotesize},
    ]
  \addplot[red  , thick, mark=square* ,mark options={fill=red  }]  plot coordinates {(24,37.73)(30,37.00)(36,35.49)(42,34.50)};% VSRS $\mu=0.5$ OK
  \addplot[black, thick, mark=*,       mark options={fill=black}]  plot coordinates {(24,38.34)(30,37.30)(36,35.80)(42,34.72)};% IMDCT-> VSRS OK
  \addplot[black, thick, mark=*,       mark options={fill=white}]  plot coordinates {(24,38.95)(30,37.55)(36,36.00)(42,34.80)};% CAVS
  \legend{VSRS 3.5,IVDCT/ED$\rightarrow$VSRS3.5,CAVS};
  \end{axis}
  % kendo
  \begin{axis}[name=kendo,  at=(dancer.right of south east), anchor=left of south west,
    width=15pc, height=15pc, grid =major,
    xstep=1, xmin=23.25, xmax=42.75,
    ystep=1, ymin=35.375, ymax=38.625,
   legend pos=south west,
    xlabel ={Quantization parameter (QP)}, xtick={24,27,30,33,36,39,42}, x label style={at={(0.5, 0.05)}, font=\footnotesize}, x tick style={font=\footnotesize}, x tick label style={font=\footnotesize},
    ylabel ={Average PSNR of rendered views [dB]}, y label style={at={(0.07,0.5)}, font={\footnotesize}}, y tick style={font=\footnotesize}, y tick label style={font=\footnotesize},
    ytick={35.50,36.00,36.50,37.00,37.50,38.00,38.50},
    y tick label style={font=\footnotesize, /pgf/number format/.cd, fixed, fixed zerofill, precision=1, /tikz/.cd},
    legend cell align=left, legend style={font=\footnotesize},
   ]
  \addplot[red  , thick, mark=square* ,mark options={fill=red  }]  plot coordinates {(24, 37.11)(30,37.00)(36,36.78)(42,36.37)};% VSRS OK
  \addplot[black, thick, mark=*,       mark options={fill=black}]  plot coordinates {(24, 37.50)(30,37.46)(36,37.36)(42,37.00)};% IMDCT VSRS OK
  \addplot[black, thick, mark=*,       mark options={fill=white}]  plot coordinates {(24, 37.85)(30,37.79)(36,37.60)(42,37.12)};% CAVS OK
  \legend{VSRS 3.5,IVDCT/ED$\rightarrow$VSRS3.5,CAVS};
  \end{axis}
  % balloons
  \begin{axis}[name=balloons,  at=(kendo.right of south east), anchor=left of south west,
    width=15pc, height=15pc, grid =major,
    xstep=1, xmin=23.25, xmax=42.75,
    ystep=1, ymin=33.875, ymax=37.125,
    legend pos=south west,
    xlabel ={Quantization parameter (QP)}, xtick={24,27,30,33,36,39,42}, x label style={at={(0.5, 0.05)}, font=\footnotesize}, x tick style={font=\footnotesize}, x tick label style={font=\footnotesize},
    ylabel ={Average PSNR of rendered views [dB]}, y label style={at={(0.07,0.5)}, font={\footnotesize}}, y tick style={font=\footnotesize}, y tick label style={font=\footnotesize},
    ytick={34.00,34.50,35.00,35.50,36.00, 36.50,37.00},
    y tick label style={font=\footnotesize, /pgf/number format/.cd, fixed, fixed zerofill, precision=1, /tikz/.cd},
    legend cell align=left, legend style={font=\footnotesize},
    ]
  \addplot[red  , thick, mark=square*  ,mark options={fill=red  }] plot coordinates {(24, 36.21)(30, 36.09)(36, 35.86)(42, 35.30)};% VSRS OK
  \addplot[black, thick, mark=*,       mark options={fill=black}] plot coordinates {(24, 36.44)(30, 36.35)(36, 36.20)(42, 35.80)};% IMDCT-> VSRS % \mu =0.5 OK
  \addplot[black, thick, mark=*,       mark options={fill=white}] plot coordinates {(24, 36.62)(30, 36.53)(36, 36.34)(42, 35.90)};% CAVS \mu =0.5 OK
  \legend{VSRS 3.5,IVDCT/ED$\rightarrow$VSRS3.5,CAVS};
  \end{axis}
  \node at(dancer.below south)  [anchor=north, align=center]{\footnotesize (a) Dancer.};
  \node at(kendo.below south)   [anchor=north, align=center]{\footnotesize (b) Kendo.};
  \node at(balloons.below south)[anchor=north, align=center]{\footnotesize (c) Balloons.};
  \end{tikzpicture}

  \begin{tikzpicture}
  % lovebird1
   \begin{axis}[name=lovebird1,
    width=15pc, height=15pc, grid =major,
    xstep=1, xmin=23.25, xmax=42.75,
    ystep=1, ymin=26.875, ymax=30.125,
    legend pos=south west,
    xlabel ={Quantization parameter (QP)}, xtick={24,27,30,33,36,39,42}, x label style={at={(0.5, 0.05)}, font=\footnotesize}, x tick style={font=\footnotesize}, x tick label style={font=\footnotesize},
    ylabel ={Average PSNR of rendered views [dB]}, y label style={at={(0.07,0.5)}, font={\footnotesize}}, y tick style={font=\footnotesize}, y tick label style={font=\footnotesize},
    ytick={27.00,27.50,28.00,28.50,29.00,29.50,30.00},
    y tick label style={font=\footnotesize, /pgf/number format/.cd, fixed, fixed zerofill, precision=1, /tikz/.cd},
    legend cell align=left, legend style={font=\footnotesize},
    ]
  \addplot[red  , thick, mark=square* ,mark options={fill=red  }]  plot coordinates {(24, 28.85)(30, 28.77)(36, 28.75)(42, 28.73)};% VSRS OK
  \addplot[black, thick, mark=*,       mark options={fill=black}]  plot coordinates {(24, 29.04)(30, 29.03)(36, 29.00)(42, 28.92)};% IMDCT -> VSRS
  \addplot[black, thick, mark=*,       mark options={fill=white}]  plot coordinates {(24, 29.10)(30, 29.07)(36, 29.06)(42, 29.02)};% CAVS
  \legend{VSRS 3.5,IVDCT/ED$\rightarrow$VSRS3.5,CAVS};
  \end{axis}

  % newspaper
   \begin{axis}[name=newspaper,  at=(lovebird1.right of south east), anchor=left of south west,
    width=15pc, height=15pc, grid =major,
    xstep=1, xmin=23.25, xmax=42.75,
    ystep=1, ymin=30.375, ymax=33.625,
   legend pos=south west,
    xlabel ={Quantization parameter (QP)}, xtick={24,27,30,33,36,39,42}, x label style={at={(0.5, 0.05)}, font=\footnotesize}, x tick style={font=\footnotesize}, x tick label style={font=\footnotesize},
    ylabel ={Average PSNR of rendered views [dB]}, y label style={at={(0.07,0.5)}, font={\footnotesize}}, y tick style={font=\footnotesize}, y tick label style={font=\footnotesize},
    ytick={30.50,31.00,31.50,32.00,32.50,33.00,33.50},
    y tick label style={font=\footnotesize, /pgf/number format/.cd, fixed, fixed zerofill, precision=1, /tikz/.cd},
    legend cell align=left, legend style={font=\footnotesize},
    ]
  \addplot[red  , thick, mark=square* ,mark options={fill=red  }]  plot coordinates {(24, 32.50)(30,32.46)(36, 32.34)(42, 31.85)};\label{p1}% VSRS $\mu=0.5$ OK
  \addplot[black, thick, mark=*,       mark options={fill=black}]  plot coordinates {(24, 32.80)(30,32.75)(36, 32.73)(42, 32.46)};\label{p2}% IMDCT-> VSRS OK
  \addplot[black, thick, mark=*,       mark options={fill=white}]  plot coordinates {(24, 33.41)(30,33.40)(36, 33.21)(42, 32.55)};\label{p3}% CAVS OK
  \legend{VSRS 3.5,IVDCT/ED$\rightarrow$VSRS3.5,CAVS};
  \end{axis}
  \node at(lovebird1.below south)[anchor=north, align=center]{\footnotesize (d) Lovebird1.};
  \node at(newspaper.below south)[anchor=north, align=center]{\footnotesize (e) Newspaper.};
  \end{tikzpicture}
  \caption{\label{fig:qp_results}Objective quality of rendered views over the quantization parameter that is used for depth map coding with H.264/AVC. Both VSRS 3.5 using IVDCT enhanced depth maps (IVDCT/ED$\rightarrow$VSRS3.5) and CAVS using MPEG depth maps outperform VSRS 3.5 using MPEG depth maps.}\vspace{-2ex}
\end{figure*}
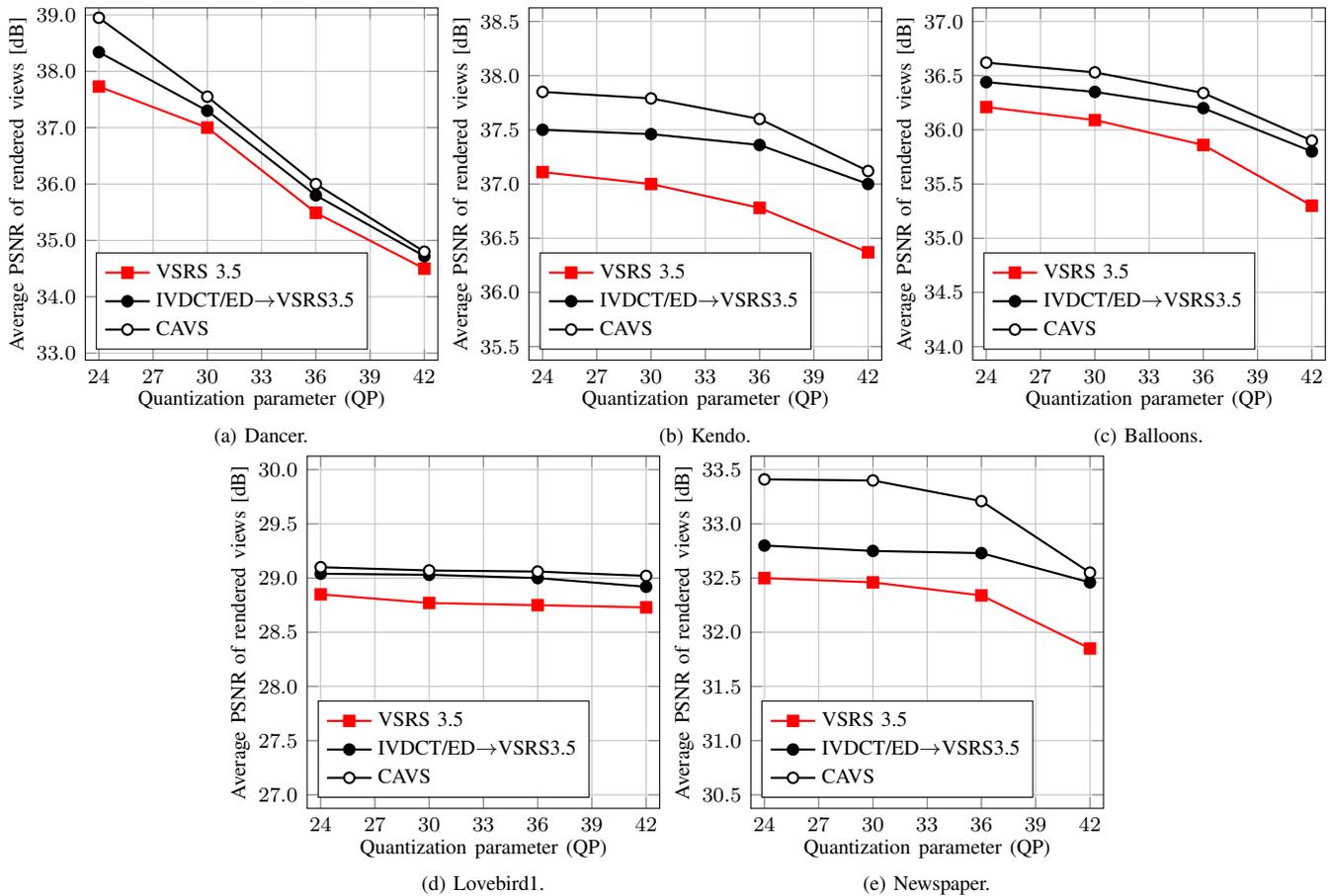

To assess our proposed algorithms at the receiver end, we coded the three reference depth maps at four quantization parameters by using the multiview coding extension of H.264/AVC~\cite{Vetro2011}. The coded depth maps are reconstructed and used for the virtual view synthesis by VSRS 3.5. The objective quality of the synthesized view is measured in terms of PSNR (in dB). Next, the reconstructed depth maps are enhanced by our IVDCT based depth enhancement scheme. The resulting enhanced reconstructed depth maps  are used for virtual view synthesis via VSRS 3.5. The quality of synthesized views using reconstructed depth maps and enhanced reconstructed depth maps at four different quantization parameters are plotted in Fig.~\ref{fig:qp_results}. We observe that our depth enhancement algorithm gives significant improvements in the quality of the synthesized views when compared to decoded only depth maps. However, our CAVS gives further improvements in the quality of synthesised views when using the reconstructed depth maps, as depicted for five test sequences in Fig.~\ref{fig:qp_results}. Thus, our proposed algorithm is also beneficial for coded MVD imagery.

\subsection{White Gaussian Noise}
\label{subsec:wgn}
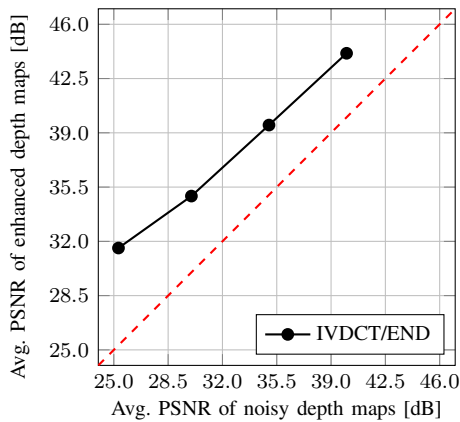
\begin{figure}[t]
  \centering
  \begin{tikzpicture}
    \begin{axis}[name=e-depthmap,
    width=15pc, height=15pc, grid =major,
    xstep=1, xmin=24, xmax=47,
    ystep=1, ymin=24, ymax=47,
    legend pos=south east,
    ylabel ={\hspace{-2ex}Avg. PSNR of enhanced depth maps~[dB]},
    xlabel ={Avg. PSNR of noisy depth maps [dB]},
    y label style={at={(0.04,0.5)}}, legend cell align=left, legend style={font=\footnotesize}, x tick style={font=\footnotesize}, y tick style={font=\footnotesize}, y label style={font=\footnotesize}, x label style={at={(0.5, 0.04)},font=\footnotesize}, xtick={25.00,28.50,32.00,35.50,39.00,42.50,46.00}, ytick={25.00,28.50,32.00,35.50,39.00,42.50,46.00},
    x tick label style={font=\footnotesize, /pgf/number format/.cd, fixed, fixed zerofill, precision=1, /tikz/.cd},
    y tick label style={font=\footnotesize, /pgf/number format/.cd, fixed, fixed zerofill, precision=1, /tikz/.cd},
    ]
    \addplot[black, thick, mark=*, mark options={fill=black}] plot coordinates {(25.30,31.57)(30.00,34.92)(35.00,39.50)(40.00, 44.13)}; % iMDCT $\mu = 0.5$
    \addplot[red, thick, dashed] plot coordinates {(24,24)(47,47)};
    \legend{IVDCT/END};
  \end{axis}
  \end{tikzpicture}
  \caption{\label{fig:noisy depth enhancement}Objective quality of enhanced noisy depth maps by using the IVDCT algorithm (IVDCT/ED) over objective quality of noisy depth maps. For this, three MPEG Dancer sequence depth maps are used as clean depth maps.}
\end{figure}

\begin{figure}[t!]
\centering
\begin{tikzpicture}
    \begin{axis}[name=rendering,grid =major,
     width=15pc, height=15pc,
     xstep=1, xmin=24.5, xmax=40.5,
     ystep=1, ymin=28.0, ymax=38.5,
     legend pos=south east,
     ylabel = {Avg. PSNR of rendered views [dB]},
     xlabel = {Avg. PSNR of noisy depth maps [dB]}, y label style={at={(0.04,0.5)}},
          legend cell align=left,
      xtick={25.00,27.50,30.00,32.50,35.00,37.50, 40.00}, ytick ={28.5000, 30.0833,31.6667,33.2500,34.8333,36.4167, 38.00},
     legend style={font=\footnotesize}, x tick style={font=\footnotesize}, y tick style={font=\footnotesize},
     x label style={at={(0.5, 0.04)},font=\footnotesize}, y label style={font=\footnotesize},
     x tick label style={font=\footnotesize, /pgf/number format/.cd, fixed, fixed zerofill, precision=1, /tikz/.cd},
     y tick label style={font=\footnotesize, /pgf/number format/.cd, fixed, fixed zerofill, precision=1, /tikz/.cd},
     ]
     \addplot[red  , thick, mark=square*, mark options={fill=red}  ] plot coordinates {(25.30,28.50)(30.00,31.55)(35.00,34.60)(40.00,37.00)};% vsrs
     \addplot[black, thick, mark=*      , mark options={fill=black}] plot coordinates {(25.30,31.57)(30.00,34.20)(35.00,36.63)(40.00,37.97)};% iMDCT-> VSRS 3.5
     \addplot[black, thick, mark=*      , mark options={fill=white}] plot coordinates {(25.30,32.25)(30.00,34.80)(35.00,37.00)(40.00,38.20)};% cavs
    \legend{VSRS 3.5, IVDCT/ED$\rightarrow$VSRS 3.5, CAVS};
    \end{axis}
\end{tikzpicture}
\caption{\label{fig:noisy depth enhancement_vsrs}Objective quality of rendered views by CAVS using noisy depth maps, by VSRS 3.5 using IVDCT enhanced noisy depth maps (IVDCT/ED), and by VSRS 3.5 using noisy depth maps over objective quality of noisy depth maps. Again, three MPEG Dancer depth maps are used as clean depth maps.}
\end{figure}
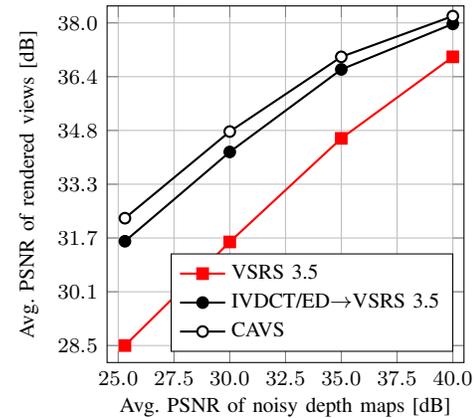

To investigate the efficiency  of our proposed algorithms, we generate noisy depth maps from ground-truth depth maps of the Dancer sequence at three viewpoints by adding white Gaussian noise (AWGN) with variance $\sigma_n^2 = 0.0001, 0.0003, 0.001, 0.003$. The noisy depth maps are enhanced by our proposed algorithm. The ground-truth depth maps are used to assess the quality of the resulting enhanced depth maps in terms of PSNR (in dB). Fig.~\ref{fig:noisy depth enhancement} shows the average PSNR of the enhanced noisy depth maps with respect to the quality of the noisy depth maps. The proposed algorithm offers gains between 4 and 6 dB when compared to the quality of the noisy depth maps.  Note, for such noisy depth maps, the enhancement algorithm mostly averages depth hypotheses from different viewpoints adaptively as per inter-view consistency information.  Moreover, with an increasing number of available depth hypotheses, the efficiency of our algorithm improves. Fig.~\ref{fig:noisy depth enhancement_vsrs} shows the average PSNR of rendered views by VSRS 3.5 using the enhanced noisy depth maps which have originally been degraded by AWGN. The enhancement of highly noisy depth maps is beneficial for view synthesis, as we observe also an improvement in rendering quality. On the other hand, the enhancement of high-quality depth maps is of limited benefit for view synthesis. Nevertheless, we observe further gains in the quality of virtual views as generated by CAVS when using noisy depth maps, as shown in Fig.~\ref{fig:noisy depth enhancement}. This confirms again the limitations of VSRS 3.5.

With both objective and subjective results, we have demonstrated the efficiency of our consistency information for view synthesis and our depth map enhancement algorithm. Hence, depth consistency testing is a promising approach to offer a better visual experience to FTV users. Moreover, inter-view consistency across multiple viewpoints is relevant for high-quality view synthesis.

\section{Conclusions}
\label{sec:conclusion}
This paper proposes a novel algorithm for depth consistency testing in depth difference subspace. It improves the inter-view depth consistency at a given viewpoint by testing multiple depth hypotheses from various reference viewpoints. With this improved depth consistency, we are able to enhance the visual experience of FTV. Further, we utilize the consistency information to enhance the depth representation at multiple viewpoints, and hence, the visual quality of synthesized views. Both objective and subjective results demonstrate the effectiveness of the presented consistency testing algorithm. In experiments, we compare the visual quality of synthesized views between our approach and convectional view synthesis algorithms such as MPEG VSRS 3.5. The visual quality of novel views is improved by both consistency-based view synthesis and depth map enhancement. Gains of up to 1.4 dB have been observed for MPEG test sequences.

\section*{Acknowledgment}
This work has been supported in part by Ericsson AB and the ACCESS Linnaeus Centre at KTH Royal Institute of Technology, Stockholm, Sweden.

\bibliographystyle{IEEEtran}
\bibliography{./bib/reference,IEEEabrv}

\begin{thebibliography}{10}
\providecommand{\url}[1]{#1}
\csname url@samestyle\endcsname
\providecommand{\newblock}{\relax}
\providecommand{\bibinfo}[2]{#2}
\providecommand{\BIBentrySTDinterwordspacing}{\spaceskip=0pt\relax}
\providecommand{\BIBentryALTinterwordstretchfactor}{4}
\providecommand{\BIBentryALTinterwordspacing}{\spaceskip=\fontdimen2\font plus
\BIBentryALTinterwordstretchfactor\fontdimen3\font minus
  \fontdimen4\font\relax}
\providecommand{\BIBforeignlanguage}[2]{{%
\expandafter\ifx\csname l@#1\endcsname\relax
\typeout{** WARNING: IEEEtran.bst: No hyphenation pattern has been}%
\typeout{** loaded for the language `#1'. Using the pattern for}%
\typeout{** the default language instead.}%
\else
\language=\csname l@#1\endcsname
\fi
#2}}
\providecommand{\BIBdecl}{\relax}
\BIBdecl

\bibitem{Tanimoto2011}
M.~Tanimoto, M.~P. Tehrani, T.~Fujii, and T.~Yendo, ``Free-viewpoint {TV},''
  \emph{IEEE Signal Process. Mag.}, vol.~28, no.~1, pp. 67--76, Jan. 2011.

\bibitem{Urey2011}
H.~Urey, K.~Chellappan, E.~Erden, and P.~Surman, ``State of the art in
  stereoscopic and autostereoscopic displays,'' \emph{Proc. IEEE}, vol.~99,
  no.~4, pp. 540--555, Apr. 2011.

\bibitem{Chai2000}
J.-X. Chai, X.~Tong, S.-C. Chan, and H.-Y. Shum, ``Plenoptic sampling,'' in
  \emph{Proc. SIGGRAPH Conf. Computer Graphics and Interactive Techniques}, New
  York, USA, 2000, pp. 307--318.

\bibitem{Benzie2007}
P.~Benzie, J.~Watson, P.~Surman, I.~Rakkolainen, K.~Hopf, H.~Urey, V.~Sainov,
  and C.~von Kopylow, ``A survey of {3DTV} displays: Techniques and
  technologies,'' \emph{IEEE Trans. Circuits Syst. Video Technol.}, vol.~17,
  no.~11, pp. 1647--1658, Nov. 2007.

\bibitem{Flierl2007}
M.~Flierl and B.~Girod, ``Multiview video compression,'' \emph{IEEE Signal
  Process. Mag.}, vol.~24, no.~6, pp. 66--76, Nov. 2007.

\bibitem{Girod2003}
M.~Magnor, P.~Ramanathan, and B.~Girod, ``Multi-view coding for image-based
  rendering using {3-D} scene geometry,'' \emph{IEEE Trans. Circuits Syst.
  Video Technol.}, vol.~13, no.~11, pp. 1092--1106, Nov. 2003.

\bibitem{Smolic2007}
A.~Smolic, K.~M$\ddot{u}$ller, N.~Stefanoski, J.~Ostermann, A.~Gotchev,
  G.~Akar, G.~Triantafyllidis, and A.~Koz, ``Coding algorithms for {3DTV}--{A}
  survey,'' \emph{IEEE Trans. Circuits Syst. Video Technol.}, vol.~17, no.~11,
  pp. 1606--1621, Nov. 2007.

\bibitem{Vetro2011}
A.~Vetro, T.~Wiegand, and G.~Sullivan, ``Overview of the stereo and multiview
  video coding extensions of the {H.264/MPEG-4} {AVC} standard,'' \emph{Proc.
  IEEE}, vol.~99, no.~4, pp. 626--642, Apr. 2011.

\bibitem{Mueller2011}
K.~M$\ddot{u}$ller, P.~Merkle, and T.~Wiegand, ``{3-D} video representation
  using depth maps,'' \emph{Proc. IEEE}, vol.~99, no.~4, pp. 643--656, Apr.
  2011.

\bibitem{Fehn2004}
C.~Fehn, ``Depth-image-based rendering {(DIBR)}, compression, and transmission
  for a new approach on {3D-TV},'' in \emph{Stereoscopic Displays and Virtual
  Reality Systems XI}, vol. 5291, no.~1.\hskip 1em plus 0.5em minus 0.4em\relax
  San Jose, CA, USA: SPIE, Jan. 2004, pp. 93--104.

\bibitem{Scharstein2002}
D.~Scharstein and R.~Szeliski, ``A taxonomy and evaluation of dense two-frame
  stereo correspondence algorithms,'' \emph{Int. J. Computer Vision}, vol.~47,
  pp. 7--42, Apr. 2002.

\bibitem{Mueller2008}
K.~M$\ddot{u}$ller, A.~Smolic, K.~Dix, P.~Kauff, and T.~Wiegand,
  ``Reliability-based generation and view synthesis in layered depth video,''
  in \emph{Proc. IEEE Int. Workshop Multimedia Signal Process.}, Cairns,
  Queensland, Australia, Oct. 2008, pp. 34--39.

\bibitem{Rana2011}
P.~K. Rana and M.~Flierl, ``View interpolation with structured depth from
  multiview video,'' in \emph{Proc. European Signal Process. Conf.}, Barcelona,
  Spain, Aug. 2011, pp. 383--387.

\bibitem{Ishibashi2012}
T.~Ishibashi, M.~Tehrani, T.~Fujii, and M.~Tanimoto, ``{FTV} format using
  global view and depth map,'' in \emph{Proc. Picture Coding Symp.}, Krakow,
  Poland, May 2012, pp. 29 --32.

\bibitem{Suzuki2013}
K.~Suzuki and M.~Tanimoto, ``{AHG08}: {T}echnical description of {GVD} (global
  view and depth) {3D} format,'' JCT-3V ITU-T SG 16 WP 3 and MPEG ISO/IEC JTC
  1/SC 29/WG 11, Geneva, Switzerland, Tech. Rep. JCT3V-C0058 M27793, Jan. 2013.

\bibitem{Cigla2009}
C.~Cigla and A.~Alatan, ``Temporally consistent dense depth map estimation via
  belief propagation,'' in \emph{3DTV Conf.}, Potsdam, Germany, May 2009, pp.
  1--4.

\bibitem{Lee2010}
S.~Lee and Y.~Ho, ``Temporally consistent depth map estimation using motion
  estimation for {3DTV},'' in \emph{Int. Workshop on Advanced Image Technol.},
  Kuala Lumpur, Malaysia, Jan. 2010, pp. 149(1--6).

\bibitem{Fu2010}
D.~Fu, Y.~Zhao, and L.~Yu, ``Temporal consistency enhancement on depth
  sequences,'' in \emph{Proc. Picture Coding Symp.}, Nagoya, Japan, Dec. 2010,
  pp. 342--345.

\bibitem{Rana2010}
P.~K. Rana and M.~Flierl, ``Depth consistency testing for improved view
  interpolation,'' in \emph{Proc. IEEE Int. Workshop Multimedia Signal
  Process.}, St. Malo, France, Oct. 2010, pp. 384--389.

\bibitem{MPEG:N12036}
MPEG, ``Call for proposals on {3D} video coding technology,''
  {ISO/IEC~JTC1/SC29/WG11}, Geneva, Switzerland, Tech. Rep. N12036, Mar. 2011.

\bibitem{Ekmekcioglu2011}
E.~Ekmekcioglu, V.~Velisavljevi\'{c}, and S.~Worrall, ``Content adaptive
  enhancement of multi-view depth maps for free viewpoint video,'' \emph{IEEE
  J. Sel. Topics Signal Process.}, vol.~5, no.~2, pp. 352--361, Apr. 2011.

\bibitem{Kurc2012}
M.~Kurc, O.~Stankiewicz, and M.~Domanski, ``Depth map inter-view consistency
  refinement for multiview video,'' in \emph{Proc. Picture Coding Symp.},
  Krakow, Poland, May 2012, pp. 137--140.

\bibitem{Li2012}
R.~Li, D.~Rusanovskyy, M.~M. Hannuksela, and H.~Li, ``Joint view filtering for
  multiview depth map sequences,'' in \emph{Proc. IEEE Int. Conf. Image
  Process.}, Orlando, USA, Sept. 2012, pp. 1329--1332.

\bibitem{Rusanovskyy2011}
D.~Rusanovskyy and M.~M. Hannuksela, ``Description of {3D} video coding
  technology proposal by {Nokia},'' ISO/IEC JTC1/SC29/WG11, Geneva,
  Switzerland, Tech. Rep. M22552, Nov. 2011.

\bibitem{Hannuksela2012}
{M. M. Hannuksela, Y. Chen, and T. Suzuki}, ``{3D-AVC} draft text 3,'' JCT-3V
  ITU-T SG 16 WP 3 and MPEG ISO/IEC JTC 1/SC 29/WG 11, Stockholm, Sweden, Tech.
  Rep. JCT3V-A1002, Jul. 2012.

\bibitem{Nokia}
{NOKIA}, ``{MPEG} {MVC}+{D} and {3D}-{AVC} based reference {3DV}-{ATM}
  software,'' [Online]: http://mpeg3dv.research.nokia.com/svn/mpeg3dv/.

\bibitem{Rana2012A}
P.~K. Rana and M.~Flierl, ``Depth pixel clustering for consistency testing of
  multiview depth,'' in \emph{Proc. European Signal Process. Conf.}, Bucharest,
  Romania, Aug. 2012, pp. 1119--1123.

\bibitem{Kolmogorov2004}
Y.~Boykov and V.~Kolmogorov, ``An experimental comparison of min-cut/max-flow
  algorithms for energy minimization in vision,'' \emph{IEEE Trans. Pattern
  Anal. Mach. Intell.}, vol.~26, no.~9, pp. 1124--1137, Sep. 2004.

\bibitem{Sun2003}
J.~Sun, N.-N. Zheng, and H.-Y. Shum, ``Stereo matching using belief
  propagation,'' \emph{IEEE Trans. Pattern Anal. Mach. Intell.}, vol.~25,
  no.~7, pp. 787--800, Jul. 2003.

\bibitem{Felzenszwalb2004A}
P.~Felzenszwalb and D.~Huttenlocher, ``Efficient belief propagation for early
  vision,'' in \emph{Proc. IEEE Conf. Computer Vision and Pattern Recognition},
  vol.~1, Washington, DC, USA, Jun. 2004, pp. 261--268.

\bibitem{Klaus2006}
A.~Klaus, M.~Sormann, and K.~Karner, ``Segment-based stereo matching using
  belief propagation and a self-adapting dissimilarity measure,'' in
  \emph{Proc. Int. Conf. Pattern Recognition}, vol.~3, Hong Kong, China, Aug.
  2006, pp. 15--18.

\bibitem{Yang2009}
Q.~Yang, L.~Wang, R.~Yang, H.~Stewenius, and D.~Nister, ``Stereo matching with
  color-weighted correlation, hierarchical belief propagation, and occlusion
  handling,'' \emph{IEEE Trans. Pattern Anal. Mach. Intell.}, vol.~31, no.~3,
  pp. 492--504, Mar. 2009.

\bibitem{Cigla2007B}
C.~Cigla, X.~Zabulis, and A.~A. Aydin, ``Segment-based stereo-matching via
  plane and angle sweeping,'' in \emph{3DTV Conf.}, Kos Island, Greece, May
  2007, pp. 1--4.

\bibitem{MPEG:M16923}
M.~Tanimoto, T.~Fujii, M.~Panahpour, and M.~Wildeboer, ``Depth estimation
  reference software {DERS} 5.0,'' ISO/IEC JTC1/SC29/WG11, Xian, China, Tech.
  Rep. M16923, Oct. 2009.

\bibitem{Zisserman2004}
R.~I. Hartley and A.~Zisserman, \emph{Multiple View Geomtery in Computer
  Vision}, 2nd~ed.\hskip 1em plus 0.5em minus 0.4em\relax Cambridge, UK:
  Cambridge University Press, 2004.

\bibitem{Tian2009}
D.~Tian, P.-L. Lai, P.~Lopez, and C.~Gomila, ``View synthesis techniques for
  {3D} video,'' in \emph{Proc. SPIE}, ser. Applications of Digital Image
  Processing XXXII, A.~G. Tescher, Ed., vol. 7443, San Diego, CA, USA, Sep.
  2009, pp. 1--11.

\bibitem{Flierl2002}
M.~Flierl, T.~Wiegand, and B.~Girod, ``Rate-constrained multihypothesis
  prediction for motion-compensated video compression,'' \emph{IEEE Trans.
  Circuits Syst. Video Technol.}, vol.~12, pp. 957--969, Nov. 2002.

\bibitem{Bertalmio2001}
M.~Bertalmio, A.~Bertozzi, and G.~Sapiro, ``Navier-stokes, fluid dynamics, and
  image and video inpainting,'' in \emph{Proc. IEEE CS Conf. Computer Vision
  and Pattern Recognition}, vol.~1, Kauai, HI, USA, Dec. 2001, pp. 355--362.

\bibitem{MPEG:M15377}
M.~Tanimoto, T.~Fujii, K.~Suzuki, N.~Fukushima, and Y.~Mori, ``Reference
  softwares for depth estimation and view synthesis,'' ISO/IEC JTC1/SC29/WG11,
  Archamps, France, Tech. Rep. M15377, Apr. 2008.

\end{thebibliography}

\end{document}